\documentclass[10pt,letterpaper]{article}

\usepackage{xcolor} % xcolor handles everything color-related; 'color' is redundant
\definecolor{navy}{rgb}{0,0.2,0.6}
\definecolor{navy-cap}{rgb}{0,0.1,0.7}
\definecolor{lgray}{gray}{0.90}
\definecolor{lightgreen}{rgb}{0.9,1,0.9}
\definecolor{darkgreen}{rgb}{0,.7,0}
\definecolor{lightorange}{rgb}{1,0.95,0.8}
\definecolor{mauve}{rgb}{0.58,0,0.82}
\definecolor{custom-orange}{rgb}{0.9,0.5,0} % Renamed to avoid overriding default 'orange'

\usepackage[margin=1in]{geometry}
\usepackage[T1]{fontenc}
\usepackage[utf8]{inputenc}
\usepackage{newtxtext}
\usepackage{newtxmath}
\usepackage{textcomp}
\usepackage{setspace}
\usepackage{soul}
\usepackage{dirtytalk} % Provides \say{...} for clean nested quotes

\usepackage{fancyhdr}
\usepackage[bottom, hang, flushmargin]{footmisc} % Pushes footnotes to absolute bottom
\usepackage{titlesec}
\titleformat{\section}{\fontsize{11}{15}\selectfont\color{navy}\bfseries}{\thesection}{1em}{}
\titleformat{\subsection}{\fontsize{11}{15}\selectfont\color{navy}\bfseries}{\thesubsection}{1em}{}
\titleformat{\subsubsection}{\fontsize{11}{15}\selectfont\color{navy}\bfseries}{\thesubsubsection}{1em}{}

\titlespacing\section{0pt}{6pt plus 2pt minus 2pt}{2pt plus 1pt}
\titlespacing\subsection{0pt}{6pt plus 2pt minus 2pt}{2pt plus 1pt}
\titlespacing\subsubsection{0pt}{6pt plus 2pt minus 2pt}{2pt plus 1pt}

\usepackage{bm}
\usepackage{mathtools} % Automatically loads amsmath with modern fixes

\DeclareMathOperator*{\argmin}{argmin}

\usepackage{booktabs} % Professional-grade horizontal lines (\toprule, \midrule, \bottomrule)
\usepackage{longtable} % Multi-page table support
\usepackage{array} % Advanced column definitions
\usepackage{multirow} % Cells spanning multiple rows
\usepackage{makecell} % Multi-line cells and padding
\usepackage{xltabular}

\usepackage{caption}
\DeclareCaptionFont{subtxt}{\fontsize{11pt}{16pt}\selectfont}
\usepackage{subcaption}
\usepackage{url}
\usepackage{hyperref}
\hypersetup{linktoc=all, colorlinks=true, linkcolor=navy-cap, citecolor=blue, urlcolor=navy} % linktoc=all is what makes the page numbers clickable
\usepackage[open, openlevel=2, atend, numbered]{bookmark} % Highly optimized PDF outline builder
\usepackage[nameinlink]{cleveref} % Smart cross-referencing (e.g., \cref{fig:X})
\Crefname{figure}{Fig.}{Figs.}

\usepackage{tocloft}
\usepackage[backend=biber, style=authoryear, sorting=nyt, maxbibnames=10, maxcitenames=2, uniquelist=false, giveninits=true, dashed=false]{biblatex}
\let\cite\parencite
\providecommand{\fb}{\boldsymbol{f}}

\providecommand{\mb}{\boldsymbol{m}}
\providecommand{\nb}{\boldsymbol{n}}

\providecommand{\tb}{\boldsymbol{t}}
\providecommand{\ub}{\boldsymbol{u}}

\providecommand{\xb}{\boldsymbol{x}}

\providecommand{\Bb}{\boldsymbol{B}}
\providecommand{\Cb}{\boldsymbol{C}}

\providecommand{\Fb}{\boldsymbol{F}}
\providecommand{\Ib}{\boldsymbol{I}}
\providecommand{\Jb}{\boldsymbol{J}}

\providecommand{\Mb}{\boldsymbol{M}}
\providecommand{\Rb}{\boldsymbol{R}}

\providecommand{\Ub}{\boldsymbol{U}}

\providecommand{\Xb}{\boldsymbol{X}}

\providecommand{\epsilonb}{\boldsymbol{\varepsilon}}
\providecommand{\sigmab}{\boldsymbol{\sigma}}

\providecommand{\thetab}{\boldsymbol{\theta}}
\providecommand{\xib}{\boldsymbol{\xi}}

\providecommand{\CrefSub}[2]{Fig.~\ref{#1}(\subref{#2})}
\providecommand{\CrefSubRange}[3]{Figs.~\ref{#1}(\subref{#2})--(\subref{#3})}

\providecommand{\nimm}{\mathrm{mm}}

\providecommand{\Res}{\operatorname{Res}}

\newif\ifanon
\anonfalse

\begin{document}
    \sloppy 
    \setcounter{page}{1}
    \begin{center}
    {\Large\bfseries Scalable Stress-Constrained Topology Optimization: A Simultaneous Approach Based on Machine Learning\par}
    \vspace{0.75em}
    \ifanon
    \vspace{0.35em}
    \else
    Xiangyu Sun\textsuperscript{1,3}, Shirin Hosseinmardi\textsuperscript{1}, Ramin Bostanabad\textsuperscript{1,2,*}\par
    \vspace{0.35em}
    {\small \textsuperscript{1}Department of Mechanical and Aerospace Engineering, University of California, Irvine, CA, USA\par
    \vspace{0.35em}
    \textsuperscript{2}Department of Civil and Environmental Engineering, University of California, Irvine, CA, USA\par
    \textsuperscript{3}Department of Mechanical Engineering, Northwestern University, Evanston, IL, USA\par
    \textsuperscript{*}Corresponding Author: \href{mailto:raminb@uci.edu}{raminb@uci.edu} \par
    \href{https://github.com/Bostanabad-Research-Group/Scalable-Stress-TO}{GitHub Repository: https://github.com/Bostanabad-Research-Group/Scalable-Stress-TO}\par}
    \fi
    \end{center}
    \vspace{0.5em}
    \noindent \textbf{Abstract}\\
    \noindent
    Large-scale topology optimization (TO) remains computationally demanding, especially for stress-constrained problems.
    To address this challenge, we propose the simultaneous neural topology optimization (SNTO) framework where state, adjoint, and design variables are represented with custom neural networks (NNs) whose parameters are \textit{jointly} optimized within a single loop using adjoint analysis and automatic differentiation.
    The loss function combines the design objective, an aggregation of the local stress constraints, and the potential energy functionals.
    Our NNs have a unique architecture that accelerates convergence while capturing the sharp gradients and boundaries which are critical to stress-constrained TO.
    To scale to millions of degrees of freedom, we develop a hardware-aware training scheme in which the design domain is partitioned into subdomains distributed across multiple graphics processing units (GPUs).
    To control peak GPU memory, we further slice each subdomain into smaller regions whose loss gradients are evaluated sequentially and accumulated.
    Unlike domain decomposition, our partitioning \textit{naturally} ensures field continuity across subdomains without additional loss terms.
    We first evaluate SNTO on two-dimensional (2D) stress-constrained benchmarks and compare the results to a conventional finite element-based approach, showing that SNTO produces designs with sharper boundaries and smaller gray areas.
    We then solve three-dimensional (3D) problems with up to eight million collocation points on eight GPUs.
    The training time decreases nearly in proportion to the number of GPUs and the peak GPU memory is reduced by increasing the number of sub-slices.
    Overall, SNTO offers an effective pathway to large-scale stress-constrained TO on limited computational resources.

    \noindent \textbf{Keywords:} Topology Optimization, Stress Constraints, Physics-Informed Machine Learning, Neural Networks, Multi-GPU Training, Adjoint Sensitivity.

    \section{Introduction}\label{sec intro}
Topology optimization (TO) determines how material should be distributed in a design domain by optimizing an objective function subject to the governing physics and design constraints. Since the seminal work of \textcite{bendsoe_generating_1988}, TO has matured into a standard design tool in many disciplines such as structural mechanics, heat transfer, fluid flow, and multi-physics systems \cite{bendsoe_topology_2004,borrvall_topology_2003,zhu_topology_2016,wu_advances_2021,yousefpour_physics_2026}. Conventional TO methods include the solid isotropic material with penalization (SIMP) approach \cite{bendsoe_optimization_1995,andreassen_efficient_2011}, the level-set method \cite{van_level_2013,chen_topology_2017}, and evolutionary structural optimization \cite{tanskanen_evolutionary_2002}, which have been extended to more advanced design problems with multiple materials \cite{zuo_mm_2017,sanders_polymat_2018,da_mm_2022,xu_stress_2021,han_multi_2025}, coupled physics \cite{giraldo_multi_2020,wang_multi_2023,wang_topology_2024}, fracture and plasticity \cite{jia_controlling_2023,han_general_2025}, and manufacturing uncertainties \cite{da_stress_2019}. In most of these settings, the objective is a global quantity (e.g., compliance or actuation displacement) and the design constraints are a few scalars such as the volume fraction. The problem becomes far more challenging when the design constraints are local and large in number.

Stress constraints are among the most popular local constraints. Since the stress limit must be satisfied at every spatial point, the number of constraints scales with the resolution of the discretization. Aggregation functions such as the p-norm or the Kreisselmeier--Steinhauser function condense the local constraints into one or a few global measures \cite{le_stress_2010,holmberg_stress_2013}. But the resulting approximation of the maximum stress can be loose, and its accuracy depends on the aggregation parameter and adaptive scaling \cite{da_silva_local_2021}. Augmented Lagrangian formulations instead keep every local constraint and handle them through a quadratic penalty with Lagrange multiplier estimates \cite{senhora_topology_2020,giraldo_polystress_2021}. In either case, the stress field is highly localized around re-entrant corners, crack tips, and loading regions where resolving the stress state requires fine meshes, further increasing the number of constraints.

Conventional stress-constrained TO methods follow a nested analysis and design (NAND) approach with three main steps: (1) the state variables are computed with the finite element (FE) method for a given design field; (2) the adjoint variables and the sensitivities are evaluated from the resulting stress distribution; and (3) the design is updated with the optimality criteria (OC) method \cite{bendsoe_optimization_1995} or the method of moving asymptotes (MMA) \cite{svanberg_method_1987}. 
Each iteration therefore requires two FE solves whose size grows with the mesh. The first solve gives the state field while the second gives the adjoint field. The stress must also be evaluated at every element in each iteration to build the adjoint load for solving the adjoint field. This cost grows with the number of local constraints.
For large-scale problems, especially in three dimensions (3D), this cost becomes prohibitive on a single machine. Large-scale implementations reduce this cost using distributed-memory parallelism and multigrid-preconditioned iterative solvers \cite{aage_topology_2015,alexandersen_large_2016,jia_fenitop_2024,alexandersen_large_2026} which have reached the billion-voxel scale in, e.g., wing design \cite{aage_giga_2017}. In these implementations, the FE mesh is partitioned into blocks, and each processor stores and operates on its own block of the mesh and the system matrices. The blocks are coupled through the nodes on their shared boundaries, whose values are exchanged between processors in every iteration of the solver so that the solution remains continuous across the partition. These frameworks are well developed and computationally efficient, but they require substantial software and hardware infrastructure, and their memory footprint and communication overhead are mostly set by the mesh and its partition rather than by the design problem itself.

Machine learning (ML) offers a different route that is not bound to a mesh since all the variables can be represented by continuous functions of the spatial coordinates rather than by nodal values on a mesh. 
To reduce the cost of TO, data-driven approaches train generative or direct design-prediction models \cite{RN941,RN965,RN1864,RN2026,xiang_topology_2026}, reinforcement learning agents \cite{wang_accelerated_2026}, or surrogate models for the state solution \cite{wang_versatile_2026,wang_data_2026}, sensitivity evaluation \cite{chi_universal_2021,senhora_machine_2022,li2026multi}, and super-resolution \cite{wang_deep_2021,lim_accelerating_2024}. The generative capabilities of these models are limited by the coverage of their training databases which are built by conventional solvers. Applications of generative models in TO have been limited to simple problems (e.g., compliance minimization) where the cost of generating the training dataset can be justified.

Physics-informed ML (PIML) removes the dependence on data by training the model against the governing equations which are represented as either a PDE system or its variational form such as the deep energy method (DEM) \cite{nguyen_DEM_2020,haghighat_pinn_2021}. PIML has been applied to TO through three strategies: (1) the design field is parameterized with a neural network (NN) while the state is computed with an FE solver \cite{hoyer_neural_2019,zhang_tonr_2021,chandrasekhar_MMTOuNN_2021,sanu_neural_2025,liu_neural_2025,kuszczak_meta_2026}; (2) the state variables are parameterized with NNs while the design field is updated with OC or MMA \cite{he_deep_TO_2023,yin_dynamically_2024,zhao_physics_2024}; and (3) both the state and design variables are parameterized with NNs and trained on a single physics-based loss \cite{jeong_complete_2023,jin_compliant_2025,yousefpour_simultaneous_2025,sun_smo_2025}. Across all three strategies, stress-constrained problems have received little attention, and most studies are limited to 2D examples with a modest number of local constraints.

Most ML-based TO works adopt a NAND approach where the state, adjoint, and design variables are updated in separate steps. An alternative to NAND is the simultaneous analysis and design (SAND) formulation \cite{sankara_truss_1994} which avoids repeated state solves by jointly solving for the state and design variables. The continuous parameterization of strategy (3) is naturally suited to SAND: the equilibrium condition and the design objective are minimized simultaneously by a gradient-based optimizer, and the sensitivities follow from automatic differentiation (AD). 
\ifanon Prior work\else Our prior work\fi\ \cite{yousefpour_simultaneous_2025, yousefpour_localized_2025, sun_smo_2025, sun_multi_2026, hosseinmardi_importance_2026} showed that this simultaneous formulation produces mesh-independent designs with super-resolution for fluid flow optimization, compliance minimization, multi-material compliant mechanism design, and multi-material multi-physics problems. Three properties of the SAND formulation are relevant to stress-constrained TO: (1) the fields are continuous functions of the spatial coordinates, so the resolution at which the constraints are evaluated can be chosen independently of the number of trainable parameters; (2) the state and adjoint fields are obtained by minimizing energy functionals, so no global stiffness matrix is assembled, stored, or factorized; and (3) the loss is a sum over collocation points (CPs), which matches the structure of a penalty over local constraints. However, applying SAND to stress-constrained TO is challenging, since the sensitivity requires an adjoint field that must remain in equilibrium with a state solution that itself changes at every step of the optimization.

Beyond this difficulty, ML-based TO has so far been demonstrated only on relatively small scale problems due to high training costs and peak GPU memory requirements. Evaluating an NN and its spatial derivatives at millions of CPs in both the forward and backward passes easily exceeds the memory of a single graphics processing unit (GPU). This limitation can be addressed by domain decomposition \cite{RN1524, RN1637} in physics-informed neural networks (PINNs) and operator learning. For example, each subdomain can have its own NN \cite{jagtap_conservative_2020, jagtap_extended_2020} that is trained only on the CPs within that subdomain so the memory required by any one network is a fraction of the total. These formulations can be implemented in parallel \cite{shukla_parallel_2021} by placing each subdomain network and its CPs on a separate GPU and exchanging only the interface quantities between devices. The convergence of such subdomain networks can be improved using shared initialization and imposing continuity of the solution and its derivatives at the interfaces \cite{si_initialization_2025}. Overlapping decompositions such as finite basis PINNs \cite{moseley_finite_2023} instead multiply each local network by a partition-of-unity window function and sum the results, so that continuity across the overlaps follows from the smoothness of the windows rather than from an interface loss. 
In all of these approaches, the subdomain solutions are coupled either through interface loss terms that penalize jumps in the solution and its fluxes, or through overlapping windows. The interface terms introduce weights that must be tuned, compete with the physics loss during training, and require interface quantities to be exchanged between devices at every iteration. For a TO problem, the difficulty is exacerbated since the state, adjoint, and density fields would each require its own interface condition, and any residual jump in the displacement across an interface produces a spurious stress, which decreases the accuracy of local stress estimates.

\begin{figure*}[!t]
    \centering
    \begin{subfigure}[t]{\textwidth}
        \captionsetup{justification=raggedright, singlelinecheck=false, skip=0pt, position=top,labelfont=small, textfont=small}
        \caption{Simultaneous optimization framework}
        \includegraphics[width=\linewidth]{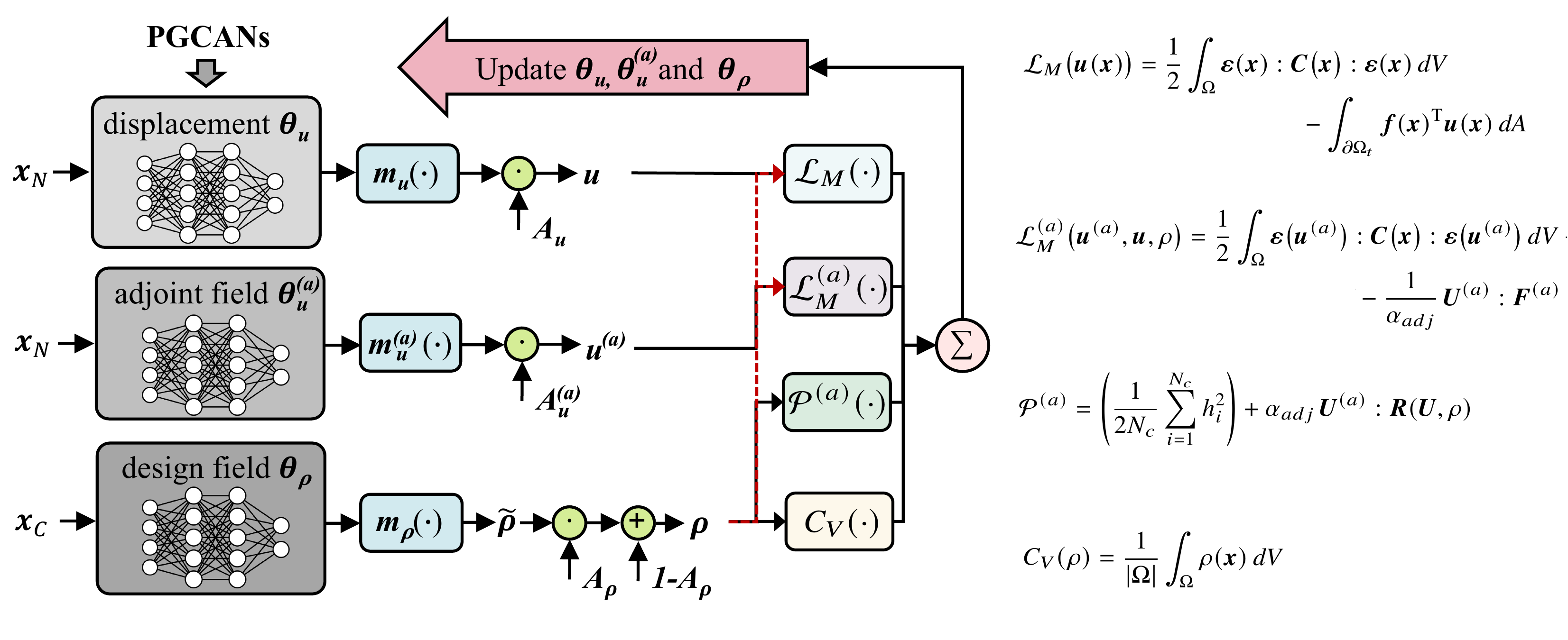}
        \label{fig schem 1}
    \end{subfigure}    
    \par\vspace{-2em}   % gap between (a) and (b): more negative = closer
    \begin{subfigure}[t]{\textwidth}
        \captionsetup{justification=raggedright, singlelinecheck=false, skip=0pt, position=top,labelfont=small, textfont=small}
        \caption{PGCAN architecture}
        \includegraphics[width=\linewidth]{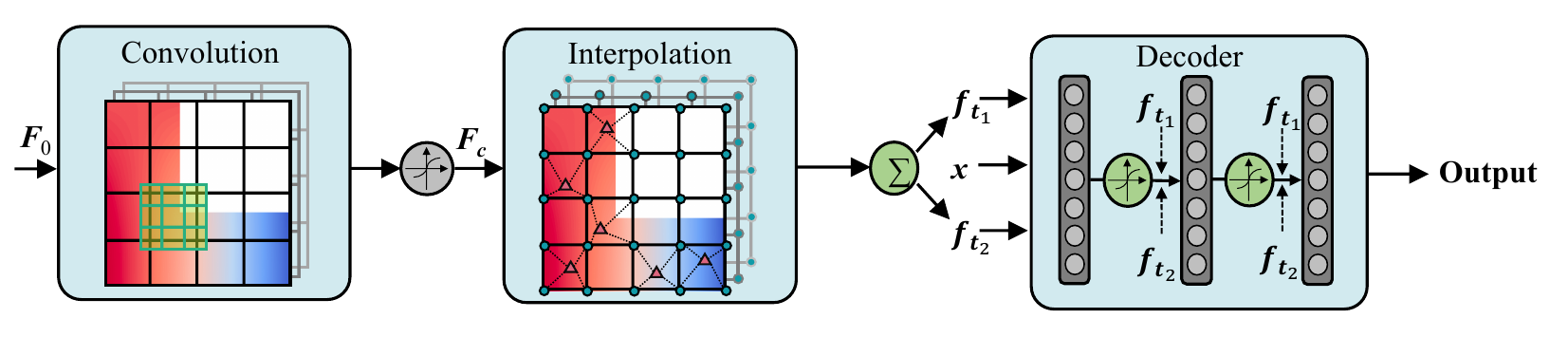}
        \label{fig schem 2}
    \end{subfigure}
    \vspace{-1em}
    \caption{\textbf{SNTO framework for stress-constrained TO:}
    \textbf{(a)} Simultaneous optimization pipeline. Three independent PGCANs generate the raw outputs $\mb_{\ub}(\xb;\thetab_u)$, $\mb_{\ub}^{(a)}(\xb;\thetab_u^{(a)})$, and $m_\rho(\xb;\thetab_\rho)$, which are analytically transformed via $A_{\ub}$, $A_{\ub}^{(a)}$, and $A_\rho$ into displacement, adjoint, and density fields that satisfy their respective constraints by construction. The potential energies $\mathcal{L}_M(\cdot)$ and $\mathcal{L}_M^{(a)}(\cdot)$, the stress constraint penalty $\mathcal{P}^{(a)}(\cdot)$, and the objective of volume fraction $C_V(\cdot)$ are combined into a single loss that is minimized simultaneously with respect to $\thetab_u$, $\thetab_u^{(a)}$, and $\thetab_\rho$. The individual loss terms are listed on the right.
    \textbf{(b)} PGCAN architecture. The trainable feature tensor $\Fb_0$ is convolved to obtain $\Fb_c$, which is interpolated at each query point $\xb$; the interpolated features are then passed to a shallow gated decoder.}
    \label{fig schem}
\end{figure*}

In this work we introduce \textit{simultaneous neural topology optimization} (SNTO) which is an ML-based framework for stress-constrained TO, see \Cref{fig schem}. SNTO represents the state, adjoint, and design fields by three separate NNs with each having as many outputs as the dimension of the variable it parameterizes, see \CrefSub{fig schem}{fig schem 1}. The three NNs are trained on a composite physics-informed loss that combines the objective, the design constraints, and the potential energy functionals. The local stress constraints are collected in a single quadratic penalty summed over all CPs, and its sensitivity is obtained by solving an adjoint problem driven by the active stress constraints. Importantly, our NNs are designed based on the parametric grid convolutional attention network (PGCAN) \cite{shishehbor_parametric_2024,yousefpour_localized_idetc} which mitigates spectral bias and captures the sharp gradients that characterize stress concentrations, see  \CrefSub{fig schem}{fig schem 2}. 
Additionally, differentiations and integrations in our composite loss are approximated with FE shape functions and reduced Gauss quadrature to increase computational efficiency \cite{sun_multi_2026}. This approximation applies to loss evaluation only and so the formulation remains mesh-free from the perspective of the design and state representations.

%========================================
%=======================================

To achieve scalability, SNTO exploits the additive property of the loss rather than domain decomposition. Since every term of our loss is a sum over the CPs,  the gradient of the loss with respect to the NN parameters is additive over any partition of the CPs. We argue that interface loss terms are unnecessary and the continuity across any partition interface is guaranteed by the NN parameterization over the entire domain. In the implementation, the design domain is partitioned into slices, each with its own group of CPs distributed across multiple GPUs. To reduce the peak GPU memory, each slice is further divided into sub-slices that are evaluated sequentially on the host GPU. The composite loss is thus constructed from one small batch of CPs at a time. The gradients are accumulated over the sub-slices and synchronized across the GPUs before updating the shared parameters. 

We summarize the contributions of this work as follows:
\begin{itemize}
    \item We formulate stress-constrained TO as a physics-informed neural framework in which the state, adjoint, and design fields are parameterized by separate NNs and trained simultaneously on a composite loss.
    \item We derive a discrete adjoint sensitivity formulation for the penalty over the local stress constraints, and solve the adjoint field by minimizing its potential energy functional.
    \item We exploit the additivity of the loss gradient to distribute the training of large-scale problems over sub-slices and multiple GPUs without any interface loss terms.
    \item We benchmark SNTO against PolyStress \cite{giraldo_polystress_2021} which shows that our SNTO framework produces sharper and lighter designs at a comparable stress level.
    % \item \fix{Add one more item about comparison to PolyStress}
    \item We demonstrate the scalability of SNTO on 3D stress-constrained designs with up to 8M local stress constraints using up to eight GPUs.
\end{itemize}

The remainder of this paper is organized as follows. \Cref{sec equ} states the governing equations and \Cref{sec metd} presents our approach including its stress-constrained formulation, neural parameterization, multi-GPU training scheme, numerical approximation with shape functions, adjoint sensitivity analysis, and curriculum training strategy. \Cref{sec res} compares SNTO with PolyStress on various 2D benchmarks including the sensitivity to the number of CPs and an FE verification of the stress constraints. Scalability and computational cost of SNTO on 3D problems with up to 8M CPs are also demonstrated. \Cref{sec con} concludes the paper and outlines directions for future work.

    \section{Governing Physics}\label{sec equ}
We consider a linear elastic solid that occupies the domain $\Omega \subset \mathbb{R}^{n_d}$ with boundary $\partial\Omega$ with $n_d \in \{2,3\}$ denoting the spatial dimension. The coordinates of a material point are denoted by $\xb = [x_1,\dots,x_{n_d}]^{\text{T}}$ and the state variable is the displacement field $\ub(\xb) = [u_1(\xb),\dots,u_{n_d}(\xb)]^{\text{T}}$. Under the infinitesimal strain assumption, the strain tensor is symmetric and can be written as:
%=======================================
\begin{equation}\label{eq strain}
\epsilonb(\xb)
=
\frac{1}{2}
\big(
\nabla \ub(\xb)
+
\nabla \ub^{\text{T}}(\xb)
\big).
\end{equation}
%=======================================
The stress tensor follows from the linear elastic constitutive relation:
%=======================================
\begin{equation}\label{eq const}
    \sigmab(\xb) = \Cb(\xb) : \epsilonb(\xb),
\end{equation}
%=======================================
where $\Cb(\xb)$ is the fourth-order stiffness tensor which is constructed by the Young's modulus $E(\xb)$ and the Poisson's ratio $\nu$ for isotropic materials. We write $\Cb(\xb)$ as a function of position because the Young's modulus is interpolated from the density field and therefore varies over the domain. Ignoring the body forces, the static equilibrium condition is given as:
%=======================================
\begin{equation}\label{eq equil}
    \nabla \cdot \sigmab(\xb) = \mathbf{0}, \quad \forall \xb \in \Omega,
\end{equation}
%=======================================
where $\nabla\cdot(\cdot)$ denotes the divergence operator. The domain boundary is partitioned into three parts, $\partial\Omega = \partial\Omega_h \cup \partial\Omega_t \cup \partial\Omega_0$ with the prescribed boundary conditions (BCs):
%=======================================
\begin{subequations} \label{eq bc}
    \begin{align}
    &\ub(\xb) = \tilde{\ub}(\xb), \quad \forall \xb \in \partial \Omega_{h}, \label{eq bc 1} \\
    &\sigmab(\xb) \cdot \nb(\xb) = \fb(\xb), \quad \forall \xb \in \partial \Omega_t, \label{eq bc 2}\\
    &\sigmab(\xb) \cdot \nb(\xb) = \mathbf{0}, \quad \forall \xb \in \partial \Omega_0, \label{eq bc 3}
    \end{align}
\end{subequations}
%=======================================
where $\partial \Omega_h$ and $\partial \Omega_t$ denote the regions with prescribed displacement and traction while $\partial \Omega_0$ indicates free surfaces. The prescribed displacement is given as $\tilde{\ub}(\xb) := [\tilde{u}_1(\xb),\dots,\tilde{u}_{n_d}(\xb)]^{\text{T}}$ and the applied traction is $\fb(\xb) := [f_1(\xb),\dots,f_{n_d}(\xb)]^{\text{T}}$ with $\nb(\xb) := [n_1(\xb),\dots,n_{n_d}(\xb)]^{\text{T}}$ being the outward unit normal vector on $\partial\Omega_t$. \Cref{eq strain} to \Cref{eq bc} define the boundary value problem whose solution is the displacement field for a given material layout.

Instead of directly solving the strong form in \Cref{eq equil}, we rely on its variational form by the principle of minimum potential energy which states that among all kinematically admissible displacement fields (i.e., those that satisfy \Cref{eq bc 1}), the equilibrium solution is the one that minimizes the total potential energy functional:
%=======================================
\begin{equation}\label{eq LM}
\mathcal{L}_M\big(\ub(\xb)\big) =
\frac{1}{2}\int_{\Omega}
\epsilonb(\xb):
\Cb\big(\xb\big):
\epsilonb(\xb)\, dV
-\int_{\partial\Omega_t}
\fb(\xb)^{\text{T}}
\ub(\xb)\, dA,
\end{equation}
%=======================================
where the first term on the right-hand side is the strain energy stored in the solid and the second term is the work done by the external tractions. The stationarity conditions of $\mathcal{L}_M(\cdot)$ recover the equilibrium equation in \Cref{eq equil} together with the traction BCs in \Cref{eq bc 2}. The free-surface conditions in \Cref{eq bc 3} appear as natural BCs with zero energy and need not be imposed on the solution space. 

The energy formulation in \Cref{eq LM} is the basis of DEM \cite{nguyen_DEM_2020,haghighat_pinn_2021} which is preferable to the residual loss based on \Cref{eq equil} for three reasons: (1) $\mathcal{L}_M(\cdot)$ involves only first derivatives of $\ub(\xb)$, which can be evaluated using shape functions as in the finite element method (FEM) \cite{hughes_FEM_2012}; (2) $\mathcal{L}_M(\cdot)$ is quadratic for linear elasticity, which makes its minimization with gradient-based optimizers well behaved; and (3) the minimizer satisfies a simple energy balance at equilibrium where twice the strain energy equals the external work. We use this identity for diagnosing how closely the trained NNs satisfy equilibrium during TO iterations.
    \section{Proposed Method}\label{sec metd}
In stress-constrained TO, we seek to minimize the volume fraction subject to the equilibrium condition and the local stress constraints:
%=======================================
\begin{subequations}\label{eq sc1}
\begin{align}
\widehat{\rho}(\xb) &=
\argmin_{\rho}
C_V(\rho)
= \argmin_{\rho}\frac{1}{|\Omega|}\int_{\Omega}\rho(\xb)\, dV,
\label{eq sc1 a}
\intertext{subject to:}
\ub(\xb;\rho) &=
\argmin_{\ub}
\mathcal{L}_M\big(\ub(\xb;\rho),\rho(\xb)\big)
= \argmin_{\ub}\bigg(
\frac{1}{2}\int_{\Omega}
\epsilonb(\xb):
\Cb\big(\xb\big):
\epsilonb(\xb)\, dV
% \\
% &\qquad
-\int_{\partial\Omega_t}
\fb(\xb)^{\text{T}}
\ub\, dA
\bigg),
\label{eq sc1 b}
\\
\frac{\sigma_{\mathrm{VM}}(\xb_i)}{\sigma_y} & - 1 \le 0,
\quad \xb_i\in\Omega, \quad i = 1,\dots, N_c,
\label{eq sc1 c}
\end{align}
\end{subequations}
%=======================================
where $\rho(\xb)\in[0,1]$ is the relative density at point $\xb$, $C_V(\rho)$ is the volume fraction of the design, $|\Omega|$ is the volume of the design domain, $\sigma_{\mathrm{VM}}(\xb)$ is the equivalent (von Mises) stress, $\sigma_y$ is the effective stress limit, and $N_c$ is the number of local stress constraints. \Cref{eq sc1 b} states that the displacement field must be in equilibrium with the current design; the displacement $\ub(\xb;\rho)$ therefore depends on $\rho(\xb)$. 

To obtain $\Cb(\xb)$ in \Cref{eq sc1 b} for a design field involving gray areas, we follow SIMP interpolation and penalize the Young's modulus. That is:
%=======================================
\begin{equation}\label{eq E}
E\big(\xb\big) = m_E(\xb)E_0,
\end{equation}
%=======================================
where $E_0$ is the Young's modulus of the solid material and $m_E(\xb)$ is the SIMP interpolation function given as:
%=======================================
\begin{equation}\label{eq me}
m_E(\xb)=\epsilon + (1-\epsilon)\rho(\xb)^{\,p}.
\end{equation}
%=======================================
Here $p$ is the penalization exponent and $\epsilon = 10^{-6}$ is a small constant that prevents numerical instability. 

The equivalent stress in \Cref{eq sc1 c} is the von Mises stress defined as:
%=======================================
\begin{equation}\label{eq svm}
\sigma_{\mathrm{VM}}(\xb)
=
\sqrt{\frac{3}{2}\,\operatorname{dev}\big(\sigmab(\xb)\big):\operatorname{dev}\big(\sigmab(\xb)\big)}, 
\quad \operatorname{dev}\sigmab(\xb) = \sigmab(\xb) - \frac{1}{3}\operatorname{tr}\big(\sigmab(\xb)\big)\Ib,
\end{equation}
%=======================================
where $\operatorname{tr}(\cdot)$ is the trace operator and $\Ib$ the second-order identity tensor.
In a 3D Cartesian coordinate system we have:
%=======================================
\begin{equation}\label{eq svm 3d}
\sigma_{\mathrm{VM}}(\xb)
=
\sqrt{
\frac{1}{2}\Big[
\big(\sigma_{xx}-\sigma_{yy}\big)^2
+\big(\sigma_{yy}-\sigma_{zz}\big)^2
+\big(\sigma_{zz}-\sigma_{xx}\big)^2
\Big]
+3\big(\tau_{xy}^2+\tau_{yz}^2+\tau_{zx}^2\big)
},
\end{equation}
%=======================================
where $\sigma_{xx}$, $\sigma_{yy}$, and $\sigma_{zz}$ are the normal stresses and $\tau_{xy}$, $\tau_{yz}$, and $\tau_{zx}$ are the shear stresses. The dependence of stress components on $\xb$ is omitted for brevity. The plane stress form used in our 2D examples follows from \Cref{eq svm 3d} by setting $\sigma_{zz}=\tau_{yz}=\tau_{zx}=0$.

Since $\sigmab(\xb)$ is obtained from the interpolated stiffness $\Cb(\xb)$, the equivalent stress in \Cref{eq svm} vanishes as the density approaches zero. If this quantity is constrained directly, a low-density region would carry arbitrarily large strains without violating the constraint, allowing the optimizer to satisfy the stress limit by lowering the density rather than by reorganizing the material. 
To address this issue, we follow \textcite{le_stress_2010,senhora_topology_2020} and evaluate the stresses with the stiffness tensor of the solid material, i.e., with $\rho(\xb)=1$, and denote the resulting equivalent stress by $\sigma^s_{eq}(\xb)$. This is the stress that the material at $\xb$ would carry if it were fully solid under the current strain field. The normalized stress violation measure $s(\xb)$ is then defined as:%=======================================
\begin{equation}\label{eq s}
    s(\xb) = \frac{\sigma^s_{eq}(\xb)}{\sigma_y} - 1,
\end{equation}
%=======================================
such that $s(\xb) \le 0$ corresponds to a satisfied constraint. Evaluating the equivalent stress in \Cref{eq s} with the solid stiffness introduces the singularity phenomenon \cite{duysinx_topology_1998}. Since $\sigma^s_{eq}(\xb)$ does not vanish as $\rho\to0$, the constraint remains active in regions that are being removed. The constraint must therefore be relaxed so that it vanishes in void regions where $s(\xb)$ has no physical meaning. To achieve this, we use a polynomial vanishing constraint of the form:
%=======================================
\begin{equation}\label{eq g}
    g(\xb) = m_E(\xb)\,s(\xb)\big(s(\xb) + 1\big)^2,
\end{equation}
%=======================================
where $m_E(\xb)$ is the interpolation function defined in \Cref{eq me}. Multiplication by $m_E(\xb)$ relaxes the stress constraint in low-density regions: as $\rho(\xb)\to 0$, $m_E(\xb)\to\epsilon$, and the contribution of the constraint becomes negligible. This relaxation allows gradient-based optimizers to remove material and evolve the design through void regions \cite{cheng_epsilon_1997,bruggi_alternative_2008}. Using the same interpolation function in \Cref{eq g,eq E} ensures that the stiffness and stress-constraint contributions decay consistently as the density decreases.

The formulation in \Cref{eq g} is a variant of the one used in PolyStress: $g_{\text{PolyStress}}(\xb) = m_E(\xb)\,s(\xb)\big(s(\xb)^2+1\big)$ \cite{senhora_topology_2020,giraldo_polystress_2021}. The two agree for a marginally active constraint as $s \to 0$. However, for regions where $s>0$ our form is larger by $2m_E s^2$ and grows more steeply with the violation. This steeper growth penalizes severe stress violations more aggressively, which is particularly beneficial in a penalty-based framework. Since only the constraints with $s>0$ should influence the optimization, we keep the positive part of $g(\xb)$:
%=======================================
\begin{equation}\label{eq h}
    h(\xb) = \max\big(g(\xb),0\big),
\end{equation}
%=======================================
and collect all local constraints in \Cref{eq sc1 c} into a single quadratic penalty functional:
%=======================================
\begin{equation}\label{eq P}
    \mathcal{P}(g(\xb)) = \frac{1}{2N_c} \sum_{i=1}^{N_c} h(\xb_i)^2, \quad \xb_i\in\Omega, \quad i = 1,\dots, N_c.
\end{equation}
%=======================================
The factor $1/2$ cancels the factor $2$ from differentiating $h^2$ and simplifies the sensitivity expression.
In \Cref{eq P}, every point that violates the yield criterion enters the penalty with a weight proportional to its squared violation. The normalization by $N_c$ keeps the magnitude of $\mathcal{P}(\cdot)$ independent of the number of CPs, enabling the same penalty weight to be used across different examples. 
$\mathcal{P}(\cdot)$ depends on both the displacement and the density fields. Its gradient with respect to the density field has an explicit part obtained by AD through $m_E(\rho)$ and an implicit part of $\partial \ub/\partial \rho$ which we estimate using the adjoint analysis in \Cref{sec metd adj}. 

The density $\rho(\xb)$ that appears in \Cref{eq sc1,eq me,eq E} is the final physical density. In our framework, the network output $m_\rho(\xb)$ is passed through a smooth Heaviside projection to promote a nearly binary design \cite{wang_projection_2011}:
%=======================================
\begin{equation}\label{eq proj}
\tilde{\rho}(\xb)
=
\frac{
\tanh(\beta\eta)
+
\tanh\big(\beta(m_\rho(\xb)-\eta)\big)
}{
\tanh(\beta\eta)
+
\tanh\big(\beta(1-\eta)\big)
},
\end{equation}
%=======================================
where $\tilde{\rho}(\xb)$ is the projected design and $\eta = 0.5$ is the threshold density. The parameter $\beta$ controls the sharpness of the transition between void and solid. The projection pushes intermediate densities toward $0$ or $1$ which reduces gray regions in the final design.
The physical density $\rho(\xb)$ coincides with $\tilde{\rho}(\xb)$ everywhere except in the regions where material is \textit{prescribed} by the analytical function described in \Cref{sec metd param}.

\subsection{Neural Parameterization}\label{sec metd param}
All of our NNs are based on the PGCAN architecture \cite{shishehbor_parametric_2024} whose underlying idea is to (1) encode a query point with trainable features that depend on its location in the domain, and (2) decode those features with a shallow network to obtain the desired outputs. Standard multilayer perceptrons (MLPs) are known to preferentially learn low-frequency content of the solution and to struggle with the sharp gradients that characterize the stress concentration field near re-entrant corners or crack tips. PGCAN alleviates this spectral bias by first arranging the trainable parameters on a spatial feature grid to localize the learning capacity and then letting its shallow MLP resolve fine-scale features.
As shown in \Cref{sec res}, PGCAN is directly applicable to complex design domains in 2D and 3D. Below, we summarize its implementation for a simple 2D problem. 

As shown in \CrefSub{fig schem}{fig schem 2} for a rectangular domain with a hole inside it, we overlay the domain with $N_{\mathrm{rep}}$ grids where each one spans the entire domain and is discretized into $N_x^e \times N_y^e$ cells. The grids are shifted relative to one another by small offsets along the diagonal direction. A trainable feature vector of length $N_f$ is assigned to each grid vertex and the resulting feature vectors collectively form the initial feature tensor:
%=======================================
\begin{equation}\label{eq fb0}
    \Fb_0 \in
    \mathbb{R}^{
        N_{\mathrm{rep}} \times N_f
        \times (N_x^e+1)
        \times (N_y^e+1)
    }.
\end{equation}
%=======================================
To incorporate information from neighboring vertices, $\Fb_0$ is processed using a depthwise convolution with a $3 \times 3$ kernel followed by a $\tanh$ activation, which yields the final feature tensor $\Fb_c$. Because the convolution is applied independently to each feature channel, it captures local spatial correlations without mixing information across channels.
For a query point $\xb$, we first identify the cell containing the point in each grid and express its position using the normalized local coordinates $\bar{\xb}\in[0,1]^{n_d}$. These coordinates are smoothed using the cosine mapping:
%=======================================
\begin{equation}\label{eq xc}
    \xb^{(c)}
    =
    \frac{1}{2}
    \left[
        1-\cos\left(\pi\bar{\xb}\right)
    \right].
\end{equation}
%=======================================
The features at the four vertices of the enclosing cell are then bilinearly interpolated using $\xb^{(c)}$. The interpolated features from all $N_{\mathrm{rep}}$ grids are summed to obtain the pointwise feature vector $\fb_{\tb}(\xb)\in\mathbb{R}^{N_f}$.
Finally, a shallow MLP decodes the spatial coordinate $\xb$ and the interpolated feature vector into the desired field value. At each hidden layer, a sigmoid gate computed from the current hidden state adaptively blends the two halves of the feature vector, $\fb_{\tb_1}(\xb)$ and $\fb_{\tb_2}(\xb)$. This gated architecture facilitates gradient propagation from the output to the grid features \cite{wang2021understanding}. In all numerical examples, we use $N_{\mathrm{rep}}=3$, $N_f=128$, and three hidden layers of width $N_f/2$ with $\tanh$ activations, where the hidden-layer width is chosen to match the dimension of each half of the feature vector. More details about the PGCAN can be found in \textcite{shishehbor_parametric_2024}.

For SNTO, we parameterize the displacement, adjoint, and density fields using three independent PGCANs with trainable parameters $\thetab_u$, $\thetab_u^{(a)}$, and $\thetab_\rho$, respectively, see \CrefSub{fig schem}{fig schem 1}. Each network maps a spatial coordinate to a raw output: $\mb_{\ub}(\xb;\thetab_u)\in\mathbb{R}^{n_d}$ for the displacement field, $\mb_{\ub}^{(a)}(\xb;\thetab_u^{(a)})\in\mathbb{R}^{n_d}$ for the adjoint field, and $m_\rho(\xb;\thetab_\rho)\in\mathbb{R}$ for the density field. Using a separate network for each field reduces interference among the loss terms and enables field-specific learning rates, which we found necessary for stable training.

We are free to evaluate the three networks at arbitrary CPs but choose specific configurations to promote computational efficiency and training stability.
Specifically, we evaluate the three fields at two sets of CPs defined on a common background grid. The nodal CPs denoted as $\xb_N$ coincide with the nodes of an FE mesh overlaid on the domain. $\xb_N$ are used to evaluate $\mb_{\ub}(\xb;\thetab_u)$ and $\mb_{\ub}^{(a)}(\xb;\thetab_u^{(a)})$. Centroid CPs are denoted as $\xb_C$ and located at the element centers and are used to evaluate $m_\rho(\xb;\thetab_\rho)$. Hereafter, the term ``CPs'' refers to either the nodal or centroid CPs, depending on the network being evaluated. The correspondence between the two sets of CPs enables the spatial gradients and integrals in the loss terms to be evaluated using FE shape functions, as detailed in \Cref{sec metd num}. Upon training, the NNs can be queried at arbitrary points in the design domain to evaluate the displacement and density fields.

The raw network outputs do not generally satisfy the BCs in \Cref{eq bc 1} or the prescribed design constraints. Rather than enforcing these conditions through additional penalty terms in the loss function, we impose them analytically through output transformations. For the displacement field, each component $u_i(\xb)$ may be prescribed as $\tilde{u}_i(\xb)$ on a corresponding boundary subset $\partial\Omega_{h,i}\subseteq\partial\Omega_h$. So, we express each displacement component as:
%=======================================
\begin{equation}\label{eq AB}
    u_i(\xb) = A_{u_i}(\xb)\, m_{u_i}(\xb;\thetab_u) + B_{u_i}(\xb),\quad i = 1,\dots,n_d,
\end{equation}
%=======================================
where $m_{u_i}(\xb;\thetab_u)$ is the $i$-th component of the displacement network output. The fixed function $A_{u_i}(\xb)$ vanishes on $\partial\Omega_{h,i}$ and is positive elsewhere with a maximum value of one, while $B_{u_i}(\xb)$ satisfies $B_{u_i}(\xb)=\tilde{u}_i(\xb)$ on $\partial\Omega_{h,i}$. Consequently, the product $A_{u_i}(\xb)m_{u_i}(\xb;\thetab_u)$ vanishes on the prescribed boundary $\partial\Omega_{h,i}$, ensuring that the BC is satisfied exactly regardless of the network parameters. Because $A_{u_i}(\xb)$ and $B_{u_i}(\xb)$ depend only on the domain geometry and the prescribed BCs, they are computed once and cached for all CPs before training. Since all examples considered in this work involve homogeneous displacement BCs, $B_{u_i}(\xb)=0$ and \Cref{eq AB} reduces to:
%=======================================
\begin{equation}\label{eq Au}
    u_i(\xb) = A_{u_i}(\xb)\, m_{u_i}(\xb;\thetab_u), \quad i = 1,\dots,n_d.
\end{equation}
%=======================================
Inspired by \textcite{mora_gaussian_2024}, we construct $A_{u_i}(\xb)$ using the minimum Euclidean distance from $\xb$ to the set $\Xb_{h,i}$ of CPs located on $\partial\Omega_{h,i}$:
%=======================================
\begin{subequations}\label{eq A dist}
\begin{align}
    d_i(\xb) &= \min_{\xb'\in\Xb_{h,i}} \big\|\xb - \xb'\big\|_2, \qquad
    \hat{d}_i(\xb) = \frac{d_i(\xb)}{\max_{\xb\in\Omega} d_i(\xb)}, \label{eq A dist a}\\
    A_{u_i}(\xb) &= 1 - \exp\big(-K_u\,\hat{d}_i(\xb)\big), \label{eq A dist b}
\end{align}
\end{subequations}
%=======================================
where $\hat{d}_i(\xb)\in[0,1]$ is the normalized distance and $K_u$ controls the rate at which $A_{u_i}(\xb)$ increases from zero on the constrained boundary toward unity in the interior. A larger value of $K_u$ confines the attenuation of the network output to a thinner region near $\partial\Omega_{h,i}$. We use $K_u=50$ throughout this work. Because this construction depends only on distances to a set of boundary points, it can be readily applied to the irregular domain geometries considered in \Cref{sec res}. The adjoint field $\ub^{(a)}(\xb)$ satisfies the same homogeneous BCs as the primal displacement field and so it uses the same weight functions, $u_i^{(a)}(\xb)=A_{u_i}(\xb)\,m_i^{(a)}(\xb;\thetab_u^{(a)})$, such that $A_{\ub}^{(a)}=A_{\ub}$ in \CrefSub{fig schem}{fig schem 1}.

We apply a similar transformation to the density field to enforce $\rho(\xb)=1$ on the loading boundary $\partial\Omega_t$ and in any regions containing prescribed solid features:
%=======================================
\begin{equation}\label{eq Arho}
    \rho(\xb)=A_\rho(\xb)\,\tilde{\rho}(\xb)+\bigl(1-A_\rho(\xb)\bigr),
\end{equation}
%=======================================
where $\tilde{\rho}(\xb)$ is the projected network output. The weight $A_\rho(\xb)$ is constructed using \Cref{eq A dist}, with the distance measured from the set of CPs on $\partial\Omega_t$ and $K_\rho=500$. \Cref{eq Arho} is a special case of \Cref{eq AB} with $B_\rho(\xb)=1-A_\rho(\xb)$. It enforces $\rho(\xb)=1$ on $\partial\Omega_t$, smoothly approaches $\tilde{\rho}(\xb)$ away from the boundary, and remains differentiable with respect to $\thetab_\rho$. The larger value of $K_\rho$ confines the prescribed solid region to a thin band near $\partial\Omega_t$. 

\subsection{Memory-Efficient Multi-GPU Training}\label{sec metd gpu}

%========================================
\begin{figure*}[!t]
    \centering
    \begin{subfigure}[t]{0.3\textwidth}
        \captionsetup{justification=raggedright, singlelinecheck=false, skip=0pt, position=top,labelfont=subtxt, textfont=subtxt}
        \caption{Single GPU}
        \includegraphics[width=\linewidth]{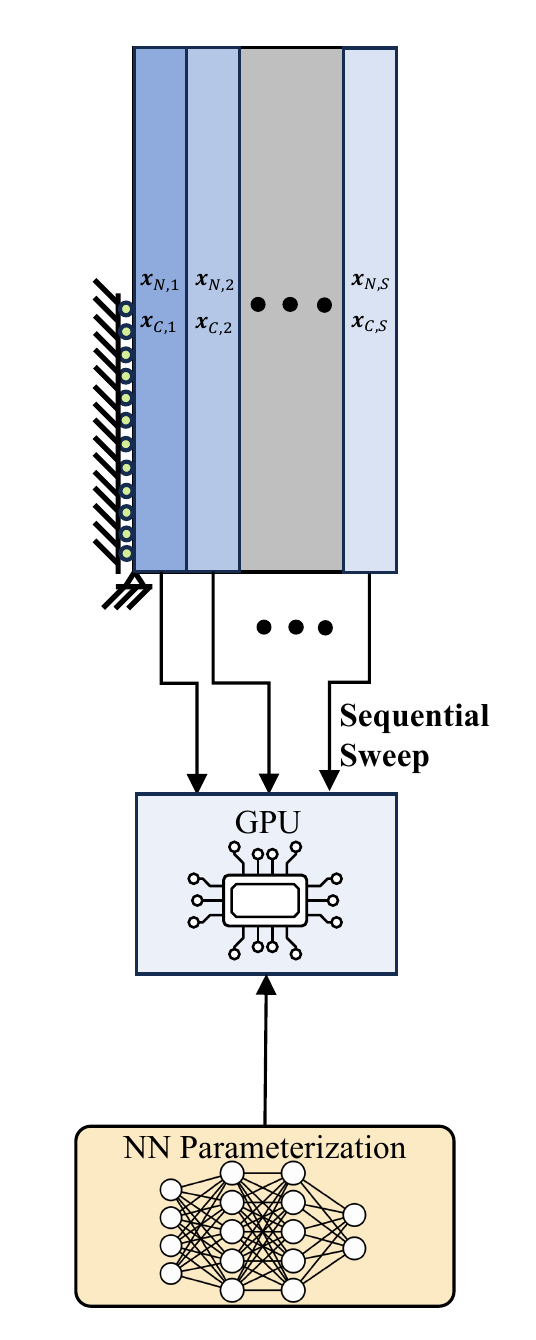}
        \label{fig gpu 1}
    \end{subfigure}
    % \hfill
    \begin{subfigure}[t]{0.495\textwidth}
        \captionsetup{justification=raggedright, singlelinecheck=false, skip=0pt, position=top,labelfont=subtxt, textfont=subtxt}
        \caption{Multiple GPUs}
        \includegraphics[width=\linewidth]{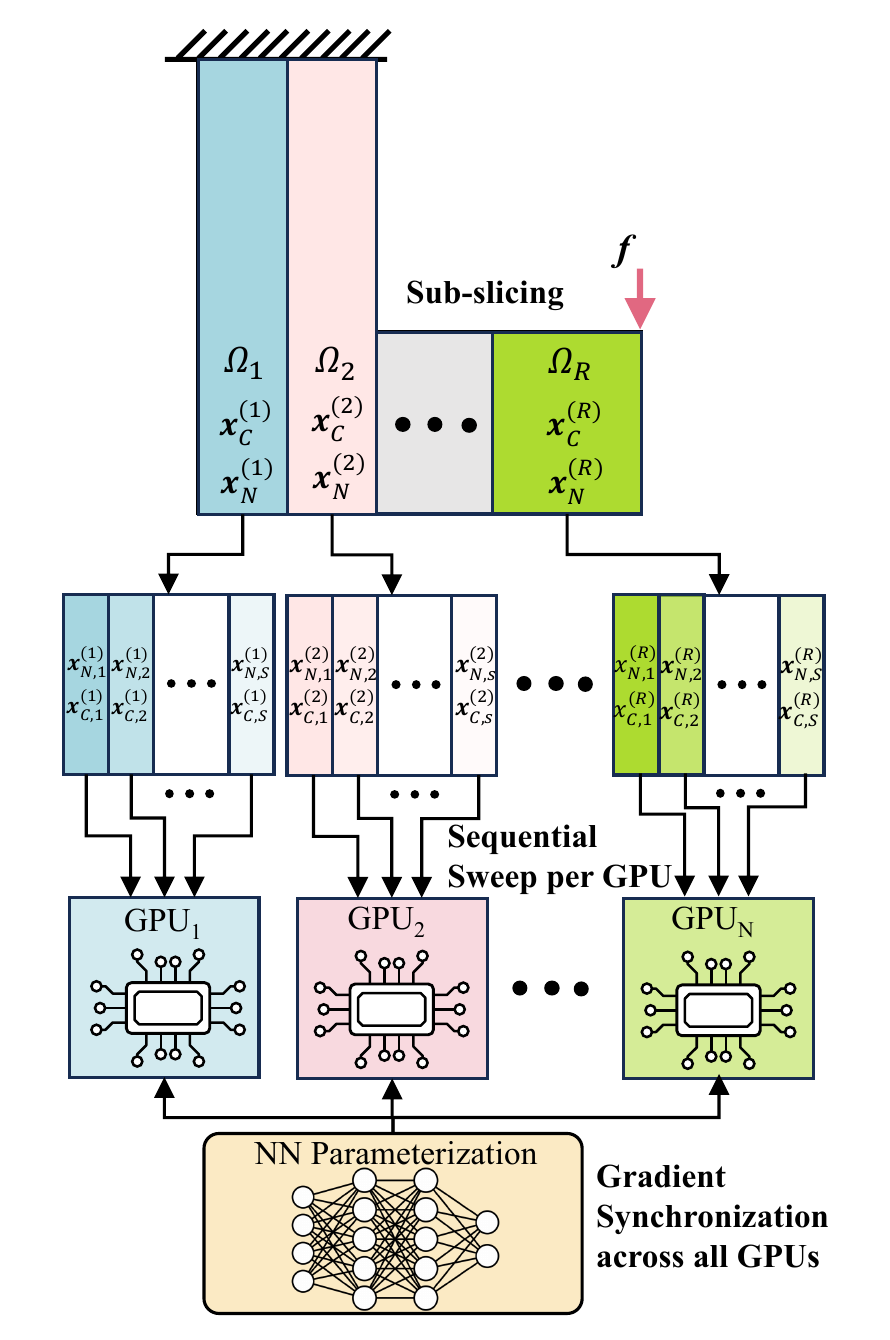}
        \label{fig gpu 2}
    \end{subfigure}
    \vspace{-1em}
    \caption{\textbf{Memory-efficient training scheme for large-scale stress-constrained TO:}
    \textbf{(a)} Single-GPU setup, in which the centroid CPs $\xb_C$ and the associated nodal CPs $\xb_N$ of the design domain are partitioned into $S$ sub-slices that are passed to the GPU sequentially;
    \textbf{(b)} multi-GPU setup, in which the design domain is sliced into $R$ subdomains $\Omega_1,\dots,\Omega_R$, one per GPU, and each subdomain is further divided into $S$ sub-slices that are swept sequentially on its GPU, so that the autograd graph never holds more than one sub-slice. The parameter gradients are accumulated over the sub-slices and synchronized across the GPUs before every update of the shared NNs.}
    \label{fig gpu}
\end{figure*}
%=======================================

Large-scale TO remains particularly challenging for ML-based approaches because GPU memory consumption is governed primarily by two factors: (1) the number of NN parameters, and (2) the number of CPs at which the loss terms and their gradients are evaluated through AD. As described in \Cref{sec metd param}, we use three separate NNs to represent the displacement, adjoint, and density fields, which are evaluated over two sets of CPs. For large-scale problems with millions of CPs in each set, constructing a single autograd graph for the loss globally across the entire domain can readily exceed the memory capacity of a single commercial GPU.

To reduce peak memory, we develop a memory-efficient training strategy that exploits the additive structure of our loss terms and can be used with either one or multiple GPUs, see \Cref{fig gpu}. Because the objective, constraint, and adjoint quantities are assembled as sums over the nodal or centroid CPs, their gradients with respect to the NN parameters can be computed for separate subsets of CPs and accumulated to recover the full-domain gradients.

In the single-GPU setting exemplified in \CrefSub{fig gpu}{fig gpu 1}, the complete set of centroid CPs $\xb_C$ is sorted according to their $x$-coordinate and partitioned into $S$ sub-slices containing equal numbers of centroid CPs, denoted by $\xb_{C,1},\ldots,\xb_{C,S}$ (for simplifying the implementation, we group the centroid CPs along the $x$-direction in this work but the partitioning scheme can be chosen arbitrarily). The nodal CPs associated with the elements in each centroid sub-slice are then identified, yielding the corresponding sets $\xb_{N,1},\ldots,\xb_{N,S}$. This partitioning is performed once before training, and all CP subsets are stored on the GPU for repeated use throughout the optimization.
The sub-slices are subsequently processed in sequence on the same GPU. For each sub-slice, the three NNs are evaluated either at the corresponding centroid or the nodal CPs to obtain the required state and design variables. The contribution of that sub-slice to the loss is then computed, and backpropagation is performed immediately. The resulting parameter gradients are accumulated, while the autograd graph associated with the current sub-slice is released before the next one is processed. As a result, although all CP subsets remain stored on the GPU, only one sub-slice is included in the active autograd graph at a time. The peak memory footprint is therefore reduced and governed by the size of a single sub-slice rather than that of the entire design domain.

Although sub-slicing reduces peak GPU memory usage, processing them sequentially can lead to long training times for large-scale problems using a single GPU. To improve computational efficiency, we extend the single-GPU strategy to multiple GPUs as shown in \CrefSub{fig gpu}{fig gpu 2}. The hierarchical partitioning consists of two stages: (1) the design domain is partitioned into $R$ disjoint subdomains, $\Omega_1,\ldots,\Omega_R$, that are distributed across $R$ GPUs; and (2) each subdomain is further divided into $S$ local sub-slices that are processed sequentially on the assigned GPU following the same procedure as in the single-GPU setting.

In the first stage, the centroid CPs are sorted by their $x$-coordinates and divided into $R$ contiguous groups containing approximately equal numbers of CPs, denoted by $\xb_C^{(1)},\ldots,\xb_C^{(R)}$. This partitioning balances the computational workload and memory usage across the GPUs. The corresponding sets of nodal CPs, $\xb_N^{(1)},\ldots,\xb_N^{(R)}$, are then constructed for the respective subdomains. In the second stage, the centroid CPs within each subdomain $\xb_C^{(i)}$ are further partitioned into $S$ local sub-slices, $\xb_{C,1}^{(i)},\ldots,\xb_{C,S}^{(i)}$, along with their corresponding nodal CP sets, $\xb_{N,1}^{(i)},\ldots,\xb_{N,S}^{(i)}$.

During training, each GPU stores a replica of the three NNs and the two sets of CPs associated with its assigned subdomain. The GPUs operate in parallel while the local sub-slices on each GPU are processed sequentially. For each local sub-slice, the loss contribution is evaluated and backpropagated, and the resulting parameter gradients are accumulated. Only the autograd graph associated with the current sub-slice is retained, thereby limiting the peak memory usage on each GPU.
After all local sub-slices have been processed, the accumulated parameter gradients are aggregated across the GPUs through an all-reduce operation implemented using the PyTorch Distributed Data Parallel (DDP) library. Because each GPU maintains an identical replica of the global NNs, applying the synchronized gradients produces identical parameter updates and preserves consistency among the network replicas throughout training. The adjoint nodal loads assembled during each sweep are aggregated across the GPUs in the same manner to obtain the global adjoint load for the subsequent training iteration.

The two levels of partitioning play complementary roles in the multi-GPU setup. Partitioning the domain across GPUs distributes both memory and computation, reducing the wall-clock cost per training iteration and allowing larger problems to be handled by adding devices. Sub-slicing within each GPU trades additional sequential passes for lower peak memory, making the reachable problem size no longer limited by the memory of a single device. In \Cref{sec res}, we independently vary the number of GPUs $R$ and the number of sub-slices $S$ to verify that the optimized topology is insensitive to these computational parameters.

We emphasize that this strategy differs fundamentally from domain-decomposition approaches commonly used in PINNs \cite{moseley_finite_2023, si_initialization_2025}. Such approaches typically assign independent subnetworks to different subdomains and enforce compatibility using interface-loss or flux-matching terms with tunable penalty weights. In contrast, we employ a single global neural representation for each field, with identical network replicas distributed across all GPUs. The field representations are therefore continuous across sub-slice and subdomain boundaries by construction. Moreover, elements adjacent to an interface are evaluated using nodal values generated by the same global networks, ensuring that the partitioning does not introduce inconsistencies in the strain calculations.
Consequently, no interface-loss terms, interface-weight tuning, or communication between subdomains is required during the local forward and backward passes. Communication is needed only to aggregate the parameter gradients and adjoint-load contributions before each global parameter update. The shared nodal CPs along subdomain boundaries introduce only a small amount of redundant computation and negligible algorithmic overhead.

\subsection{Numerical Approximation}\label{sec metd num}

The loss terms in \CrefSub{fig schem}{fig schem 1} involve spatial derivatives of the displacement and adjoint fields, as well as integrals over the design domain. Although these quantities can be evaluated using AD, constructing and retaining the required derivative graphs at millions of CPs would incur prohibitive computational and memory costs. Following \ifanon the approach of \textcite{sun_multi_2026}\else our previous work \cite{sun_multi_2026}\fi, we employ the shape-function machinery of the FEM \cite{hughes_FEM_2012}. The nodal and centroid CPs are chosen to coincide with the nodes and element centers of a fixed quadrilateral (in 2D) or hexahedral (in 3D) background mesh. Their relationship is defined by the standard element connectivity. The displacement and adjoint fields are evaluated at the nodal CPs whereas the density field is evaluated at the centroid CPs. Spatial derivatives are computed using cached shape-function gradient matrices and the domain integrals are evaluated using Gauss quadrature. We emphasize that the background mesh serves only for quadrature and differentiation and the unknowns remain the network parameters, preserving the global continuous representations of all three fields in the design domain.

Within an element $e$ occupying the subdomain $\Omega^{(e)}$ with nodal coordinates $\xb_j^{(e)}$, $j=1,\ldots,n_s$, each displacement component is interpolated as:
%=======================================
\begin{equation}\label{eq sf 1}
    u_i^{(e)}(\xb) = \sum_{j=1}^{n_s}N_j\big(\xib^{(e)}(\xb)\big)\, u_i\big(\xb^{(e)}_j;\thetab_u\big), \quad i=1,\dots,n_d, \quad \xb\in\Omega^{(e)},
\end{equation}
%=======================================
where $n_s$ denotes the number of nodes per element ($n_s=4$ in 2D and $n_s=8$ in 3D), $N_j(\cdot)$ is the shape function associated with node $j$, and $\xib^{(e)}(\xb)=[\xi_1^{(e)},\ldots,\xi_{n_d}^{(e)}]^{\mathrm{T}}\in[-1,1]^{n_d}$ denotes the local coordinates of $\xb$ in the parent element. The nodal values $u_i(\xb_j^{(e)};\thetab_u)$ are obtained by applying the analytical transformation in \Cref{eq Au} to the corresponding network outputs. For a trilinear hexahedral element in 3D the shape function is
%=======================================
\begin{equation}\label{eq sf 2}
    N_j\big(\xib^{(e)}\big) = \frac{1}{8}\big(1+\xi^{(e)}_{1,j}\xi^{(e)}_1\big)\big(1+\xi^{(e)}_{2,j}\xi^{(e)}_2\big)\big(1+\xi^{(e)}_{3,j}\xi^{(e)}_3\big), \quad j = 1,\dots,8,
\end{equation}
%=======================================
where $\xi_{k,j}^{(e)}\in\{-1,1\}$ is the $k$th local coordinate of node $j$ (the bilinear quadrilateral 2D element is defined similarly as the product of two factors). The physical and local coordinates are related through the isoparametric mapping $\xb(\xib^{(e)})=\sum_j N_j(\xib^{(e)})\xb_j^{(e)}$, with the Jacobian $\Jb(\xib^{(e)})=\partial\xb/\partial\xib^{(e)}$. Differentiating \Cref{eq sf 1} and applying the chain rule yields the gradient of each displacement component with respect to the physical coordinates:
%=======================================
\begin{equation}\label{eq sf grad}
 \frac{\partial u_i^{(e)}(\xb)}{\partial \xb} = \Jb^{-\text{T}}\big(\xib^{(e)}\big) \sum_{j=1}^{n_s}\frac{\partial N_j\big(\xib^{(e)}\big)}{\partial \xib^{(e)}}\, u_i\big(\xb^{(e)}_j;\thetab_u\big), \quad i=1,\dots,n_d.
\end{equation}
%=======================================
Since the mesh is fixed, $\Jb^{-\text{T}}(\cdot)$ and the shape function derivatives can be precomputed and combined into the strain-displacement matrix that is stored for every element.

To express the kinematic and constitutive relations compactly, we adopt Voigt notation. In 3D, the strain and stress vectors are defined as:
%=======================================
\begin{equation} \label{eq voigt}
    \epsilonb^v(\xb) =
    \begin{bmatrix}
        \varepsilon_{xx} & \varepsilon_{yy} & \varepsilon_{zz} & \gamma_{xy} & \gamma_{yz} & \gamma_{zx}
    \end{bmatrix}^{\text{T}}, \qquad
    \sigmab^v(\xb) =
    \begin{bmatrix}
        \sigma_{xx} & \sigma_{yy} & \sigma_{zz} & \tau_{xy} & \tau_{yz} & \tau_{zx}
    \end{bmatrix}^{\text{T}},
\end{equation}
%=======================================
where $\gamma_{xy}=2\varepsilon_{xy}$, $\gamma_{yz}=2\varepsilon_{yz}$, and $\gamma_{zx}=2\varepsilon_{zx}$ are the engineering shear strains (the dependence of the individual stress and strain components on $\xb$ is omitted for brevity). The nodal displacement components of element $e$ are assembled into
$\ub^{(e)}=[u_x^{(1)},u_y^{(1)},u_z^{(1)},\ldots,u_x^{(n_s)},u_y^{(n_s)},u_z^{(n_s)}]^{\mathrm{T}}\in\mathbb{R}^{n_d n_s}$
where $u_x^{(j)}$, $u_y^{(j)}$, and $u_z^{(j)}$ denote the displacement components at node $j$. The discrete kinematic relation is then written as:
%=======================================
\begin{equation}\label{eq Bu}
    \epsilonb^v(\xb) = \Bb\big(\xib^{(e)}\big)\, \ub^{(e)}, \qquad
    \Bb = \big[\, \Bb_1 \;\; \Bb_2 \;\; \cdots \;\; \Bb_{n_s} \,\big],
\end{equation}
%=======================================
where each nodal block of the strain--displacement matrix $\Bb$ is constructed from the physical-coordinate derivatives of the shape functions in \Cref{eq sf grad}:
%=======================================
\begin{equation} \label{eq B}
\Bb_j
=
\begin{bmatrix}
\dfrac{\partial N_j}{\partial x} & 0 & 0 \\[4pt]
0 & \dfrac{\partial N_j}{\partial y} & 0 \\[4pt]
0 & 0 & \dfrac{\partial N_j}{\partial z} \\[4pt]
\dfrac{\partial N_j}{\partial y} & \dfrac{\partial N_j}{\partial x} & 0 \\[4pt]
0 & \dfrac{\partial N_j}{\partial z} & \dfrac{\partial N_j}{\partial y} \\[4pt]
\dfrac{\partial N_j}{\partial z} & 0 & \dfrac{\partial N_j}{\partial x}
\end{bmatrix}, \quad j = 1,\dots,n_s.
\end{equation}
%=======================================
For an isotropic material, the stress follows from:
% \fix{is there a reason you now denote C with D? Is it only because you've moved to Voigt notation? if yes, then why not just write $\bm{C}^v$?}
%=======================================
\begin{equation} \label{eq DBu}
    \sigmab^v(\xb) = \Cb^v(\xb)\,\epsilonb^v(\xb)=\Cb^v(\xb)\,\Bb\big(\xib^{(e)}\big)\,\ub^{(e)},
\end{equation}
%=======================================
% where the \fix{material matrix new name?} is written in terms of the Lam\'e parameters:
where $\Cb^v(\xb)$ is the elasticity tensor $\Cb(\xb)$ in Voigt notation which is written in terms of the Lam\'e parameters:
%=======================================
\begin{equation}\label{eq D}
\Cb^v(\xb)
=
\begin{bmatrix}
\lambda+2\mu & \lambda & \lambda & 0 & 0 & 0 \\
\lambda & \lambda+2\mu & \lambda & 0 & 0 & 0 \\
\lambda & \lambda & \lambda+2\mu & 0 & 0 & 0 \\
0 & 0 & 0 & \mu & 0 & 0 \\
0 & 0 & 0 & 0 & \mu & 0 \\
0 & 0 & 0 & 0 & 0 & \mu
\end{bmatrix},
\qquad
\lambda(\xb)=\frac{\nu E(\xb)}{(1+\nu)(1-2\nu)},
\qquad
\mu(\xb)=\frac{E(\xb)}{2(1+\nu)},
\end{equation}
%=======================================
where $\nu$ is Poisson's ratio and $E(\xb)$ is the interpolated Young's modulus defined in \Cref{eq E} (the corresponding plane-stress constitutive stiffness matrix is used for the 2D examples). We denote the stiffness matrix of the fully solid material by $\Cb^v_0$ for which $E=E_0$. The solid-material stress used to evaluate the stress constraint is then given by $\sigmab^{v,s}(\xb)=\Cb^v_0\Bb\ub^{(e)}$, and the equivalent stress in \Cref{eq svm} can be expressed in the following quadratic form:
%=======================================
\begin{equation}\label{eq svm M}
    \sigma_{\mathrm{VM}}
    = \sqrt{\sigmab^{v\text{T}} \Mb\, \sigmab^v},
    \qquad
    \Mb =
    \begin{bmatrix}
        1 & -\tfrac{1}{2} & -\tfrac{1}{2} & 0 & 0 & 0 \\
        -\tfrac{1}{2} & 1 & -\tfrac{1}{2} & 0 & 0 & 0 \\
        -\tfrac{1}{2} & -\tfrac{1}{2} & 1 & 0 & 0 & 0 \\
        0 & 0 & 0 & 3 & 0 & 0 \\
        0 & 0 & 0 & 0 & 3 & 0 \\
        0 & 0 & 0 & 0 & 0 & 3
    \end{bmatrix},
\end{equation}
%=======================================
which recovers the 3D von Mises stress in \Cref{eq svm 3d}. The same operators are used for the adjoint field, with $\ub^{(e)}$ replaced by the nodal adjoint displacement vector $\ub^{(a,e)}$ in \Cref{eq Bu,eq DBu} to obtain the corresponding adjoint strain and stress.

All integrals in loss terms are evaluated by Gauss quadrature with a single integration point at the element center, $\xib^{(e)*} = \mathbf{0}$, i.e., reduced integration. The strain energy can be written as:
%=======================================
\begin{equation}\label{eq sf int}
\frac{1}{2}\int_{\Omega} \epsilonb:\Cb:\epsilonb \, dV \approx \frac{1}{2}\sum_{e=1}^{n_e} \omega_g\, \sigmab^{v\text{T}}\big(\xib^{(e)*}\big)\,\epsilonb^{v}\big(\xib^{(e)*}\big)\det\Jb\big(\xib^{(e)*}\big),
\end{equation}
%=======================================
where $n_e$ is the number of elements and $\omega_g=8$ in 3D ($\omega_g=4$ in 2D) is the single-point Gauss quadrature weight. The external work is evaluated as the sum of the products of the prescribed nodal forces and the corresponding displacements over all loaded nodes. The density $\rho(\xb)$, interpolated Young's modulus, and local stress constraint are evaluated at the element centers. Consequently, the number of local stress constraints equals the number of centroid CPs or elements, such that $N_c=n_e$. Consistent with \ifanon the findings of \textcite{sun_multi_2026}\else our previous findings \cite{sun_multi_2026}\fi, full Gauss quadrature increases the computational cost without noticeably affecting the optimized designs. The single-point quadrature rule also facilitates direct comparison with FE-based methods such as PolyStress, in which the stress constraints are evaluated at the element centers. 

For each sub-slice in both the single- and multi-GPU settings, the corresponding strain--displacement matrices $\Bb$ are reconstructed from the local nodal CPs and element connectivity during preprocessing. These matrices are then cached together with the Jacobian determinants, element connectivity, and analytical weight functions on the assigned GPU. Consequently, the loss and its gradient are computed by first evaluating the three networks at the nodal and centroid CPs and then applying parallel matrix--vector operations over all background elements within each sub-slice.

\subsection{Adjoint Sensitivity Analysis}\label{sec metd adj}
We obtain the gradients of the volume fraction objective $C_V(\rho)$ and the energy functionals with respect to the trainable parameters $\{\thetab_u, \thetab_u^{(a)}, \thetab_\rho \}$ directly via AD. The stress penalty $\mathcal{P}(g)$, however, depends on the density not only explicitly through $m_E(\rho)$ but also implicitly through the displacement field $\ub(\xb)$, which varies with the current design. In our simultaneous formulation, the displacement field is represented by an independent network. Consequently, differentiating $\mathcal{P}(g)$ with respect to $\thetab_\rho$ through AD captures only the explicit dependence and omits the implicit contribution through $\ub(\xb)$. We therefore employ the adjoint method to recover this contribution. Because $\mathcal{P}(g)$ is evaluated as a sum over a finite set of CPs, we formulate the adjoint derivation in a discrete setting consistent with the FE-based approach in \textcite{giraldo_polystress_2021}. We collect the displacements at all mesh nodes into the matrix $\Ub=[\ub(\xb_1),\ldots,\ub(\xb_{N_n})]^{\mathrm{T}}\in\mathbb{R}^{N_n\times n_d}$, where $N_n$ denotes the total number of mesh nodes.

In discrete form, the equilibrium condition in \Cref{eq sc1 b} requires the residual force to vanish:
%=======================================
\begin{equation}\label{eq R}
\Rb(\Ub,\rho) = \frac{\partial \mathcal{L}_M}{\partial\Ub} = \Fb_{int}(\Ub,\rho) - \Fb_{ext} = \mathbf{0},
\end{equation}
%=======================================
where $\Fb_{int}$ is the internal force assembled from the element stresses using \Cref{eq DBu,eq sf int} while $\Fb_{ext}$ is the external force that is independent of the design. Since $\Rb = \mathbf{0}$ at equilibrium, contracting it with an arbitrary field and adding the resulting term to \Cref{eq P} does not change the penalty. Hence: 
%=======================================
\begin{equation}\label{eq Pa}
\mathcal{P}^{(a)}(g(\xb)) = \left( \frac{1}{2N_c} \sum_{i=1}^{N_c} h_i^2 \right) + \alpha_{adj}\, \Ub^{(a)}:\Rb(\Ub,\rho),
\end{equation}
%=======================================
where $h_i$ is a shorthand for $h(\xb_i)$ in \Cref{eq h}, $\Ub^{(a)}\in\mathbb{R}^{N_n\times n_d}$ is the matrix of nodal adjoint displacements assembled from the adjoint field $\ub^{(a)}(\xb)$ in the same manner as $\Ub$, the colon operator $(\mathbin{:})$ denotes the sum of the entrywise products, and $\alpha_{adj}>0$ is a scaling factor whose role is discussed in \Cref{sec metd train}. Applying the chain rule to differentiate \Cref{eq Pa} with respect to the density $\rho$ yields:
%=======================================
\begin{equation}\label{eq P_rho1}
\frac{\partial \mathcal{P}^{(a)}(\cdot)}{\partial \rho} =\frac{1}{N_c} \sum_{i=1}^{N_c} h_i \bigg(\frac{\partial h_i}{\partial \rho} + \frac{\partial h_i}{\partial \Ub}:\frac{\partial \Ub}{\partial \rho}\bigg)
+ \alpha_{adj}\,\Ub^{(a)}:\bigg(\frac{\partial \Fb_{int}}{\partial\Ub}\frac{\partial\Ub}{\partial \rho} + \frac{\partial \Fb_{int}}{\partial \rho}\bigg).
\end{equation}
%=======================================
The derivative $\partial\Ub/\partial\rho$ is generally unavailable. To eliminate the terms involving this derivative, we set $\Ub^{(a)}$ to the solution of the following adjoint problem:
%=======================================
\begin{equation} \label{eq disc_adj}
\bigg(\frac{\partial \Fb_{int}}{\partial \Ub}\bigg)^{\text{T}}\Ub^{(a)} = -\frac{1}{N_c\,\alpha_{adj}}\sum_{i=1}^{N_c}h_i\frac{\partial h_i}{\partial \Ub}
=: \frac{1}{\alpha_{adj}}\Fb^{(a)}.
\end{equation}
%=======================================
With this choice, the sensitivity reduces to:
%=======================================
\begin{equation}\label{eq P_rho2}
\frac{\partial \mathcal{P}^{(a)}(\cdot)}{\partial \rho} =\frac{1}{N_c} \sum_{i=1}^{N_c} h_i \frac{\partial h_i}{\partial \rho}
+\alpha_{adj}\,\Ub^{(a)}:\frac{\partial \Fb_{int}}{\partial\rho}.
\end{equation}
%=======================================
The first term represents the explicit sensitivity of the vanishing constraint to the density through $m_E(\rho)$, whereas the second term accounts for the implicit effect of a density perturbation on the equilibrium displacement field. When evaluating $\partial\Fb_{int}/\partial\rho$, the displacement field is held fixed, such that only the interpolated modulus $m_E(\rho)$ is differentiated. The contraction of this derivative with the adjoint displacement can be expressed elementwise as the inner product of the adjoint strain and the solid-material stress, scaled by $\partial m_E/\partial\rho$. This term therefore quantifies the change in the equilibrium response caused by a local change in material stiffness. The nodal adjoint force $\Fb^{(a)}$ in \Cref{eq disc_adj} receives contributions only from active constraints. For an active constraint ($g_i>0$ and $h_i=g_i$), the chain rule gives:
%=======================================
\begin{equation}\label{eq h_U}
    \frac{\partial h_i}{\partial \Ub}
    = \frac{\partial g_i}{\partial \sigma_{\mathrm{VM}}} \frac{\partial \sigma_{\mathrm{VM}}}{\partial \sigmab^{v}} \cdot \frac{\partial \sigmab^{v}}{\partial \Ub},
\end{equation}
%=======================================
where every stress is the solid-material stress $\sigmab^{v,s}$ at $\xb_i$. From \Cref{eq g,eq s}:
%=======================================
\begin{equation}\label{eq g_seq}
    \frac{\partial g_i}{\partial \sigma_{\mathrm{VM}}}
    = \frac{1}{\sigma_y}\frac{\partial g_i}{\partial s}
    =
    \frac{m_E(\xb_i)}{\sigma_y}\left(3s^2 + 4s + 1\right),
\end{equation}
%=======================================
which is positive for every $s>0$ and grows faster with the violation than the corresponding PolyStress derivative $m_E(3s^2+1)/\sigma_y$. From \Cref{eq svm M}, the derivative of the equivalent stress with respect to the Voigt stress vector is:
%=======================================
\begin{equation}\label{eq seq_s}
    \frac{\partial \sigma_{\mathrm{VM}}}{\partial \sigmab^{v}}=\frac{\Mb\sigmab^{v}}{\sigma_{\mathrm{VM}}}.
\end{equation}
%=======================================
Furthermore, evaluating \Cref{eq DBu} using the solid-material constitutive matrix gives:
%=======================================
\begin{equation}\label{eq s_U}
    \frac{\partial \sigmab^{v}}{\partial \ub^{(e)}}=\Cb^v_0\Bb\big(\xib^{(e)*}\big).
\end{equation}
%=======================================
Combining \Cref{eq h_U,eq g_seq,eq seq_s,eq s_U}, the contribution of an active constraint at the $i$-th CP to $\Fb^{(a)}$ is the $n_d n_s$-dimensional vector $-\frac{1}{N_c}h_i\frac{\partial g_i}{\partial\sigma_{\mathrm{VM}}}\Bb^{\mathrm{T}}\Cb^v_0\Mb\sigmab^{v,s}/\sigma_{\mathrm{VM}}$. These element-level contributions are computed from the stress at the centroid CPs and assembled to the nodal CPs of the corresponding elements. Because inactive constraints make no contribution, the resulting adjoint load is concentrated in regions where the equivalent stress exceeds the yield strength.

Instead of solving the linear system in \Cref{eq disc_adj}, which would require assembling and factorizing the tangent stiffness $\partial\Fb_{int}/\partial\Ub$, we recast the adjoint system variationally. Specifically, \Cref{eq disc_adj} corresponds to the stationarity condition of the quadratic functional:
%=======================================
\begin{equation}\label{eq LMa}
\mathcal{L}_M^{(a)}\big(\ub^{(a)},\ub,\rho\big) =
\frac{1}{2}\int_{\Omega}
\epsilonb\big(\ub^{(a)}\big):
\Cb\big(\xb\big):
\epsilonb\big(\ub^{(a)}\big)\, dV
-\frac{1}{\alpha_{adj}}\,\Ub^{(a)}:\Fb^{(a)},
\end{equation}
%=======================================
which mirrors the potential energy in \Cref{eq LM}, with the adjoint field in place of the (primal) displacement field and $\Fb^{(a)}$ in place of the external load. We minimize it over $\thetab_u^{(a)}$ just as we minimize $\mathcal{L}_M(\cdot)$ over $\thetab_u$. The adjoint problem therefore needs no separate solver; it is simply one more energy functional in the loss, leveraging the same cached matrices. We refer to the first and second terms on the right-hand side of \Cref{eq LMa} as the strain energy and body work of the adjoint field, respectively.

With the adjoint quantities in place, we convert the constrained problem in \Cref{eq sc1} into the unconstrained minimization of a single loss:
%---------------------------------------
\begin{equation}\label{eq penal}
\mathcal{L}\big(\ub,\ub^{(a)},\rho\big)
=
C_V(\rho)
+\omega_C\, \mathcal{P}^{(a)}\big(g(\xb)\big)
+\omega_D\Big( \mathcal{L}_M\big(\ub,\rho\big) + \mathcal{L}_M^{(a)}\big(\ub^{(a)},\ub,\rho\big)\Big),
\end{equation}
%---------------------------------------
where $\omega_C$ and $\omega_D$ are global penalty weights for the stress-constraint term and the two potential-energy functionals, respectively. The four terms serve distinct purposes. The functional $\mathcal{L}_M(\cdot)$ drives the displacement field toward equilibrium for the current design while $\mathcal{L}_M^{(a)}(\cdot)$ drives the adjoint field toward equilibrium under the adjoint load $\Fb^{(a)}$. The terms $C_V(\rho)$ and $\mathcal{P}^{(a)}(\cdot)$ govern the design update. We enforce this separation by detaching the appropriate nodes from the computational graph during backpropagation. In $\mathcal{L}_M(\cdot)$ and $\mathcal{L}_M^{(a)}(\cdot)$, the interpolated modulus is treated as constant with respect to $\thetab_\rho$. So, these functionals only update the parameters of the displacement and adjoint PGCANs. In $\mathcal{P}^{(a)}(\cdot)$, $\Ub$ and $\Ub^{(a)}$ are treated as constants with respect to $\thetab_u$ and $\thetab_u^{(a)}$, respectively. Consequently, the penalty only affects the design network through \Cref{eq P_rho2}. Thus, minimizing \Cref{eq penal} with respect to all parameters produces a SAND iteration, with each network receiving only the sensitivities associated with its respective field.

Two practical implementation details are worth noting. First, $\Fb^{(a)}$ is a global quantity and cannot be assembled independently within the individual subdomains and sub-slices. We therefore construct $\Fb^{(a)}$ using the displacement field from the previous iteration. Because the design evolves gradually, this one-step lag has a negligible effect on the optimization. Second, the assembled adjoint load is dynamically scaled by $\alpha_{adj}$ before being used in \Cref{eq LMa}. This scaling maintains a consistent magnitude of the adjoint field throughout training.

\subsection{Stabilization and Curriculum Training}\label{sec metd train}
The displacement and adjoint fields generally do not satisfy their respective equilibrium conditions at the beginning of training because the PGCAN parameters are randomly initialized. Moreover, the equilibrium solution of the adjoint field depends on the evolving displacement and density fields, posing a significant challenge to the simultaneous optimization framework.
We resolve this instability issue by adopting a three-stage training strategy that aligns with the underlying physical hierarchy. During Stage 1, which covers the first $5\%$ of the total epochs ($n_{\text{tot}}=30,000$), we minimize only $\mathcal{L}_M(\cdot)$ to bring the displacement network close to mechanical equilibrium. Then, in Stage 2 we minimize $\mathcal{L}_M(\cdot) + \mathcal{L}_M^{(a)}(\cdot)$ over the subsequent $5\%$ of total epochs to update the adjoint field against the obtained displacement solution, keeping the design network frozen across both warm-up stages. Finally, in Stage 3 we minimize the full loss in \Cref{eq penal} to update all three networks concurrently.

To establish a reasonable initial design, we adjust the bias in the final layer of the PGCAN decoder for $m_\rho(\xb)$ such that the projected density field has an initial volume fraction of approximately $0.60$ for all examples. This scalar bias uniformly shifts the decoder output without altering its spatial variation. The density network remains frozen during Stages 1 and 2, maintaining this volume fraction for the first $10\%$ of the training epochs.

Although Stages 1 and 2 are dedicated to establishing the initial force equilibrium, our studies indicate that maintaining equilibrium for the adjoint field is challenging. The magnitude of the adjoint load $\Fb^{(a)}$ depends on both the proportion of active constraints and the severity of their violations, which causes it to vary by several orders of magnitude during training.
Without appropriate scaling, the corresponding equilibrium solution changes too abruptly for the adjoint PGCAN to evolve in a stable manner. To address this issue, in \Cref{eq Pa,eq disc_adj,eq LMa} we introduce the scaling factor $\alpha_{adj}$ defined as
%=======================================
\begin{equation}\label{eq alpha adj}
    \alpha_{adj} = \frac{\max_{k}\big|F^{(a)}_{k}\big|}{F_{goal}}, \qquad F_{goal} = c_F\,N_c,
\end{equation}
%=======================================
where $F^{(a)}_k$ ranges over all entries of the assembled adjoint force and $c_F$ is a constant that sets the maximum magnitude of the scaled force $\Fb^{(a)}/\alpha_{adj}$ in every iteration.
We set $c_F = 10$ for 2D examples and $c_F=1$ for 3D examples. Since we scale the adjoint term in \Cref{eq Pa} by $\alpha_{adj}$ and scale the adjoint load back in \Cref{eq disc_adj}, the sensitivity in \Cref{eq P_rho2} remains intact. This normalization is crucial for stabilizing simultaneous TO and will be further discussed in \Cref{sec res}.

Following \ifanon earlier work\else our earlier work\fi\ \cite{sun_smo_2025,sun_multi_2026} and the curriculum learning literature \cite{soviany_curriculum_2022}, we apply continuation to the hyperparameters that control the nonconvexity of the problem. Because the volume fraction $C_V(\cdot)$ serves as the objective rather than a constraint, no volume continuation is imposed. 
The SIMP exponent ($p=3\to5$), the projection sharpness in \Cref{eq proj} ($\beta=4\to8$), and the constraint weight in \Cref{eq penal} ($\omega_C=1\to10^3$) are held at their initial values during the first $10\%$ warm-up epochs. Each parameter then increases linearly between epochs $0.1\,n_{\text{tot}}$ and $0.6\,n_{\text{tot}}$ and remains constant thereafter.
Starting with a small $\omega_C$ lets $C_V(\cdot)$ remove material freely while the stress field equilibrates, and the subsequent growth of $\omega_C$ enforces the constraints more aggressively, akin to the role of the multiplier update in augmented Lagrangian methods. The weight of the energy functionals is held fixed at $\omega_D = 10^3$. 

In addition, the effective stress limit $\sigma_y$ in \Cref{eq sc1 c} during training is defined as $\alpha_s\sigma_{\mathrm{y0}}$, where $\sigma_{\mathrm{y0}}$ is the yield strength of the material and $\alpha_s$ is a safety factor that decays linearly from $\alpha_s = 1$ to $\alpha_s = 0.9$ over the same interval and is then held constant for all examples.  Consequently, the constraints are tightened progressively as their weight grows so that the final design maintains a $10\%$ margin below the nominal limit $\sigma_{\mathrm{y0}}$, providing a buffer for discretization errors in stress evaluations.
We can widen this safety margin at the cost of a proportionally heavier design.

All NN parameters are updated with the Adam optimizer \cite{kingma_adam_2017}. We train the displacement and adjoint networks with learning rates of $10^{-3}$ and $10^{-2}$, respectively; the adjoint network takes the higher learning rate because its target changes with every design update and requires rapid tracking. The design network adopts a substantially smaller rate ($2 {\times} 10^{-5}$ in 2D, $10^{-5}$ in 3D) to restrict per-epoch density updates and preserve state equilibrium. We also clip the parameter gradients of the state and design networks to a maximum norm of $0.1$ and $0.01$, respectively, to bound the update steps further. Finally, we exclude elements whose density falls below $10^{-3}$ from loss assembly and refresh the active elements every $20$ epochs. As the material is removed and the active sets of CPs shrink, the per-epoch cost drops as training proceeds.
    \section{Results and Discussion}\label{sec res}
\begin{figure*}[!t]
    \centering
    \begin{subfigure}[t]{0.24\textwidth}
        \captionsetup{justification=raggedright, singlelinecheck=false, skip=0pt, position=top,labelfont=subtxt, textfont=subtxt}
        \caption{L beam 2D}
        \includegraphics[width=\linewidth]{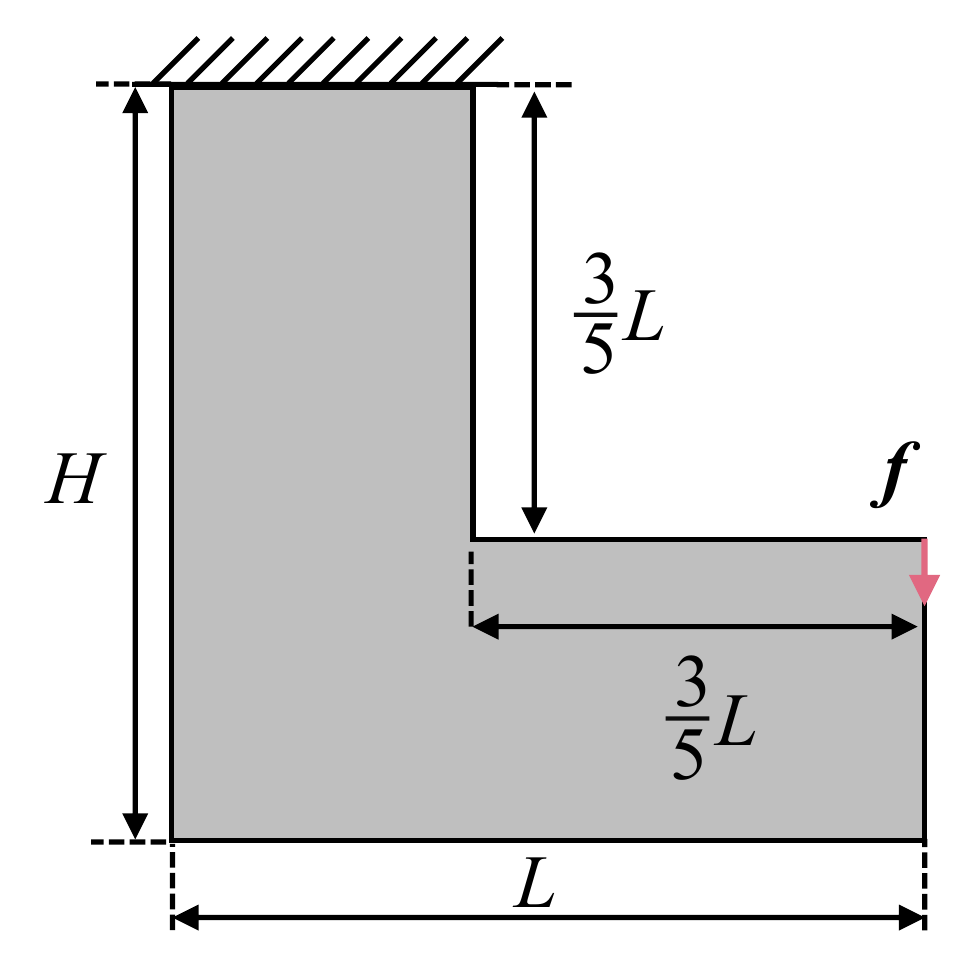}
        \label{fig exam2D 1}
    \end{subfigure}
    \hfill
    \begin{subfigure}[t]{0.24\textwidth}
        \captionsetup{justification=raggedright, singlelinecheck=false, skip=0pt, position=top,labelfont=subtxt, textfont=subtxt}
        \caption{Crack 2D}
        \includegraphics[width=\linewidth]{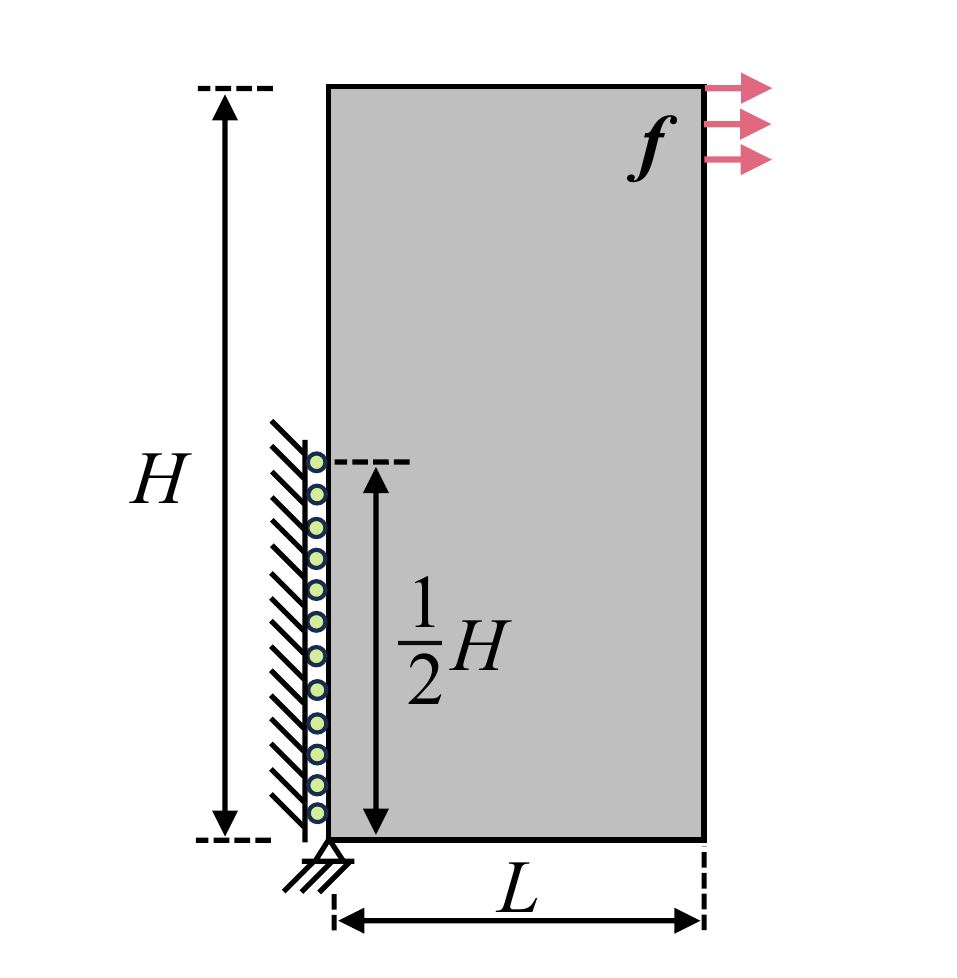}
        \label{fig exam2D 2}
    \end{subfigure}
    \hfill
    \begin{subfigure}[t]{0.24\textwidth}
        \captionsetup{justification=raggedright, singlelinecheck=false, skip=0pt, position=top,labelfont=subtxt, textfont=subtxt}
        \caption{Hook 2D}
        \includegraphics[width=\linewidth]{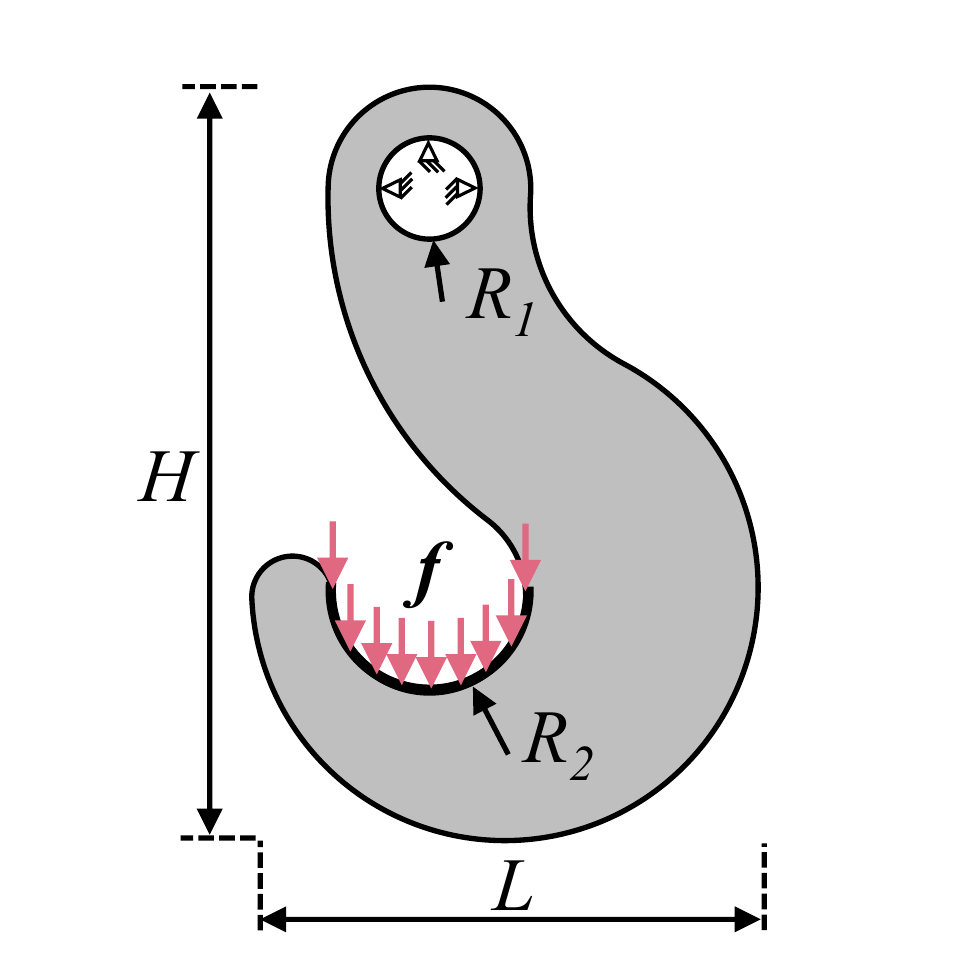}
        \label{fig exam2D 3}
    \end{subfigure}
    \hfill
    \begin{subfigure}[t]{0.24\textwidth}
        \captionsetup{justification=raggedright, singlelinecheck=false, skip=0pt, position=top,labelfont=subtxt, textfont=subtxt}
        \caption{Portal frame 2D}
        \includegraphics[width=\linewidth]{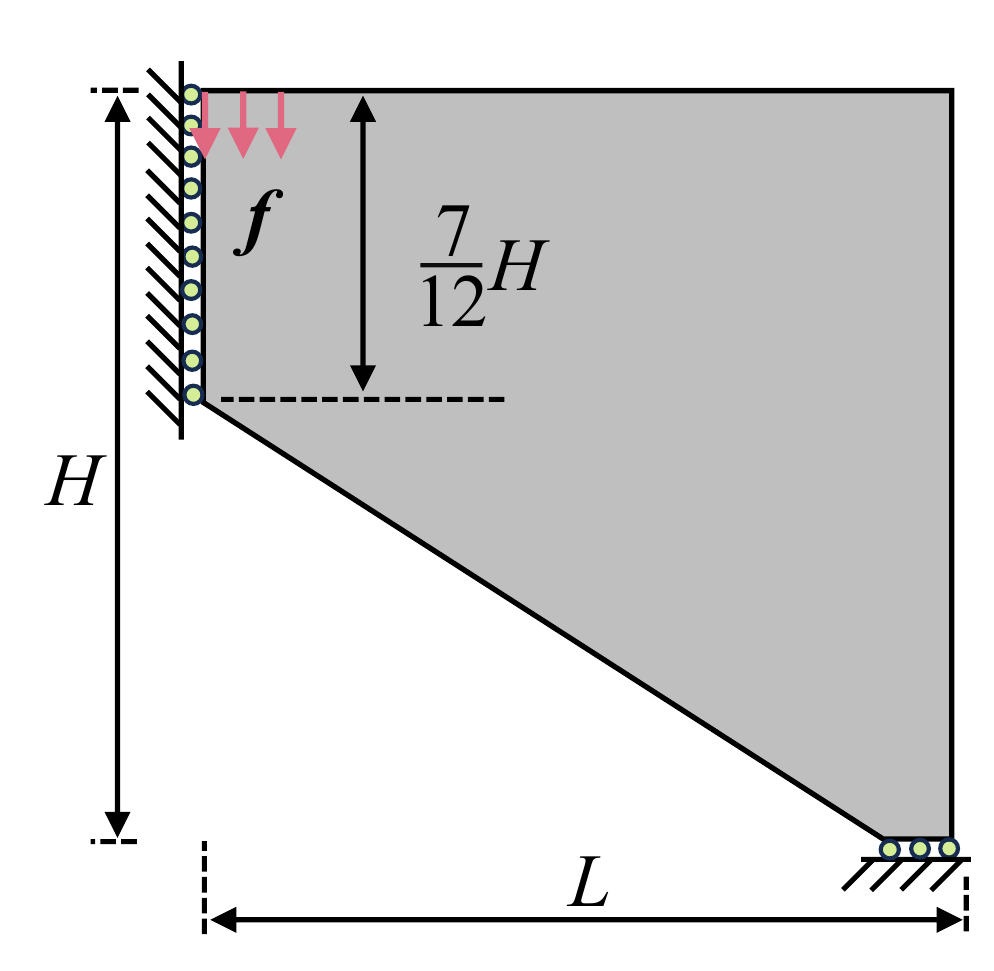}
        \label{fig exam2D 4}
    \end{subfigure}
    \caption{\textbf{Design domains, boundary conditions, geometries and applied external loads for 2D benchmarks:}
    \textbf{(a)} the L beam example with a re-entrant corner, \textbf{(b)} the crack example with a constrained edge segment, \textbf{(c)} the hook example with distributed loading along the inner contact surface, and \textbf{(d)} the portal frame example with the half design domain.}
    \label{fig exam2D}
\end{figure*}

\begin{figure*}[!b]
    \centering
    \begin{subfigure}[t]{0.29\textwidth}
        \captionsetup{justification=raggedright, singlelinecheck=false, skip=0pt, position=top,labelfont=subtxt, textfont=subtxt}
        \caption{L beam 3D}
        \includegraphics[width=\linewidth]{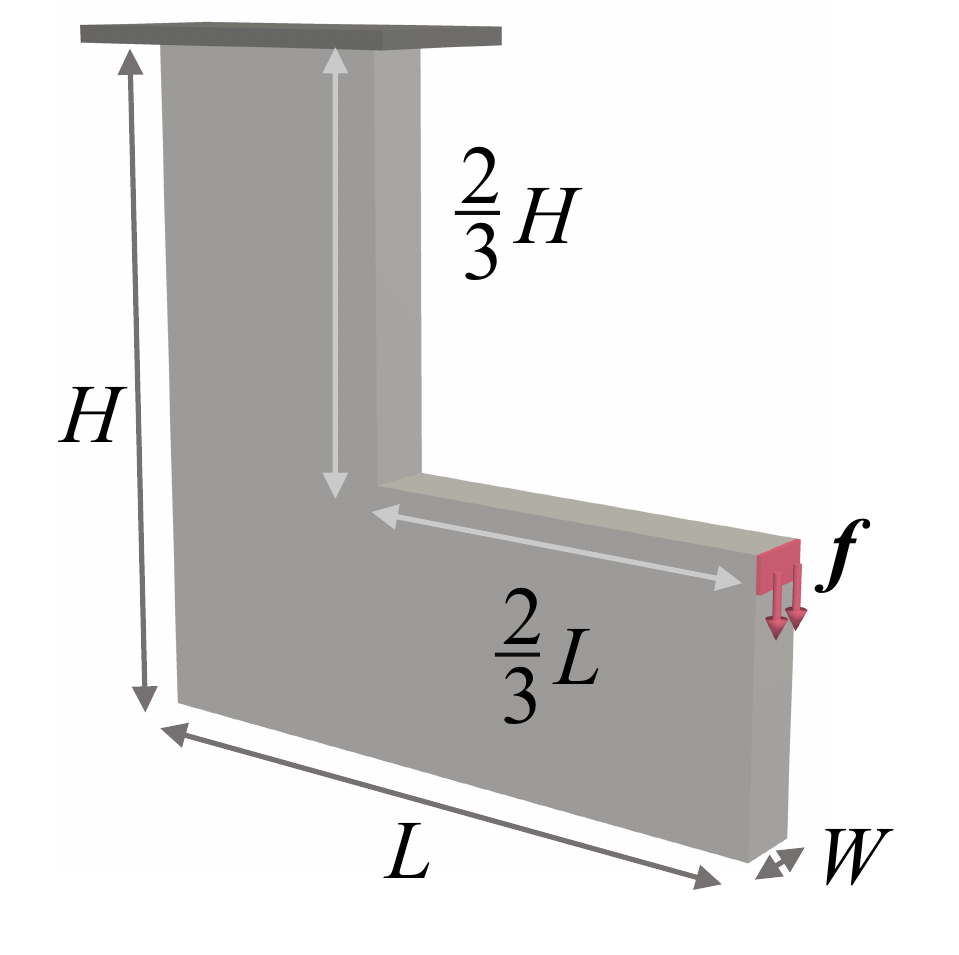}
        \label{fig exam3D 1}
    \end{subfigure}
    \begin{subfigure}[t]{0.29\textwidth}
        \captionsetup{justification=raggedright, singlelinecheck=false, skip=0pt, position=top,labelfont=subtxt, textfont=subtxt}
        \caption{Portal frame 3D}
        \includegraphics[width=\linewidth]{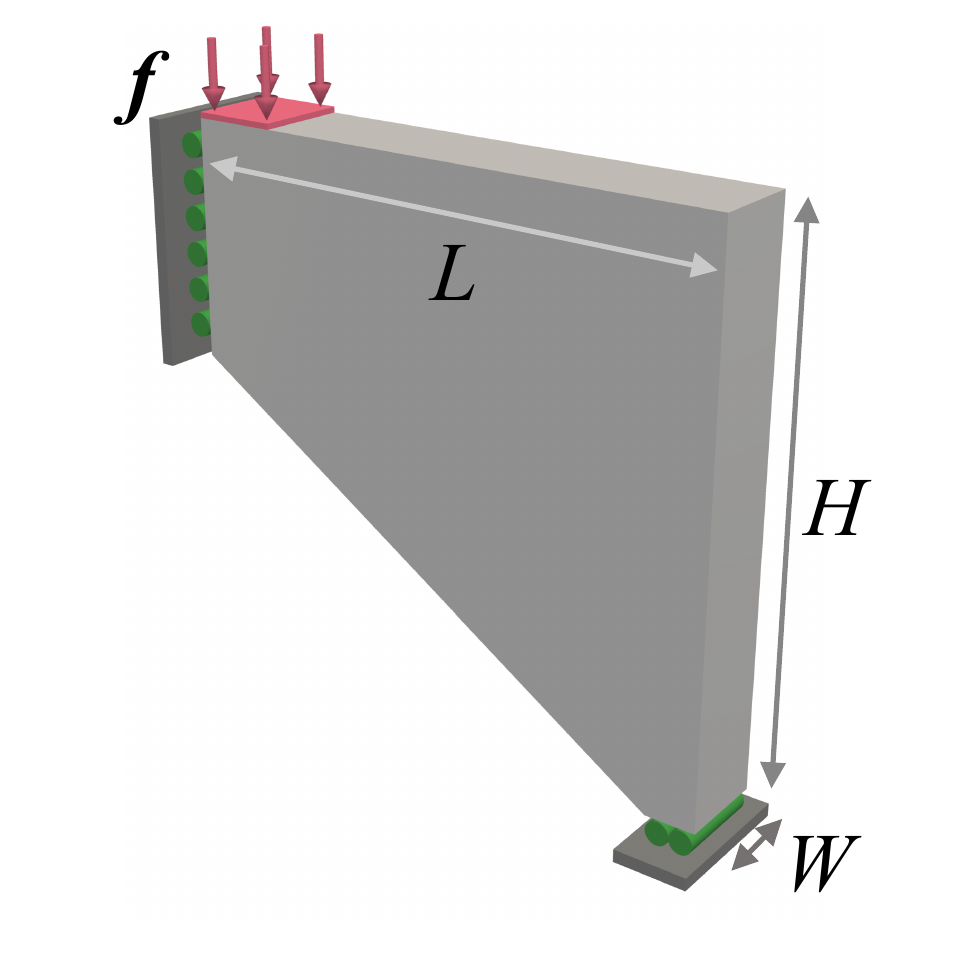}
        \label{fig exam3D 2}
    \end{subfigure}
    \vspace{-2em}
    \caption{\textbf{Large-scale 3D stress-constrained TO benchmarks used to evaluate the scalability of the proposed framework:}
    \textbf{(a)} the L beam example with a re-entrant corner;
    \textbf{(b)} the portal frame example with a horizontally constrained side face, a supported lower surface, and a distributed top load surface.}
    \label{fig exam3D}
\end{figure*}

We assess SNTO on four 2D and two 3D stress-constrained benchmarks in \Cref{sec res 2D,sec res 3D}, respectively. 
The 2D cases are visualized in \Cref{fig exam2D} and used to verify SNTO against PolyStress \cite{giraldo_polystress_2021}, an open-source FE-based code that enforces local stress constraints through an augmented Lagrangian method. The 3D benchmarks are shown in \Cref{fig exam3D} and we use them to demonstrate scalability up to eight million local stress constraints and quantify the cost of GPU parallelization and the sub-slicing scheme described in \Cref{sec metd gpu}. \Cref{tab dimension} lists the domain dimensions and load magnitudes across all 2D and 3D benchmarks. 

%========================================
%=======================================

The L beam shown in \CrefSub{fig exam2D}{fig exam2D 1} is a canonical benchmark where the re-entrant corner creates a stress singularity. The crack example in \CrefSub{fig exam2D}{fig exam2D 2} represents half a plate with a central edge crack loaded in tension, where stress concentrates at the crack tip. Invoking symmetry, we model only its right half and represent the crack faces as traction-free edges. The hook geometry shown in \CrefSub{fig exam2D}{fig exam2D 3} is adopted from \textcite{giraldo_polystress_2021} and features curved boundaries with a distributed load applied along the lower half of the inner arc. A solid ring of thickness $0.02\,\mathrm{mm}$ is prescribed around the inner arc to maintain a feasible load path. The portal frame in \CrefSub{fig exam2D}{fig exam2D 4} is also modeled as a half domain with symmetry conditions on the left edge. We assume plane stress conditions for all 2D examples with thickness $W$ matching that of the 3D examples. 

PolyStress employs a polygonal mesh and is a MATLAB code that we run on an Intel\textsuperscript{\textregistered} Core\textsuperscript{TM} i7-11700K CPU with 64~GB of memory. For our 2D comparisons, we use PolyStress with approximately $250{,}000$ elements and a filter radius of $0.05\,\nimm$. We repeat each run three times with random density initializations and report the median results. We implement SNTO in PyTorch and train all 2D cases for $30{,}000$ epochs on a single NVIDIA A100 GPU with 40~GB of memory. Following the approximation scheme described in \Cref{sec metd num}, for each 2D example we discretize the domain using quadrilateral elements. To assess sensitivity to the number of elements and initialization, we use meshes with 125K, 250K, 500K, and 1M nodal CPs and repeat the training using five random seeds at each mesh with the 250K mesh matching the PolyStress settings.
The three coarser cases fit on the GPU as a single slice. The 1M case does not fit and its CPs are therefore split into two sub-slices that are processed one after the other, see \CrefSub{fig gpu}{fig gpu 1}. We emphasize that the background mesh in SNTO is used only for numerical quadrature and differentiation with SNTO retaining a mesh-free parameterization that does not need a filter since the state, design, and adjoint fields are represented as continuous functions over the domain.
We also report the resolution of PGCAN's feature grid $\Res=(N_x^e,N_y^e)$ in \Cref{tab mesh}, which we choose proportional to the design domain dimensions so that the grid cells are approximately square. 

%========================================
%=======================================

%=======================================
\begin{table*}[!t]
    \centering
    \renewcommand{\arraystretch}{1.5}
    \small
    \setlength\tabcolsep{6pt}
    \caption{\textbf{Geometry and loading of the benchmark examples:}
    $H$ and $L$ are the height and width of the design domain, and $W$ is the out-of-plane dimension, which is the plane stress thickness in 2D and the extrusion depth in 3D. For the hook, $R_1$ and $R_2$ are the radii of the fixed hole at the top and the loaded hole at the bottom, respectively; the upper half of the $R_1$ rim is clamped and the lower half of the $R_2$ rim carries the load. $L_s$ is the length of the loaded boundary segment, which for the hook is the arc length $\pi R_2$ of the loaded half-rim, and $f$ is the total external force distributed uniformly over $L_s$ in 2D and over $L_s \times W$ in 3D.}
    \label{tab dimension}
    \begin{tabular}{l|c|c|c|c|c|c|c} 
    \hline
    \textbf{Example} & $H$ ($\mathrm{mm}$) & $L$ ($\mathrm{mm}$) & $W$ ($\mathrm{mm}$) & $R_1$ ($\mathrm{mm}$) & $R_2$ ($\mathrm{mm}$) & $L_s$ ($\mathrm{mm}$) & $f$ ($\mathrm{N}$)\\ 
    \hline
    L beam 2D          & 1   & 1 & 0.1   & --   & --    & 0.05       & 0.2 \\
    Crack 2D          & 2   & 1 & 0.1   & --   & --    & 0.1       & 0.5 \\
    Hook 2D          & 1.5   & 1 & 0.1   & 0.1   & 0.19    & 0.597       & 0.5 \\
    Portal frame 2D          & 0.6   & 0.6 & 0.1   & --   & --    & 0.05       & 0.15 \\
    L beam 3D          & 1   & 1 & 0.1   & --   & --    & 0.05       & 0.2 \\
    Portal frame 3D          & 0.6   & 0.6 & 0.1   & --   & --    & 0.05       & 0.15 \\
    \hline
    \end{tabular}
\end{table*}
%=======================================
The 3D examples and their boundary conditions are displayed in \Cref{fig exam3D}. We generate them by extruding the L beam and the portal frame along the out-of-plane depth $W$ given in \Cref{tab dimension}. We evaluate each example using five structured hexahedral meshes containing 500K, 1M, 2M, 4M, and 8M nodal CPs on the same NVIDIA A100 GPUs with 40~GB memory used for the 2D examples. The hardware allocation grows with mesh resolution: the 500K cases are trained on one GPU with two sub-slices, the 1M and 2M cases on two GPUs with two and four sub-slices per GPU, and the 4M and 8M cases on four GPUs with four and ten sub-slices per GPU, respectively. Since the allocation differs between cases, the training times reported for them are not directly comparable across resolutions. We study the effect of the number of GPUs and sub-slices at a fixed problem size separately in \Cref{sec res scal}. 

For all 2D and 3D benchmarks we examine how the number of CPs affects the optimization trajectory. Throughout, we consider an isotropic, linear-elastic solid with Young's modulus $E_0=70\,\mathrm{GPa}$, Poisson's ratio $\nu=0.25$, and yield strength $\sigma_{\mathrm{y0}}=100\,\mathrm{MPa}$. These properties represent an aluminum-like material. The initial design has a nearly uniform volume fraction of $0.6$ in all cases. Problem sizes are characterized by the number of nodal CPs in the background mesh. Each element contains one centroid CP and one corresponding local stress constraint such that $N_c=n_e$. The number of centroid CPs is slightly smaller than the number of nodal CPs. For brevity, we denote the nominal nodal CP counts by 125K, 250K, 500K, 1M, 2M, 4M, and 8M, where ``K" and ``M" denote thousand and million, respectively. The exact numbers of nodal and centroid CPs are reported in \Cref{tab mesh}. 

%=======================================
\begin{table*}[!b]
    \centering
    \renewcommand{\arraystretch}{1.5}
    \small
    \setlength\tabcolsep{4pt}
    \caption{\textbf{Mesh and feature-grid resolutions:}
    Numbers of nodal and centroid CPs of the meshes used for the 2D examples (top) and the 3D examples (bottom); the nominal sizes are given in the column headers. The number of centroid CPs equals the number of local stress constraints ($N_c = n_e$). PolyStress uses a single polygonal mesh per 2D example, whereas SNTO uses structured quadrilateral meshes in 2D and structured hexahedral meshes in 3D. All SNTO 2D cases are trained on a single GPU; the 125K, 250K, and 500K meshes are processed as one slice, while the 1M meshes are divided into two sub-slices. The last column lists the resolution $\Res$ of the PGCAN feature grid.}
    \resizebox{\textwidth}{!}{%
    \begin{tabular}{l|cc|cc|cc|cc|cc|c}
        \hline
        \multirow{2}{*}{\textbf{Example (2D)}}
        & \multicolumn{2}{c|}{\textbf{PolyStress}}
        & \multicolumn{2}{c|}{\textbf{SNTO 2D 125K}}
        & \multicolumn{2}{c|}{\textbf{SNTO 2D 250K}}
        & \multicolumn{2}{c|}{\textbf{SNTO 2D 500K}}
        & \multicolumn{2}{c|}{\textbf{SNTO 2D 1M}}
        & \textbf{$\Res$} \\
        \cline{2-12}
        & \textbf{Nodal CPs} & \textbf{Centroid CPs}
        & \textbf{Nodal CPs} & \textbf{Centroid CPs}
        & \textbf{Nodal CPs} & \textbf{Centroid CPs}
        & \textbf{Nodal CPs} & \textbf{Centroid CPs}
        & \textbf{Nodal CPs} & \textbf{Centroid CPs}
        & $(N_x^e, N_y^e)$ \\
        \hline
        L beam 
            & 251251 & 251000
            & 125153 & 124270
            & 250891 & 249641
            & 502704 & 500934
            & 1006008 & 1003503
            & (24, 24) \\
        Crack 
            & 251695 & 250632
            & 125900 & 125148
            & 251525 & 250462
            & 501501 & 500000
            & 1002510 & 1000387
            & (24, 48) \\
        Hook 
            & 495406 & 250000
            & 125219 & 124123
            & 251262 & 249704
            & 501870 & 499666
            & 1001114 & 997998
            & (24, 36) \\
        Portal frame 
            & 495557 & 250000
            & 125000 & 124261
            & 250790 & 249743
            & 500823 & 499343
            & 1000785 & 998693
            & (24, 24) \\
        \hline
        \multirow{2}{*}{\textbf{Example (3D)}}
        & \multicolumn{2}{c|}{\textbf{SNTO 3D 500K}}
        & \multicolumn{2}{c|}{\textbf{SNTO 3D 1M}}
        & \multicolumn{2}{c|}{\textbf{SNTO 3D 2M}}
        & \multicolumn{2}{c|}{\textbf{SNTO 3D 4M}}
        & \multicolumn{2}{c|}{\textbf{SNTO 3D 8M}}
        & \textbf{$\Res$} \\
        \cline{2-12}
        & \textbf{Nodal CPs} & \textbf{Centroid CPs}
        & \textbf{Nodal CPs} & \textbf{Centroid CPs}
        & \textbf{Nodal CPs} & \textbf{Centroid CPs}
        & \textbf{Nodal CPs} & \textbf{Centroid CPs}
        & \textbf{Nodal CPs} & \textbf{Centroid CPs}
        & $(N_x^e, N_y^e, N_z^e)$ \\
        \hline
        L beam 
            & 501207 & 469680
            & 1055835 & 1003912
            & 2050125 & 1968128
            & 4023125 & 3893760
            & 8027859 & 7821000
            & (36, 36, 5) \\
        Portal frame 
            & 501009 & 472736
            & 1018451 & 972832
            & 2012615 & 1941408
            & 4037629 & 3924720
            & 8027481 & 7848560
            & (36, 36, 5) \\
        \hline
    \end{tabular}
    }
    \label{tab mesh}
\end{table*}
%no nodes and elements, Ks Cp_n,c
%=======================================

\subsection{2D Benchmarks with Comparison Against an FE-based Method}\label{sec res 2D}
\begin{figure*}[!t]
    \centering
    \includegraphics[width = \textwidth]{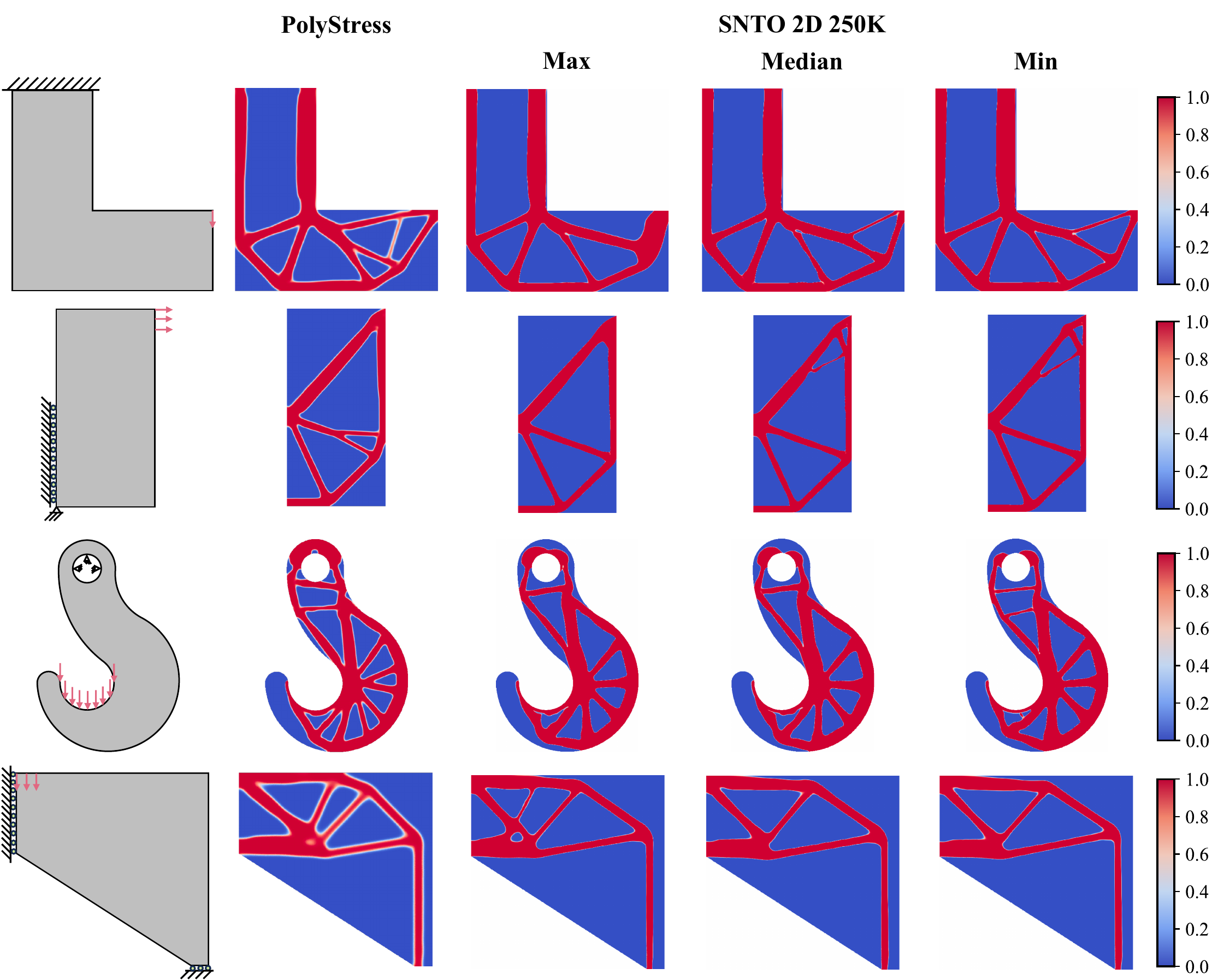} 
    \caption{\textbf{Comparison of optimized 2D topologies obtained by PolyStress and SNTO:}
    The left column shows the problem geometry for each benchmark, followed by the PolyStress solution (the second column) and three SNTO solutions from the five runs with 250K nodal CPs corresponding to the maximum, median, and minimum final volume fraction.
    The density field is plotted from void (blue) to solid material (red), with the same color scale used across all cases.
    The comparison shows that SNTO recovers physically interpretable geometries containing corners and cracks while producing designs with reduced gray regions compared to PolyStress. With SNTO, the density field can be visualized at any resolution, which is set here to the centroids corresponding to the 250K nodal CPs used in training.}
    \label{fig comp 2D}
\end{figure*}

For a direct comparison between SNTO and PolyStress on the 2D examples, we adopt the same continuation scheme for the SIMP exponent and the sharpness parameter in both frameworks. We keep the augmented Lagrangian parameters at the default values for PolyStress \cite{giraldo_polystress_2021} and set the number of outer iterations to $100$. \Cref{fig comp 2D} shows the median PolyStress design of each example next to the SNTO designs with the largest, median, and smallest final volume fraction for the 250K case. 
We observe that both methods produce well-defined load-bearing frames on all four examples but differ notably in how they relieve stress concentrations and remove redundant material.
In the L beam, PolyStress rounds the re-entrant corner more locally whereas SNTO pulls the column inward to form a smoother and longer curve to alleviate stress localization. 
For the crack example, our design with the maximum volume fraction is similar to the PolyStress design. Our designs with the median and minimum volume fractions differ from it mainly in how the structure is reinforced near the loading region: PolyStress relies on fewer and thicker members, whereas SNTO distributes the load through a finer, truss-like arrangement of members.
We observe the largest design disparity in the hook example where PolyStress fills the head around the fixed hole more densely compared to SNTO. As demonstrated in \Cref{sec res 2D const}, the stress on top of the head is well below the limit so SNTO effectively removes material in this area to achieve a smaller volume fraction without violating stress constraints. In contrast, the PolyStress design exhibits unnecessary redundancy. 
Portal frame designs remain comparable across both frameworks, with slight differences in the interior member sizes and layout.

%========================================
%=======================================

%=======================================
\begin{table*}[!t]
    \centering
    \renewcommand{\arraystretch}{1.5}
    \footnotesize
    \setlength\tabcolsep{6pt}
    \caption{\textbf{Median volume fraction, gray area fraction, and wall-clock runtime of the optimized 2D topologies:}
    For each example, the volume fraction of the final design, the gray area fraction (fraction of centroid CPs with $0.1 < \rho < 0.9$), and the wall-clock runtime in seconds are reported for PolyStress and for SNTO with 125K, 250K, 500K, and 1M nodal CPs. Each value is the median over 3 independent runs for PolyStress and 5 for SNTO.}
    \resizebox{\textwidth}{!}{%
    \begin{tabular}{l|l|c|c|c|c|c}
        \hline
        \textbf{Example}
        & \textbf{Metric}
        & \textbf{PolyStress 2D}
        & \textbf{SNTO 2D 125K}
        & \textbf{SNTO 2D 250K}
        & \textbf{SNTO 2D 500K}
        & \textbf{SNTO 2D 1M} \\
        \hline

        \multirow{3}{*}{L beam 2D}
        & Volume fraction & 0.365 & 0.341 & 0.349 & 0.349 & 0.343 \\
        & Gray fraction   & 0.087 & 0.027 & 0.035 & 0.026 & 0.023 \\
        & Runtime (sec)     & 2435 & 1432 & 2722 & 5221 & 10705 \\
        \hline

        \multirow{3}{*}{Crack 2D}
        & Volume fraction & 0.258 & 0.244 & 0.247 & 0.241 & 0.248 \\
        & Gray fraction   & 0.054 & 0.021 & 0.017 & 0.012 & 0.015 \\
        & Runtime (sec)     & 2559 & 1276 & 2341 & 4423 & 8758 \\
        \hline

        \multirow{3}{*}{Hook 2D}
        & Volume fraction & 0.579 & 0.486 & 0.505 & 0.507 & 0.505 \\
        & Gray fraction   & 0.094 & 0.014 & 0.025 & 0.027 & 0.023 \\
        & Runtime (sec)     & 6789  & 1650 & 3316 & 6424 & 12621 \\
        \hline

        \multirow{3}{*}{Portal frame 2D}
        & Volume fraction & 0.272 & 0.246 & 0.236 & 0.241 & 0.246 \\
        & Gray fraction   & 0.128 & 0.043 & 0.030 & 0.031 & 0.055 \\
        & Runtime (sec)     & 9023  & 1411 & 2632 & 5015 & 9812 \\
        \hline

    \end{tabular}
    }
    \label{tab 2D obj}
\end{table*}
%=======================================

\Cref{fig comp 2D} also highlights a clear difference in the solid--void transitions produced by PolyStress and SNTO. Our designs exhibit sharp interfaces and near binary densities while PolyStress produces noticeable bands of intermediate density, particularly for the L beam and portal frame. We quantify this behavior in \Cref{tab 2D obj} using the gray fraction defined as the fraction of centroid CPs with densities between $0.1$ and $0.9$. Across all examples, the gray fractions obtained with SNTO are two to seven times smaller than those obtained with PolyStress, ranging from $0.01$ to $0.06$ for SNTO and from $0.05$ to $0.13$ for PolyStress across the considered mesh resolutions. SNTO also consistently produces designs with lower volume fractions while satisfying the stress constraints as discussed in \Cref{sec res 2D const}. Bearing in mind the hardware differences, our SNTO framework matches the PolyStress wall-clock runtime at 250K nodal CPs for the crack and L beam and is approximately two to three times faster for the hook and portal frame examples. 

\Cref{tab 2D obj} also provides an initial assessment of SNTO's sensitivity to the number of nodal CPs. Across the four cases ranging from 125K to 1M nodal CPs, the volume and gray area fractions vary by at most $0.02$ and $0.03$, respectively. These values match per-seed variations so we can consider the final designs as nodal CP resolution-independent over this range. However, the runtime scales proportionally with the number of nodal CPs as expected from the assembly of the loss in \Cref{sec metd num}. Below, we further examine the effects of the number of nodal CPs on the optimization trajectory, final design, and stress violations, respectively.

\subsubsection{Topology Evolution During Optimization}\label{sec res 2D comp}
%========================================
\begin{figure*}[!t]
    \centering
    \begin{subfigure}[t]{\textwidth}
        \includegraphics[width=0.99\linewidth]{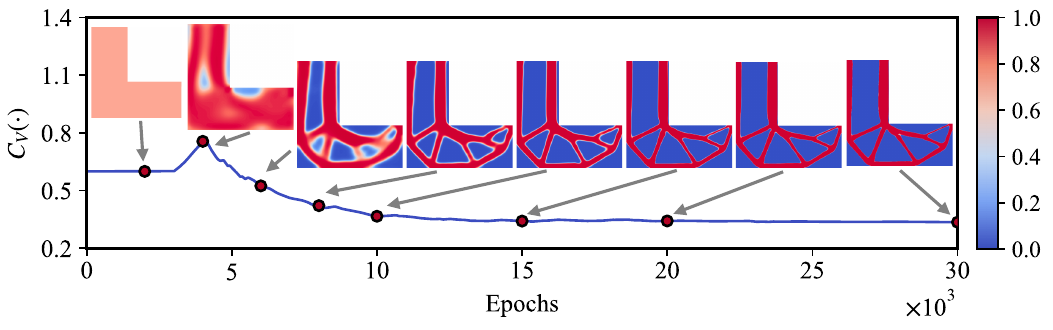}
        \vspace{-15em}
        \captionsetup{justification=raggedright, singlelinecheck=false, skip=-1.5pt, position=top,labelfont=subtxt, textfont=subtxt}
        \caption[]{}
        \label{fig top evo2d 1}
    \end{subfigure}
    \par\addvspace{-0.5em}
    \begin{subfigure}[t]{\textwidth}
        \includegraphics[width=0.99\linewidth]{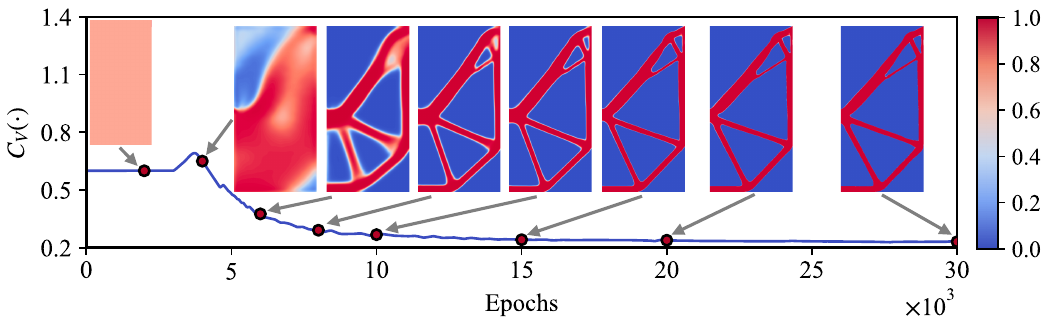}
        \vspace{-15em}
        \captionsetup{justification=raggedright, singlelinecheck=false, skip=-1.5pt, position=top,labelfont=subtxt, textfont=subtxt}
        \caption[]{}
        \label{fig top evo2d 2}
    \end{subfigure}
    \par\addvspace{-0.5em}
    \begin{subfigure}[t]{\textwidth}
        \includegraphics[width=0.99\linewidth]{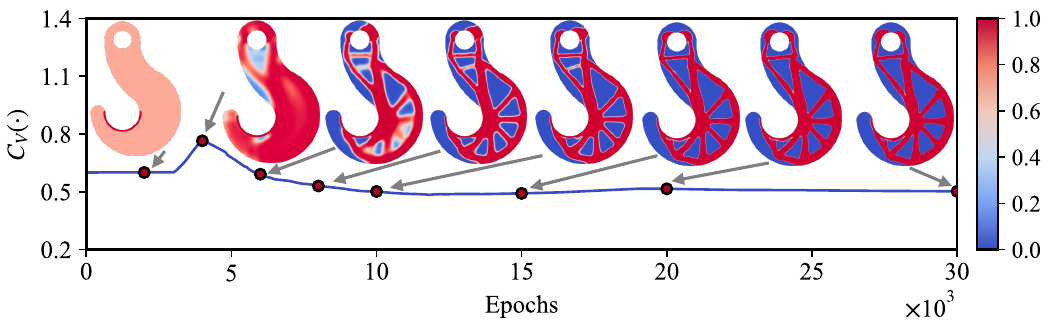}
        \vspace{-15em}
        \captionsetup{justification=raggedright, singlelinecheck=false, skip=-1.5pt, position=top,labelfont=subtxt, textfont=subtxt}
        \caption[]{}
        \label{fig top evo2d 3}
    \end{subfigure}
    \par\addvspace{-0.5em}
    \begin{subfigure}[t]{\textwidth}
        \includegraphics[width=0.99\linewidth]{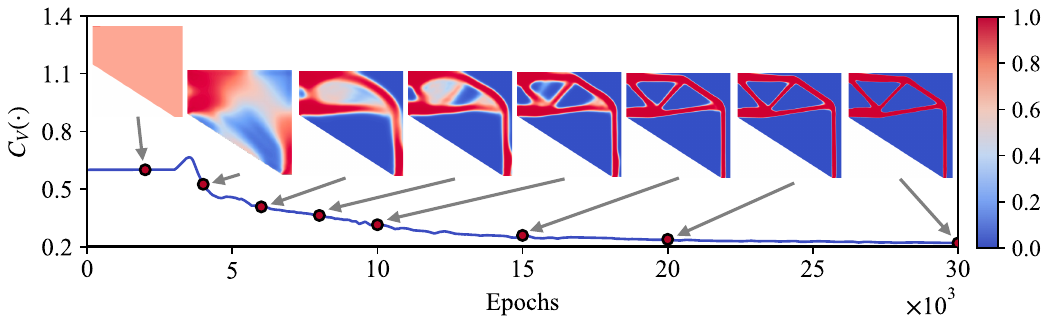}
        \vspace{-15em}
        \captionsetup{justification=raggedright, singlelinecheck=false, skip=-1.5pt, position=top,labelfont=subtxt, textfont=subtxt}
        \caption[]{}
        \label{fig top evo2d 4}
    \end{subfigure}
    \par\addvspace{-0.5em}
    \caption{\textbf{Evolution of the volume fraction and the selected topology during SNTO training:}
    Volume fraction histories and density snapshots are shown for the \textbf{(a)} L beam, \textbf{(b)} crack, \textbf{(c)} hook, and \textbf{(d)} portal frame benchmarks with 250K nodal CPs.
    Red markers indicate the epochs at which the intermediate designs are sampled; the first marker lies in the warm-up stage, the second at the peak of the volume fraction, and the last at the end of training.
    The snapshots show the transition from a uniform initial design to a diffuse layout to a crisp binary structure as the volume fraction decreases and settles.}
    \label{fig top evo2d}
\end{figure*}
%=======================================
We visualize the topology evolution during training for all 2D examples in \Cref{fig top evo2d} with the color bar distinguishing solid from void regions. The four cases share a common trend corresponding to the curriculum training scheme described in \Cref{sec metd train}. We initialize the density field at $\rho \approx 0.6$ everywhere in the domain and freeze the design network during the first $3{,}000$ epochs to allow the displacement and adjoint fields to reach equilibrium. 
As the density update begins, widespread stress violations activate the penalty term $\omega_C\mathcal{P}^{(a)}$, which adds material to reduce the stresses. The volume fraction consequently increases to a peak of $0.65$--$0.75$. As the stress violations diminish, $C_V(\cdot)$ becomes increasingly influential and drives the progressive removal of material from regions where the stress constraints are inactive.
The snapshots at the peak confirm that stress-relieving topological features emerge first while the domain remains mostly gray. For example, the L beam's corner fillet and the crack-tip cap are already present at $4{,}000$ epochs. Therefore, the stress constraints
primarily outline the design and the volume fraction objective trims excess material.

The volume fraction decreases monotonically, then plateaus after the peak, with final topologies established by roughly $10{,}000$ epochs across all examples. The crack design settles earliest, whereas the hook design continues to evolve beyond $15{,}000$ epochs as the optimizer dissolves a non-load-bearing member in the top neck region. The remaining epochs only refine the design rather than change its topology. The solid-void interface becomes crisper as the projection sharpness $\beta$ and the exponent $p$ reach their terminal values.

\begin{figure*}[!t]
    \centering
    \begin{subfigure}[t]{0.32\textwidth}
        \includegraphics[width=\linewidth]{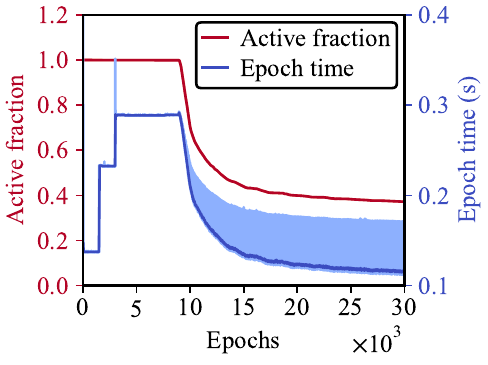}
        \vspace{-14.5em}
        \captionsetup{justification=raggedright, singlelinecheck=false, skip=-6.5pt, position=top,labelfont=subtxt, textfont=subtxt}
        \caption[]{}
        \label{fig hist evo 1}
    \end{subfigure}
    \hfill
    \begin{subfigure}[t]{0.32\textwidth}
        \includegraphics[width=\linewidth]{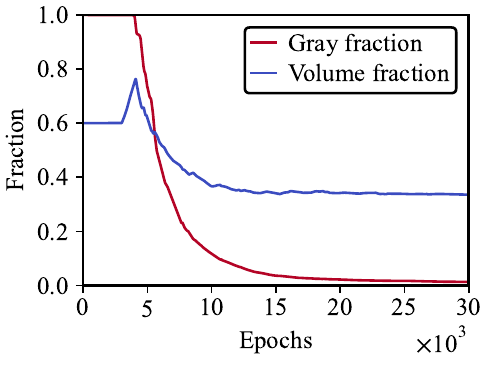}
        \vspace{-14.5em}
        \captionsetup{justification=raggedright, singlelinecheck=false, skip=-3.5pt, position=top,labelfont=subtxt, textfont=subtxt}
        \caption[]{}
        \label{fig hist evo 2}
    \end{subfigure}
    \hfill
    \begin{subfigure}[t]{0.32\textwidth}
        \includegraphics[width=\linewidth]{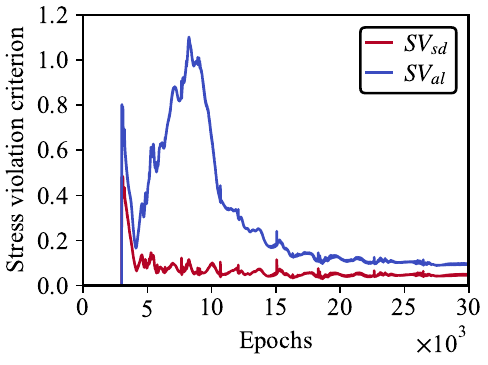}
        \vspace{-14.5em}
        \captionsetup{justification=raggedright, singlelinecheck=false, skip=-3.5pt, position=top,labelfont=subtxt, textfont=subtxt}
        \caption[]{}
        \label{fig hist evo 3}
    \end{subfigure}

    \vspace{0.6em}

    \begin{subfigure}[t]{0.32\textwidth}
        \includegraphics[width=\linewidth]{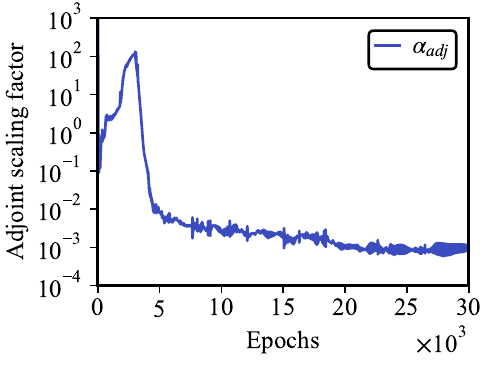}
        \vspace{-14.5em}
        \captionsetup{justification=raggedright, singlelinecheck=false, skip=-6.5pt, position=top,labelfont=subtxt, textfont=subtxt}
        \caption[]{}
        \label{fig hist evo 4}
    \end{subfigure}
    \hfill
    \begin{subfigure}[t]{0.32\textwidth}
        \includegraphics[width=\linewidth]{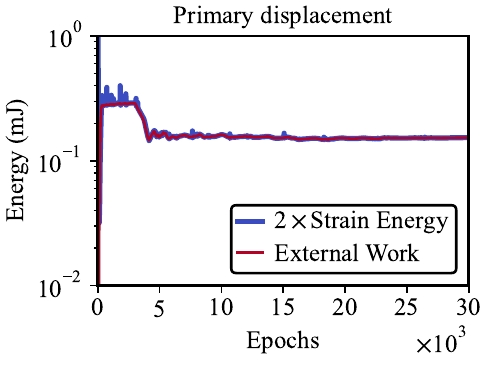}
        \vspace{-14.5em}
        \captionsetup{justification=raggedright, singlelinecheck=false, skip=-3.5pt, position=top,labelfont=subtxt, textfont=subtxt}
        \caption[]{}
        \label{fig hist evo 5}
    \end{subfigure}
    \hfill
    \begin{subfigure}[t]{0.32\textwidth}
        \includegraphics[width=\linewidth]{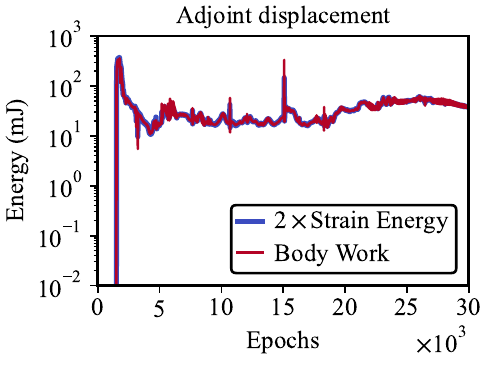}
        \vspace{-14.5em}
        \captionsetup{justification=raggedright, singlelinecheck=false, skip=-3.5pt, position=top,labelfont=subtxt, textfont=subtxt}
        \caption[]{}
        \label{fig hist evo 6}
    \end{subfigure}
    \caption{\textbf{Training histories of SNTO for the L beam example with 250K nodal CPs:}
    \textbf{(a)} active element fraction and wall-clock time per epoch;
    \textbf{(b)} volume fraction and gray area fraction;
    \textbf{(c)} stress-violation measures $SV_{sd}$ and $SV_{al}$;
    \textbf{(d)} adjoint scaling factor $\alpha_{adj}$ of \Cref{eq alpha adj};
    \textbf{(e)} twice the strain energy of the displacement field compared with the external work;
    \textbf{(f)} twice the strain energy of the adjoint field compared with the body work.
    At equilibrium the two quantities in (e) and (f) coincide (Clapeyron's theorem), so their agreement is a direct check that both networks satisfy \Cref{eq LM,eq LMa} throughout the optimization.}
    \label{fig hist evo}
\end{figure*}

We track additional metrics during training that describe how the state, the constraints, and the cost of an epoch progress. We present them only for the L beam with 250K nodal CPs in \Cref{fig hist evo} since the histories are qualitatively similar for all examples. The cost of each training epoch decreases as the design converges as we only keep active centroid CPs with densities $\rho>10^{-3}$. From \CrefSub{fig hist evo}{fig hist evo 1}, all elements remain active until approximately $9{,}000$ epochs. The active element fraction then gradually decreases to approximately $0.40$ because elements with densities below the threshold are excluded from the loss evaluation. The wall-clock time per epoch decreases accordingly from $0.29\text{ s}$ to $0.12\text{ s}$. We note that the two increases in computational cost before $3{,}000$ epochs reflect the warm-up schedule described in \Cref{sec metd train}. During the first $1{,}500$ epochs, only the primal loss $\mathcal{L}_M(\cdot)$ is evaluated, resulting in the lowest computational cost. Introducing the adjoint loss $\mathcal{L}_M^{(a)}(\cdot)$ at $1{,}500$ epochs increases the cost, which reaches its maximum at $3{,}000$ epochs when the full loss function is activated.
\CrefSub{fig hist evo}{fig hist evo 2} shows that SNTO achieves near-binary designs well before training is over. We hold all CPs at intermediate densities (i.e., $\rho \in [0.1, 0.9]$) during the warm-up phase by freezing the density network. The gray fraction reduces quickly post warm-up phase, reaching approximately $0.1$ after only a third of the epochs. It then decreases gradually throughout the remainder of the optimization and reaches the final value of $0.035$ reported in \Cref{tab 2D obj}.

We assess the satisfaction of the local stress constraints during training using the two violation measures shown in \CrefSub{fig hist evo}{fig hist evo 3}.
Both measures are based on the von Mises stress defined in \Cref{eq svm} and identify centroid CPs at which the stress exceeds the tightened training limit $\sigma_y = \alpha_s\sigma_{\mathrm{y0}}$ (\Cref{sec metd train}).
The measure $SV_{sd}$ is the percentage of solid centroid CPs with $\rho\geq0.5$ that violate the stress limit. The measure $SV_{al}$ is the number of violating active centroid CPs normalized by $C_V \times n_e$, which represents the effective number of material-containing centroid CPs.
Since the stress constraints are activated only after the warm-up stage, we report $SV_{sd}$ and $SV_{al}$ beginning at epoch $3{,}000$. The value of $SV_{sd}$ decreases rapidly and then stabilizes at a small nonzero level. This residual is inherent to the formulation rather than an indication of incomplete convergence. The adjoint force in \Cref{eq disc_adj} is assembled only from active stress constraints, and therefore no stress sensitivity would remain if the active set became empty. The value of $SV_{al}$ can exceed one because its numerator includes all stress-violating active centroid CPs in both solid and low-density regions, whereas the denominator $C_V \times n_e$ represents the effective number of fully solid elements. Because low-density regions have little stiffness, they may develop large strains and contribute many violations when their stresses are evaluated using the solid-material modulus.

\CrefSub{fig hist evo}{fig hist evo 4} shows the evolution of the adjoint scaling factor $\alpha_{adj}$ defined in \Cref{eq alpha adj}. Its variation over five orders of magnitude during training highlights the importance of adjoint normalization. The scaling factor initially increases due to the severe stress violations in the uniform-density design. Once the density network begins training and the stress concentrations are reduced, $\alpha_{adj}$ decreases by four orders of magnitude and eventually stabilizes near $10^{-3}$. Without normalization, the magnitude of the field represented by the adjoint network would undergo similarly large changes over only a few thousand epochs, potentially destabilizing the force equilibrium in the coupled optimization. The factor $\alpha_{adj}$ controls the magnitude of the normalized adjoint field while preserving the true adjoint sensitivity used to update the design as discussed in \Cref{sec metd train}.

%========================================
%=======================================
We verify mechanical equilibrium with Clapeyron's theorem, which requires no FE solve: for a linear elastic field in equilibrium, twice the strain energy equals the work done by the applied load (\Cref{sec equ}). \Cref{fig hist evo} plots both quantities for the L beam. We observe that the primal curves coincide from the first few hundred warm-up epochs onward, staying in close agreement even while the design changes most rapidly between $3{,}000$ and $5{,}000$ epochs (\CrefSub{fig hist evo}{fig hist evo 5}). Similarly, the adjoint curves agree as soon as their corresponding network activates, though with more fluctuations (\CrefSub{fig hist evo}{fig hist evo 6}); the adjoint load shifts in spatial distribution from epoch to epoch as local constraints switch state. Nevertheless, both primal and adjoint networks maintain equilibrium throughout the entire training, which is the premise of our simultaneous formulation in \Cref{sec metd adj}.

\subsubsection{Mesh Sensitivity Analysis}\label{sec res 2D mesh}
%========================================
\begin{figure*}[!b]
    \centering
    \includegraphics[width = \textwidth]{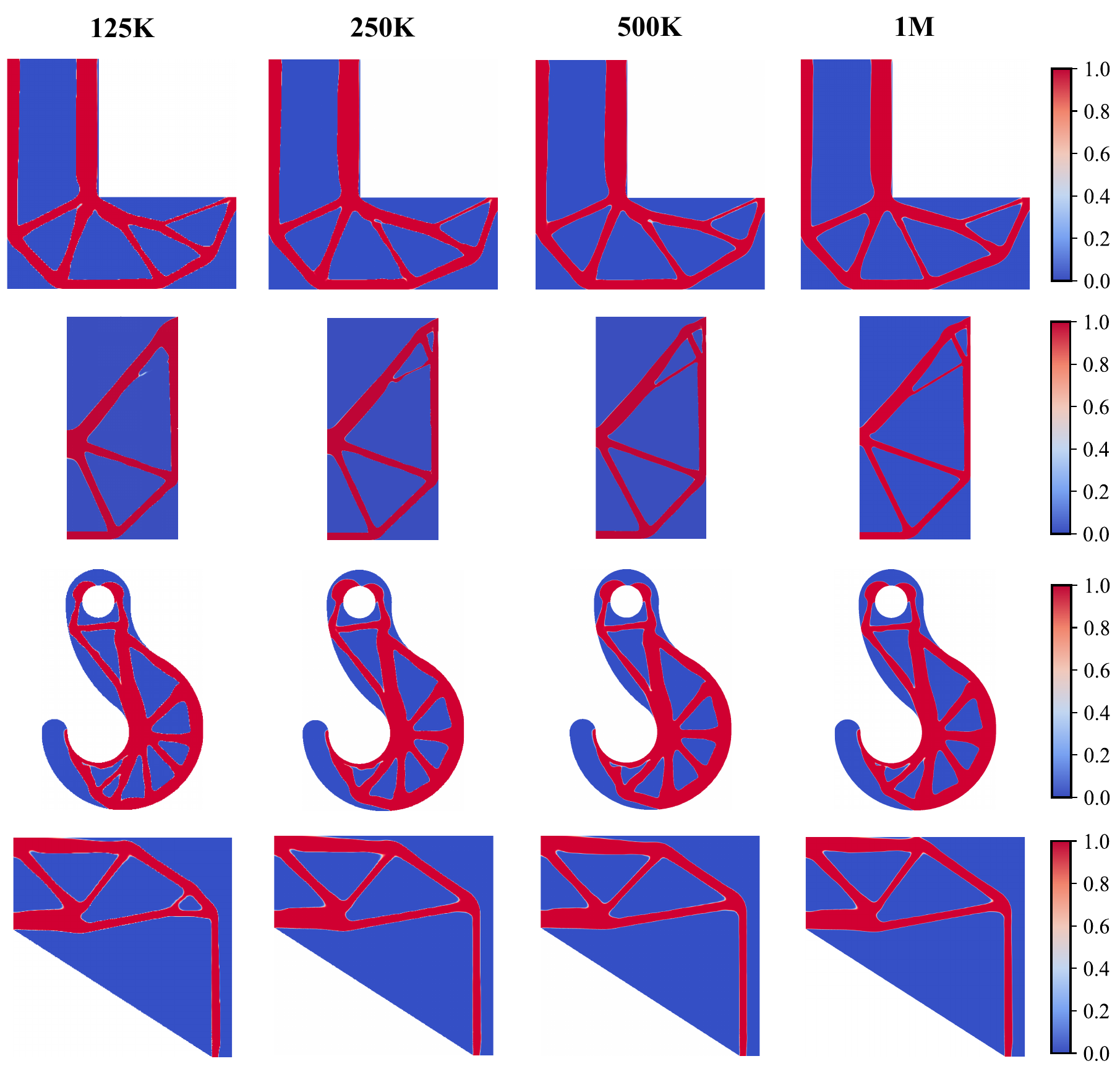} 
    \caption{\textbf{Mesh-resolution study for optimized 2D topologies:}
    Final density fields of the run with the median volume fraction are shown for the L beam, crack, hook, and portal frame benchmarks with 125K, 250K, 500K, and 1M nodal CPs.
    The consistent topologies across refinements indicate that SNTO is essentially mesh independent, preserving the main structural features as the number of local stress constraints increases by nearly an order of magnitude.
    The density scale is shared across all panels, with blue denoting void and red denoting solid material.
    The SNTO density fields are plotted at the centroids of the nodal CPs used in training, although they can be evaluated at any resolution.}
    \label{fig top mesh 2D}
\end{figure*}
%=======================================
We showed in \Cref{tab 2D obj} that the final volume and gray fractions achieved via SNTO are largely insensitive to the number of nodal CPs used during training. We now examine whether the optimized topologies exhibit similar consistency. \Cref{fig top mesh 2D} compares the density fields of the designs with the median volume fraction at each of the four nodal CP resolutions. The principal load-carrying features are preserved across all resolutions, including the fillet at the re-entrant corner of the L beam, the cap around the crack tip, the radial spokes of the hook, and the triangulated upper region of the portal frame. Small variations occur only in secondary members. In the crack example, the 125K case has fewer features near the top right corner than the other three cases. At 250K, the fine members in this region appear with rough and irregular boundaries, whereas the 500K and 1M cases show consistent topologies in the same regions. In the hook benchmark, the struts supporting the loaded rim are arranged differently at 125K resolution. Variations of the same kind are observed across random seeds at a fixed resolution, which suggests that they arise from the non-convexity of the optimization problem rather than from mesh dependence.
The observed mesh insensitivity is achieved through a different mechanism from that of conventional FE-based TO which relies on filtering. In SNTO, it arises from the continuous neural parameterization of the fields and as long as the numerical differentiation and integration are performed with sufficient accuracy, further increasing the number of CPs minimally affects the designed topology.

%========================================
\begin{figure*}[!b]
    \centering
    \begin{subfigure}[t]{\textwidth}
        \centering
        \includegraphics{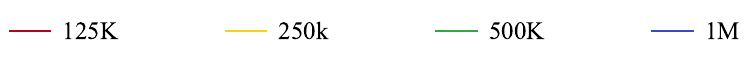}
    \end{subfigure}
    \begin{subfigure}[t]{0.32\textwidth}
        \includegraphics[width=\linewidth]{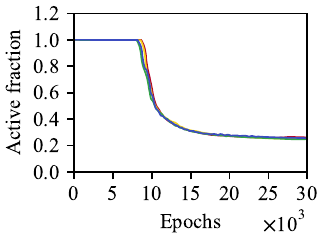}
        \vspace{-14em}
        \captionsetup{justification=raggedright, singlelinecheck=false, skip=-5.3pt, position=top,labelfont=subtxt, textfont=subtxt}
        \caption[]{}
        \label{fig hist mesh crack 2d 1}
    \end{subfigure}
    \begin{subfigure}[t]{0.32\textwidth}
        \includegraphics[width=\linewidth]{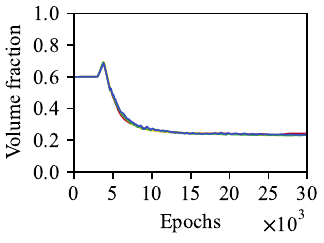}
        \vspace{-14em}
        \captionsetup{justification=raggedright, singlelinecheck=false, skip=-2.9pt, position=top,labelfont=subtxt, textfont=subtxt}
        \caption[]{}
        \label{fig hist mesh crack 2d 2}
    \end{subfigure}
    \begin{subfigure}[t]{0.32\textwidth}
        \includegraphics[width=\linewidth]{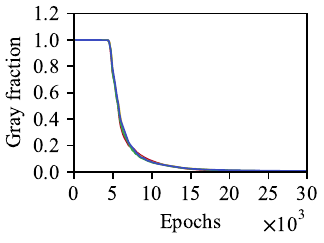}
        \vspace{-14em}
        \captionsetup{justification=raggedright, singlelinecheck=false, skip=-2.9pt, position=top,labelfont=subtxt, textfont=subtxt}
        \caption[]{}
        \label{fig hist mesh crack 2d 3}
    \end{subfigure}
    \begin{subfigure}[t]{0.32\textwidth}
        \includegraphics[width=\linewidth]{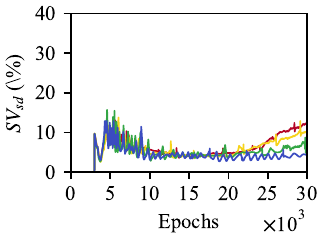}
        \vspace{-14em}
        \captionsetup{justification=raggedright, singlelinecheck=false, skip=-2.9pt, position=top,labelfont=subtxt, textfont=subtxt}
        \caption[]{}
        \label{fig hist mesh crack 2d 4}
    \end{subfigure}
    \begin{subfigure}[t]{0.32\textwidth}
        \includegraphics[width=\linewidth]{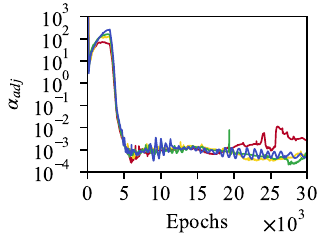}
        \vspace{-14em}
        \captionsetup{justification=raggedright, singlelinecheck=false, skip=-2.9pt, position=top,labelfont=subtxt, textfont=subtxt}
        \caption[]{}
        \label{fig hist mesh crack 2d 5}
    \end{subfigure}
    \begin{subfigure}[t]{0.32\textwidth}
        \includegraphics[width=\linewidth]{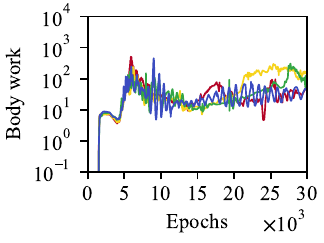}
        \vspace{-14em}
        \captionsetup{justification=raggedright, singlelinecheck=false, skip=-2.9pt, position=top,labelfont=subtxt, textfont=subtxt}
        \caption[]{}
        \label{fig hist mesh crack 2d 6}
    \end{subfigure}
    \caption{\textbf{Sensitivity of evolution histories with respect to number of nodal CPs using the Crack benchmark:}
    Evolution histories are compared for the runs with 125K, 250K, 500K, and 1M nodal CPs.
    The panels report \textbf{(a)} active element fraction, \textbf{(b)} volume fraction, \textbf{(c)} gray area fraction, \textbf{(d)} stress violation over the realized solid material, \textbf{(e)} adjoint scaling factor, and \textbf{(f)} body work contribution in the adjoint problem.
    The overlapping trends show that the training dynamics and constraint response remain consistent under mesh refinement.}
    \label{fig hist mesh crack 2d}
\end{figure*}
%=======================================

The training histories also exhibit limited sensitivity to mesh resolution. \Cref{fig hist mesh crack 2d} presents six monitored quantities for the crack example using 125K to 1M nodal CPs. The histories of the active element, volume, and gray area fractions are nearly indistinguishable across the four resolutions as shown in \CrefSubRange{fig hist mesh crack 2d}{fig hist mesh crack 2d 1}{fig hist mesh crack 2d 3}, respectively. The other three quantities depend directly on the local stress field and exhibit some differences across resolutions. 

The solid stress-violation measure $SV_{sd}$ in \CrefSub{fig hist mesh crack 2d}{fig hist mesh crack 2d 4} follows a similar trajectory at all four resolutions during the first two-thirds of training. Thereafter, the 125K and 250K curves gradually increase, while the 500K and 1M curves remain near their plateaus. At this stage, the design contains fine members and the density projection has reached its full sharpness, which produces the steepest stress gradients along the member boundaries. The coarser CPs resolve these gradients less accurately and so a larger fraction of the solid centroid CPs is counted as violating the limit. The 125K curve also shows fluctuations that the others do not with its corresponding adjoint scaling factor $\alpha_{adj}$ in \CrefSub{fig hist mesh crack 2d}{fig hist mesh crack 2d 5} showing the same trend: after $20{,}000$ epochs, its $\alpha_{adj}$ has jumps of approximately one order of magnitude while the other three curves remain smooth. Both features coincide with the removal of small members, such as the brace near the top right corner noted in \Cref{fig top mesh 2D}. Each removal redistributes the stress in its neighborhood and abruptly changes the set of active stress constraints. On a coarse mesh, a member represented by only a few centroid CPs disappears abruptly, whereas finer meshes allow the same member to thin gradually. The adjoint body work in \CrefSub{fig hist mesh crack 2d}{fig hist mesh crack 2d 6} shows no systematic dependence on resolution, confirming that the adjoint field reaches equilibrium in all cases.

In summary, we find that mesh refinement improves the stability of the optimization rather than altering the design. At 500K CPs and above, the late drift in $SV_{sd}$ and the fluctuations in $\alpha_{adj}$ subside, while the topology and the design metrics remain consistent with those obtained at 125K. This stability comes at a price, since the training time grows in proportion to the number of CPs (\Cref{tab 2D obj}). The other three 2D examples show the same trend; their optimization histories are given in the Appendix, see \Cref{app hist mesh}.

\subsubsection{Stress Constraint Verification}\label{sec res 2D const}

%========================================
\begin{figure*}[!t]
    \centering
    \begin{subfigure}[t]{0.48\textwidth}
        \includegraphics[width=\linewidth]{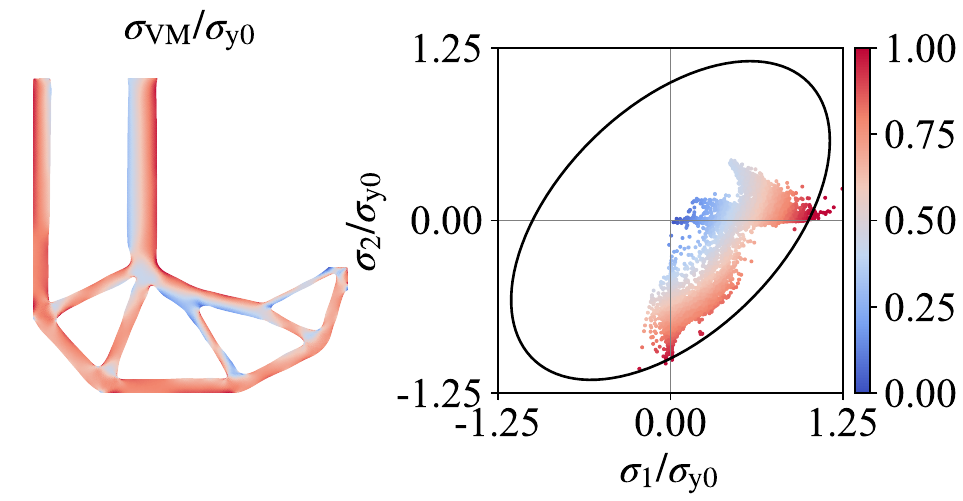}
        \vspace{-13.5em}
        \captionsetup{justification=raggedright, singlelinecheck=false, skip=-6.5pt, position=top,labelfont=subtxt, textfont=subtxt}
        \caption[]{}
        \label{fig stress vol 2d 1}
    \end{subfigure}
    \begin{subfigure}[t]{0.48\textwidth}
        \includegraphics[width=\linewidth]{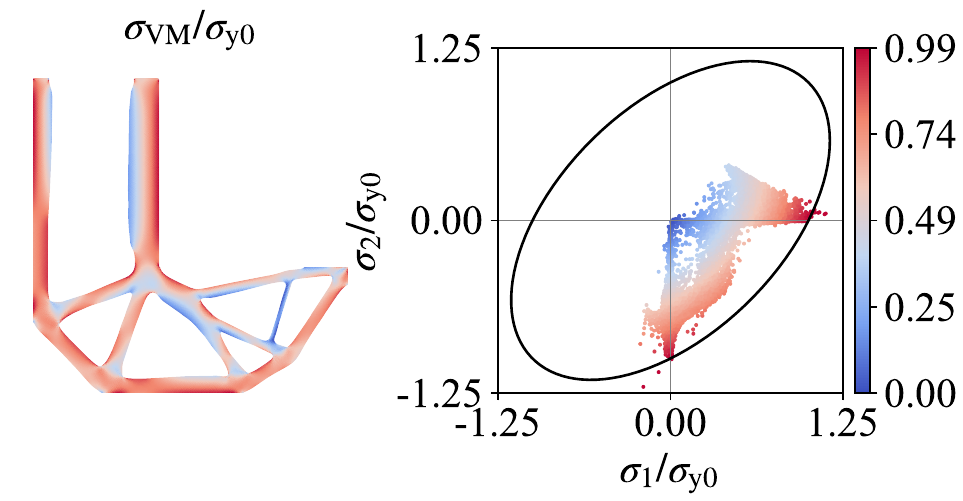}
        \vspace{-13.5em}
        \captionsetup{justification=raggedright, singlelinecheck=false, skip=-3.5pt, position=top,labelfont=subtxt, textfont=subtxt}
        \caption[]{}
        \label{fig stress vol 2d 2}
    \end{subfigure}
    \begin{subfigure}[t]{0.48\textwidth}
        \includegraphics[width=\linewidth]{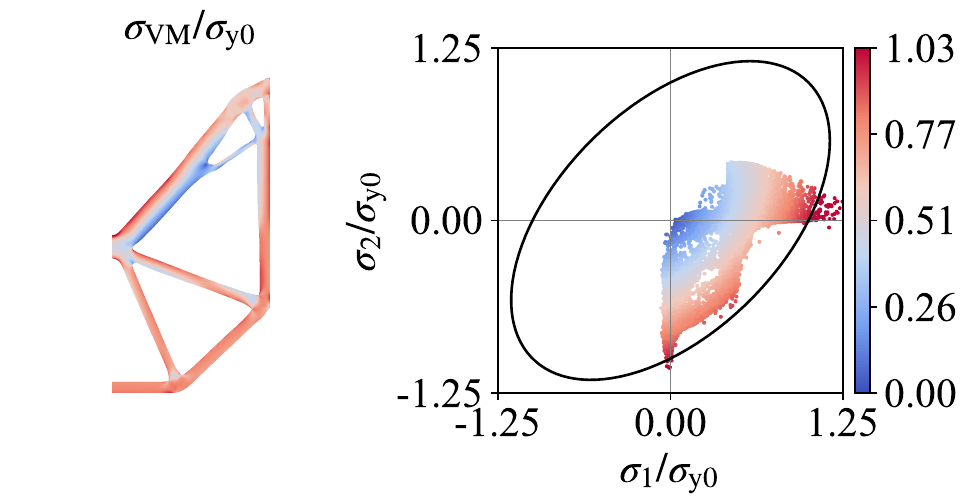}
        \vspace{-13.5em}
        \captionsetup{justification=raggedright, singlelinecheck=false, skip=-3.5pt, position=top,labelfont=subtxt, textfont=subtxt}
        \caption[]{}
        \label{fig stress vol 2d 3}
    \end{subfigure}
    \begin{subfigure}[t]{0.48\textwidth}
        \includegraphics[width=\linewidth]{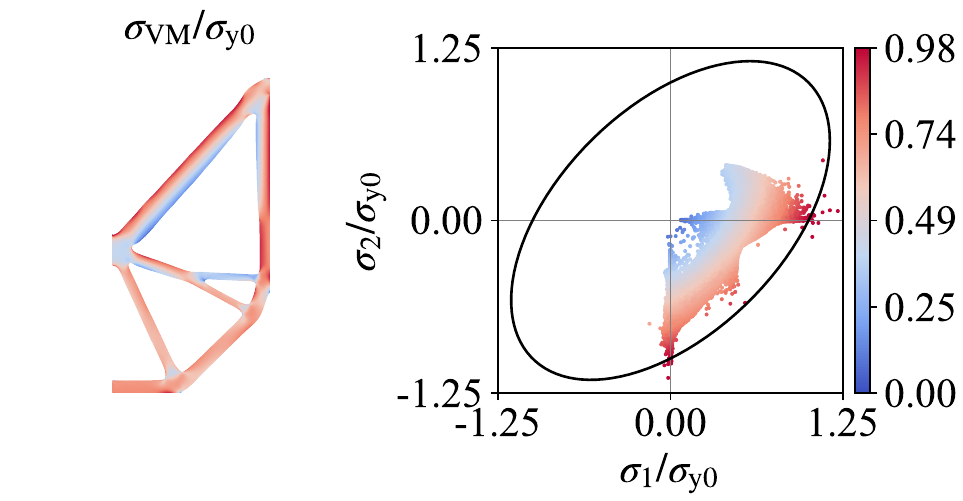}
        \vspace{-13.5em}
        \captionsetup{justification=raggedright, singlelinecheck=false, skip=-3.5pt, position=top,labelfont=subtxt, textfont=subtxt}
        \caption[]{}
        \label{fig stress vol 2d 4}
    \end{subfigure}
    \begin{subfigure}[t]{0.48\textwidth}
        \includegraphics[width=\linewidth]{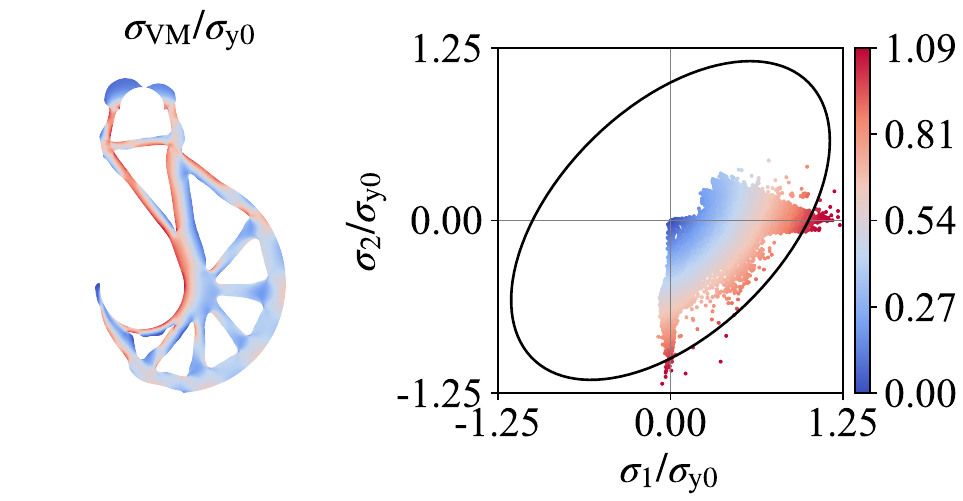}
        \vspace{-13.5em}
        \captionsetup{justification=raggedright, singlelinecheck=false, skip=-3.5pt, position=top,labelfont=subtxt, textfont=subtxt}
        \caption[]{}
        \label{fig stress vol 2d 5}
    \end{subfigure}
    \begin{subfigure}[t]{0.48\textwidth}
        \includegraphics[width=\linewidth]{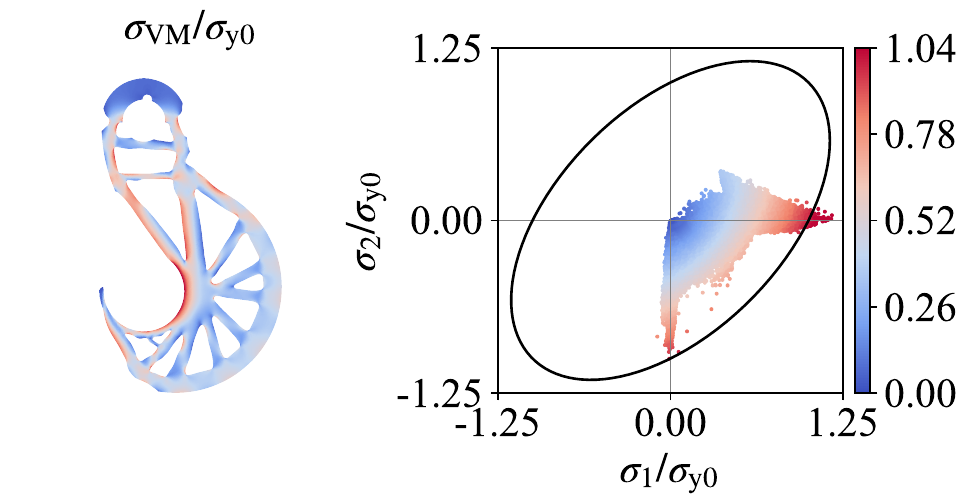}
        \vspace{-13.5em}
        \captionsetup{justification=raggedright, singlelinecheck=false, skip=-3.5pt, position=top,labelfont=subtxt, textfont=subtxt}
        \caption[]{}
        \label{fig stress vol 2d 6}
    \end{subfigure}
    \begin{subfigure}[t]{0.48\textwidth}
        \includegraphics[width=\linewidth]{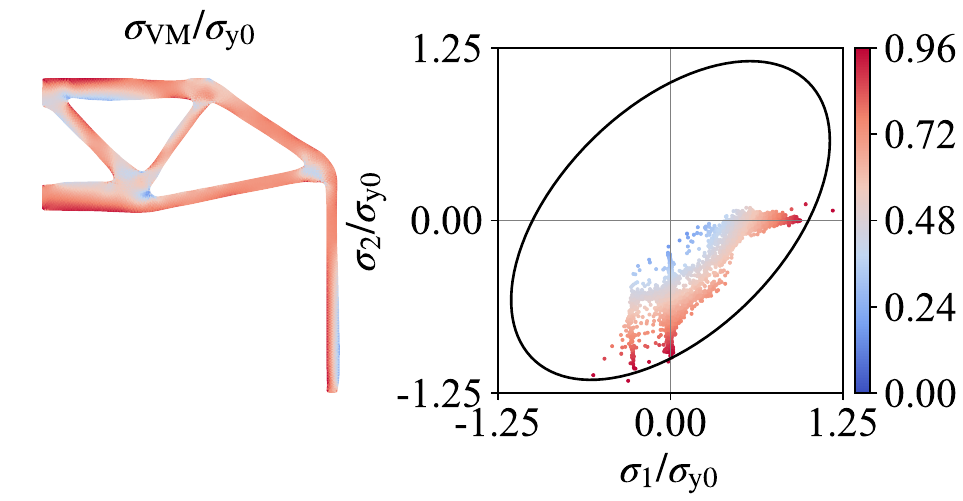}
        \vspace{-13.5em}
        \captionsetup{justification=raggedright, singlelinecheck=false, skip=-3.5pt, position=top,labelfont=subtxt, textfont=subtxt}
        \caption[]{}
        \label{fig stress vol 2d 7}
    \end{subfigure}
    \begin{subfigure}[t]{0.48\textwidth}
        \includegraphics[width=\linewidth]{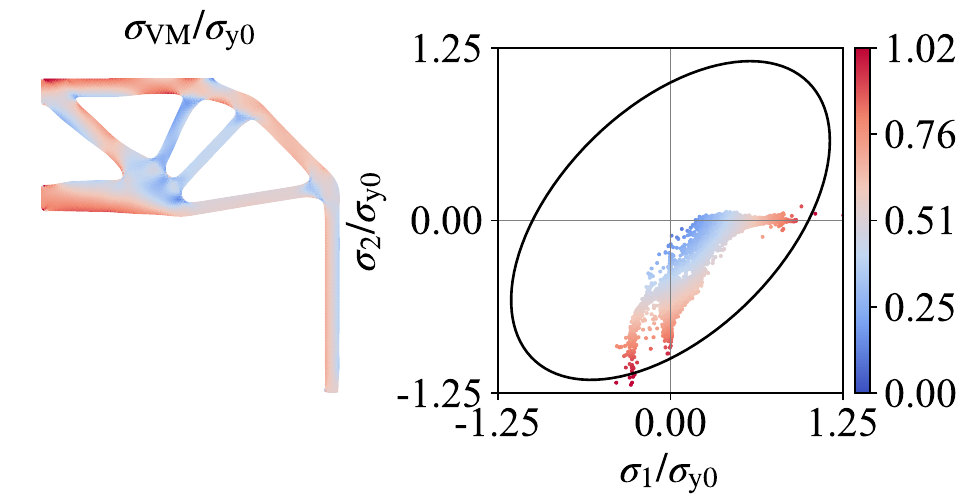}
        \vspace{-13.5em}
        \captionsetup{justification=raggedright, singlelinecheck=false, skip=-3.5pt, position=top,labelfont=subtxt, textfont=subtxt}
        \caption[]{}
        \label{fig stress vol 2d 8}
    \end{subfigure}
    \caption{\textbf{Stress-constraint verification of optimized 2D designs:} Distributions of normalized von Mises stress, $\sigma_{\mathrm{VM}}/\sigma_{\mathrm{y0}}$, with the corresponding normalized principal-stress states are shown for SNTO with 250K nodal CPs and PolyStress on the left and right, respectively. The stress is computed with FEniCS on the thresholded and remeshed designs. In the scatter plots, each point is the principal-stress state of one element, the black ellipse is the von Mises yield envelope, and the colors indicate the local stress ratio.}
    \label{fig stress vol 2d}
\end{figure*}
%=======================================
We validate the final designs of both SNTO and PolyStress with an independent FE analysis. We extract the solid boundary as the $\rho = 0.5$ contour of the density field and then smooth it. The resulting geometry is remeshed with linear triangular elements and the linear elastic problem is solved with the open-source code FEniCS \cite{fenics_website}. \Cref{fig stress vol 2d} illustrates the element-wise von Mises stress normalized by $\sigma_{\mathrm{y0}}$ for the SNTO (with 250K CPs) and PolyStress designs on the left and right, respectively. For both methods the run with the median volume fraction is shown. The color scale is capped at the $99.9$th percentile of the stress ratio so that a few extreme elements do not hide the rest of the stress field. Beside each stress field, the principal stresses of all elements are plotted in the normalized principal-stress plane. The von Mises criterion forms the ellipse $\sigma_1^2 - \sigma_1\sigma_2 + \sigma_2^2 = \sigma_{\mathrm{y0}}^2$ in this plane. Points outside the ellipse violate the stress constraint and each point is colored by the stress ratio of its element.

Both methods satisfy the stress constraint almost everywhere and only a small fraction of the elements lies outside the ellipse. We compare the two methods by the $99.9$th percentile of the stress ratio which is the upper limit of each color scale in \Cref{fig stress vol 2d}. PolyStress reaches a slightly lower value in the L beam, the crack and the hook while SNTO reaches a lower value in the portal frame. These differences are small and do not exceed about $0.05$. Note that SNTO reaches these stress levels with a lower volume fraction in all four examples (see \Cref{tab 2D obj}). The residual stress violations are intrinsic to both formulations rather than a defect of either. Both methods rely on the active constraints for the design update where the adjoint force of SNTO in \Cref{eq disc_adj} and the multiplier updates of PolyStress are both assembled from the active set. The few violating elements accordingly sit at the tension and compression segments of the point cloud. These elements are in the nearly uniaxial states typical of the thin members that carry the load.

The stress fields show that both methods place material where the load path demands it and that the lighter SNTO designs result from removing idle material. In the L beam and crack examples in \CrefSubRange{fig stress vol 2d}{fig stress vol 2d 1}{fig stress vol 2d 4}, the stress concentrates at the geometric singularities of the design domain which are, respectively, the rounded re-entrant corner and the crack-tip cap along with the diagonal beam that carries the load. In the hook benchmark in \CrefSub{fig stress vol 2d}{fig stress vol 2d 5} and \CrefSub{fig stress vol 2d}{fig stress vol 2d 6}, the load path runs along the inner curve where the bending moment is largest and the stress is high. The region around the upper clamped hole is nearly unstressed. PolyStress nevertheless closes a ring of material there whereas SNTO supports the hole with two separate clamps. This difference accounts for a large part of the gap in final volume fractions reported in \Cref{tab 2D obj} where SNTO clearly reaches the lower value. The portal frame case in \CrefSub{fig stress vol 2d}{fig stress vol 2d 7} and \CrefSub{fig stress vol 2d}{fig stress vol 2d 8} differs in the same way since PolyStress divides the frame into three cells whereas SNTO forms two cells and uses less material.

%========================================
\begin{figure*}[!t]
    \centering
    \begin{subfigure}[t]{0.23\textwidth}
        \includegraphics[width=\linewidth]{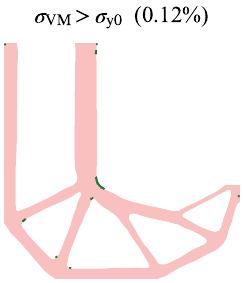}
        \vspace{-14.2em}
        \captionsetup{justification=raggedright, singlelinecheck=false, skip=-6.5pt, position=top,labelfont=subtxt, textfont=subtxt}
        \caption[]{}
        \label{fig stress vol loc 1}
    \end{subfigure}
    \begin{subfigure}[t]{0.23\textwidth}
        \includegraphics[width=\linewidth]{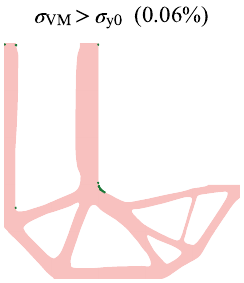}
        \vspace{-14.2em}
        \captionsetup{justification=raggedright, singlelinecheck=false, skip=-3.5pt, position=top,labelfont=subtxt, textfont=subtxt}
        \caption[]{}
        \label{fig stress vol loc 2}
    \end{subfigure}
    \begin{subfigure}[t]{0.23\textwidth}
        \includegraphics[width=\linewidth]{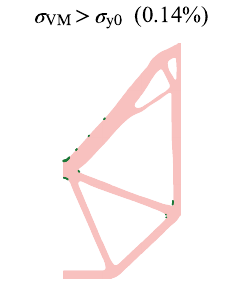}
        \vspace{-14.2em}
        \captionsetup{justification=raggedright, singlelinecheck=false, skip=-3.5pt, position=top,labelfont=subtxt, textfont=subtxt}
        \caption[]{}
        \label{fig stress vol loc 3}
    \end{subfigure}
    \begin{subfigure}[t]{0.23\textwidth}
        \includegraphics[width=\linewidth]{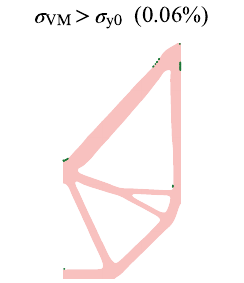}
        \vspace{-14.2em}
        \captionsetup{justification=raggedright, singlelinecheck=false, skip=-3.5pt, position=top,labelfont=subtxt, textfont=subtxt}
        \caption[]{}
        \label{fig stress vol loc 4}
    \end{subfigure}
    \begin{subfigure}[t]{0.23\textwidth}
        \includegraphics[width=\linewidth]{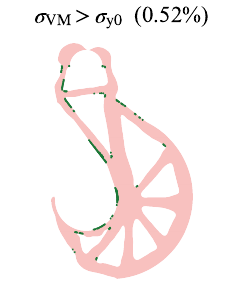}
        \vspace{-14.2em}
        \captionsetup{justification=raggedright, singlelinecheck=false, skip=-3.5pt, position=top,labelfont=subtxt, textfont=subtxt}
        \caption[]{}
        \label{fig stress vol loc 5}
    \end{subfigure}
    \begin{subfigure}[t]{0.23\textwidth}
        \includegraphics[width=\linewidth]{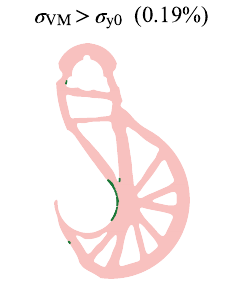}
        \vspace{-14.2em}
        \captionsetup{justification=raggedright, singlelinecheck=false, skip=-3.5pt, position=top,labelfont=subtxt, textfont=subtxt}
        \caption[]{}
        \label{fig stress vol loc 6}
    \end{subfigure}
    \begin{subfigure}[t]{0.23\textwidth}
        \includegraphics[width=\linewidth]{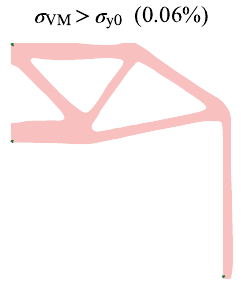}
        \vspace{-14.2em}
        \captionsetup{justification=raggedright, singlelinecheck=false, skip=-3.5pt, position=top,labelfont=subtxt, textfont=subtxt}
        \caption[]{}
        \label{fig stress vol loc 7}
    \end{subfigure}
    \begin{subfigure}[t]{0.23\textwidth}
        \includegraphics[width=\linewidth]{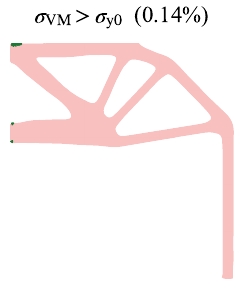}
        \vspace{-14.2em}
        \captionsetup{justification=raggedright, singlelinecheck=false, skip=-3.5pt, position=top,labelfont=subtxt, textfont=subtxt}
        \caption[]{}
        \label{fig stress vol loc 8}
    \end{subfigure}
    \caption{\textbf{Location of stress violations in the optimized 2D designs:} Elements whose von Mises stress exceeds $\sigma_{\mathrm{y0}}$ in the FE verification of \Cref{fig stress vol 2d} are marked in green. The left column (a, c, e, g) shows the SNTO designs with 250K nodal CPs and the right column (b, d, f, h) shows the PolyStress designs for the L beam, crack, hook, and portal frame examples. The percentage of violating elements is reported in each panel. For both methods the violations are confined to a few elements at the re-entrant corners, crack tip, loading points, and support boundaries.}
    \label{fig stress vol loc}
\end{figure*}
%=======================================

\Cref{fig stress vol loc} shows the elements whose stress exceeds $\sigma_{\mathrm{y0}}$ and reports their percentage in each design. For both methods the violating elements make up less than one percent of the mesh. SNTO has a slightly larger fraction than PolyStress in the L beam ($0.12\%$ versus $0.06\%$), the crack ($0.14\%$ versus $0.06\%$) and the hook ($0.52\%$ versus $0.19\%$), whereas the order is reversed in the portal frame ($0.06\%$ versus $0.14\%$). In the L beam and crack examples the violations sit at the re-entrant corner, the crack tip and the junctions of thin members. In the hook they lie along the inner curve of the hook and around the clamped hole, in which the stress field in \CrefSub{fig stress vol 2d}{fig stress vol 2d 5} is highest. In the portal frame they are limited to the fixed supports on the left edge. These locations coincide with the geometric singularities and loading points discussed above. The slightly larger fractions of SNTO are consistent with its lower volume fractions, since thinner members carry the same load at a higher stress.

\subsection{3D Benchmarks with Scalability to Millions of Local Stress Constraints}\label{sec res 3D}

%========================================
\begin{figure*}[!b]
    \centering
    \begin{subfigure}[t]{0.48\textwidth}
        \centering
        \includegraphics[width=\linewidth]{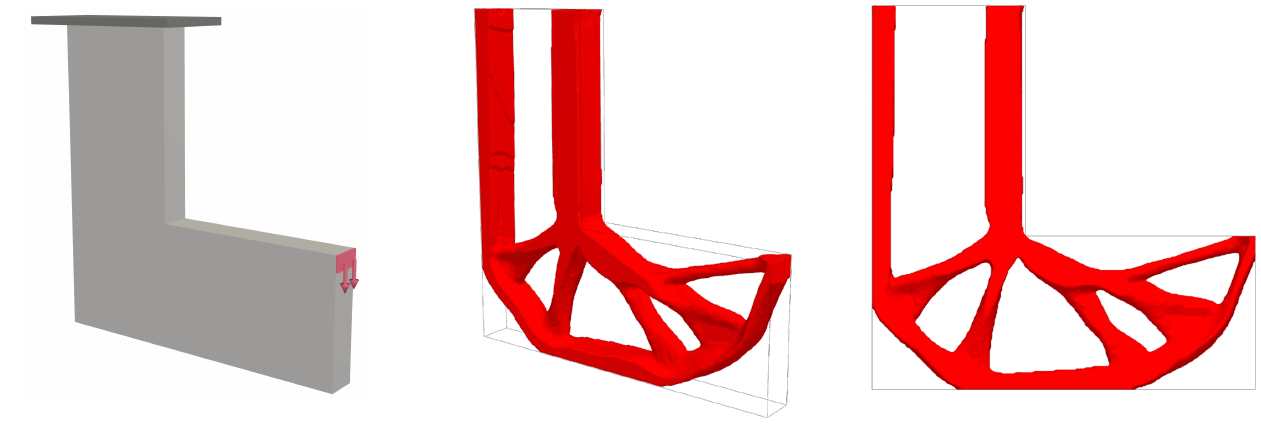}
        % \vspace{-13.5em}
        % \captionsetup{justification=raggedright, singlelinecheck=false, skip=-3.5pt, position=top,labelfont=subtxt, textfont=subtxt}
        \caption{L beam}
        \label{fig comp 3D left}
    \end{subfigure}%
    \hfill % This keeps them on the same line and spaces them evenly
    \begin{subfigure}[t]{0.48\textwidth}
        \centering
        \includegraphics[width=\linewidth]{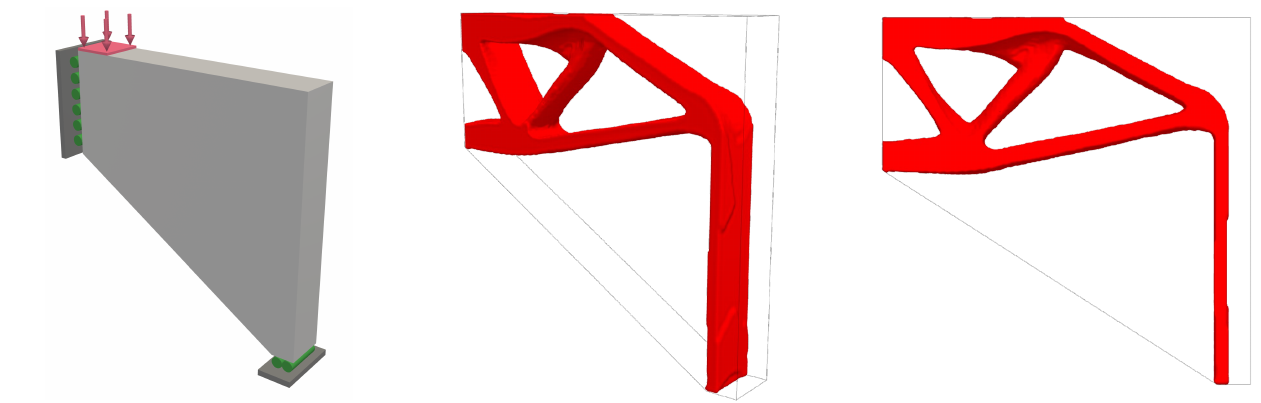}
        % \vspace{-13.5em}
        % \captionsetup{justification=raggedright, singlelinecheck=false, skip=-3.5pt, position=top,labelfont=subtxt, textfont=subtxt}
        \caption{Portal frame}
        \label{fig comp 3D right}
    \end{subfigure}
    
    \caption{\textbf{Optimized topologies (SNTO with 2M nodal CPs):}
    \textbf{(a)} the L beam and \textbf{(b)} the portal frame, each shown from an isometric view and a front view. The front views closely resemble the 2D designs in \Cref{fig comp 2D}, while the isometric views reveal 3D features such as the smoothly rounded load path at the re-entrant corner of the L beam and the fillet through the thickness where the diagonal strut of the portal frame meets the upper chord.
    }
    \label{fig comp 3D}
\end{figure*}
%=======================================
To our knowledge, no open-source implementation is available for 3D stress-constrained TO; we therefore study the benchmarks in \Cref{fig exam3D} via SNTO only. These studies aim to demonstrate SNTO's performance when the number of local stress constraints grows to millions and the training is distributed over several GPUs.
The training protocol and the hyperparameters carry over from the 2D examples with two adjustments that preserve numerical stability across random runs in 3D. First, the learning rate of the density network is reduced from $2{\times}10^{-5}$ to $10^{-5}$. 3D domains contain many more CPs, and a more conservative density update prevents abrupt changes in the stiffness so that the displacement and adjoint fields remain in equilibrium at every epoch. Second, the target magnitude of the scaled adjoint force is reduced from $10$ to $1$ in \Cref{eq alpha adj}, which makes it easier to keep the adjoint field in equilibrium.

\Cref{fig comp 3D} presents the optimized topologies with 2M nodal CPs for the 3D L beam and the portal frame from two viewing angles. The front view of each example closely resembles its 2D counterpart, see \Cref{fig top mesh 2D,fig comp 2D}. 
While there are similarities and the loading as well as BCs are uniform through the depth, the 3D designs are not planar extrusions of their 2D counterparts. For instance, in the portal frame benchmark the central diagonal strut thins where it meets the upper chord and forms a fillet through the thickness that relieves the stress concentration.

\subsubsection{Scalability}\label{sec res scal}

We increase the number of nodal CPs sixteenfold from 500K to 8M CPs to study its effect on the optimized topologies and computational costs. \Cref{tab 3D obj} summarizes the final design metrics at the five resolutions. Every L beam design remains nearly binary whereas the portal frame designs have a slightly larger gray area fraction around 0.04. In both examples the gray fractions are comparable to those of the corresponding 2D designs in \Cref{tab 2D obj}. 
The volume fraction of the L beam varies by less than $0.01$ across the five resolutions, which is comparable to the seed-to-seed variation observed in 2D, whereas the portal frame shows a larger variation with a range of $0.034$. In both examples the 8M design has the highest volume fraction. A plausible explanation is that the denser centroid CPs capture sharper stress concentrations at the re-entrant corner and at the loaded and supported regions, which the optimizer balances with more material in their vicinity.

The runtimes are not directly comparable across the columns of \Cref{tab 3D obj} since the number of GPUs, sub-slices, and CPs changes. In \Cref{sec res cost} we study the scalability and cost systematically and here we focus on a complementary trend. Specifically, we observe in \Cref{tab 3D obj} that the portal frame is consistently slower than the L beam, even though its volume fraction is smaller. The difference stays about $5\%$ up to 2M CPs but grows to $13\%$ at 4M and $23\%$ at 8M CPs, which are the two cases trained on four GPUs. We attribute this trend to the distribution of active elements across the slices. Each slice holds the same number of elements, but the number of active elements per slice diverges as material is removed. In the portal frame the void forms mainly below the horizontal strut, and the slices covering that region lose most of their active elements while those through the top frame and the vertical leg keep theirs. Since the parameter gradients are synchronized after every epoch, the epoch time is set by the GPU holding the most active elements. This imbalance grows with the number of GPUs, whereas in the L beam the material is distributed more evenly along the slicing direction.

%=======================================
\begin{table*}[!b]
    \centering
    \renewcommand{\arraystretch}{1.5}
    \footnotesize
    \setlength\tabcolsep{6pt}
    \caption{\textbf{Volume fraction, gray area fraction, and wall-clock runtime of the optimized 3D benchmarks:}
    For each example, the values are reported for the five mesh resolutions listed in \Cref{tab mesh}. Each value is the mean over two independent runs, except for the 8M cases, which were run once. All cases are trained for 30{,}000 epochs on NVIDIA A100 GPUs. The number of GPUs and the number of sub-slices per GPU are listed in the first row of each example.}
    \resizebox{\textwidth}{!}{%
    \begin{tabular}{l|l|c|c|c|c|c}
        \hline
        \textbf{Example}
        & \textbf{Metric}
        & \textbf{SNTO 3D 500K}
        & \textbf{SNTO 3D 1M}
        & \textbf{SNTO 3D 2M}
        & \textbf{SNTO 3D 4M}
        & \textbf{SNTO 3D 8M} \\
        \hline

        \multirow{4}{*}{L beam 3D}
        & GPUs $\times$ sub-slices & $1 \times 2$ & $2 \times 2$ & $2 \times 4$ & $4 \times 4$ & $4 \times 10$ \\
        & Volume fraction & 0.342 & 0.347 & 0.350 & 0.345 & 0.351 \\
        & Gray fraction   & 0.015 & 0.013 & 0.009 & 0.015 & 0.015 \\
        & Runtime (sec)     & 7661  & 8912  & 17784 & 22037 & 50532 \\
        \hline

        \multirow{4}{*}{Portal frame 3D}
        & GPUs $\times$ sub-slices & $1 \times 2$ & $2 \times 2$ & $2 \times 4$ & $4 \times 4$ & $4 \times 10$ \\
        & Volume fraction & 0.234 & 0.224 & 0.228 & 0.248 & 0.258 \\
        & Gray fraction   & 0.046 & 0.030 & 0.038 & 0.042 & 0.034 \\
        & Runtime (sec)     & 7880  & 9026  & 18707 & 24892 & 62036 \\
        \hline

    \end{tabular}
    }
    \label{tab 3D obj}
\end{table*}
%=======================================

\begin{figure*}[!t]
    \centering
    \includegraphics[width = 0.85\textwidth]{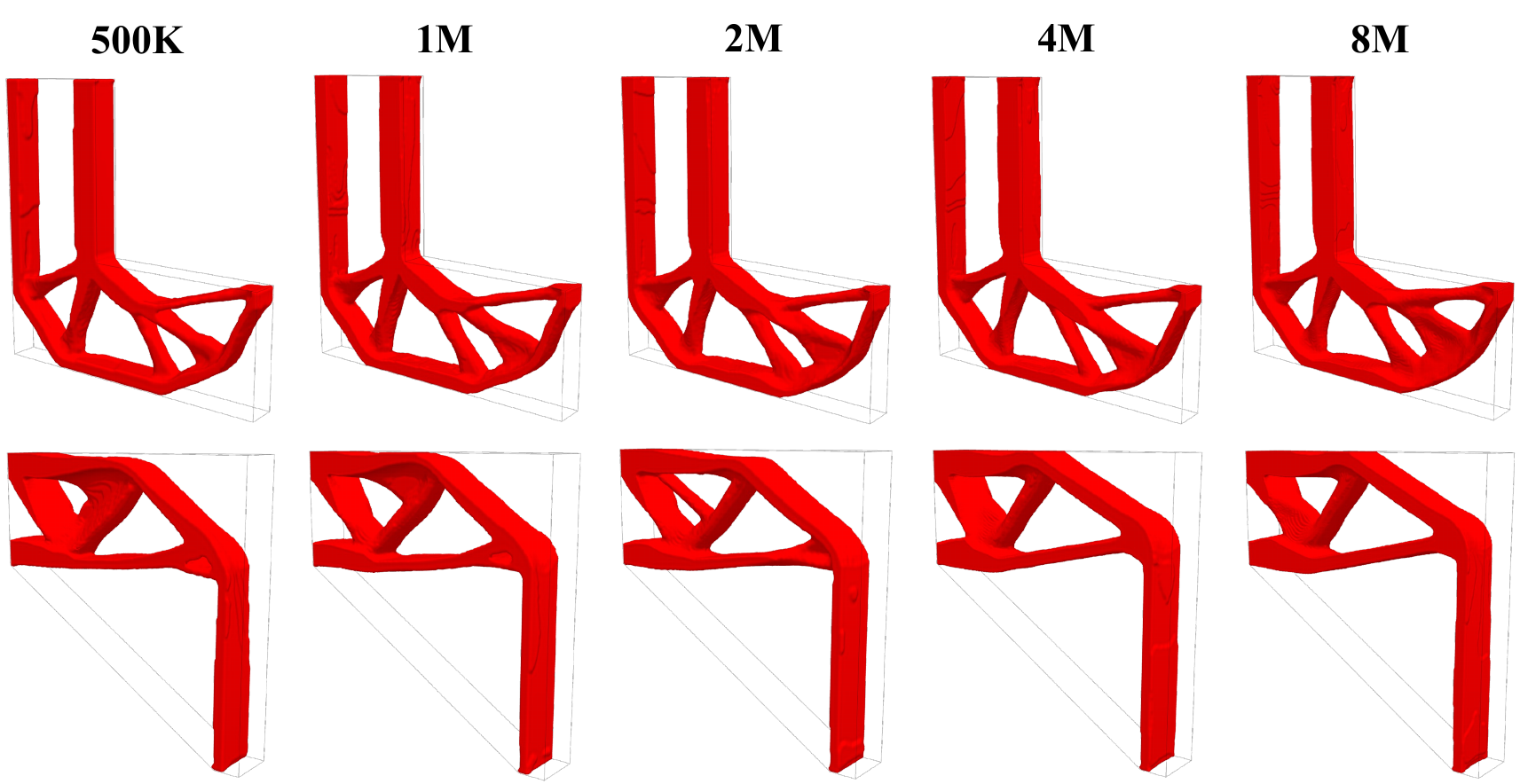} 
    \caption{\textbf{Mesh-resolution study for optimized 3D topologies:}
    Final designs are shown for the L beam (top) and portal frame (bottom) benchmarks with progressively more nodal CPs.
    The main structural features are preserved as the number of local stress constraints grows by more than an order of magnitude, which shows that SNTO is essentially mesh independent even without any filtering.}
    \label{fig top mesh 3D}
\end{figure*}

To further assess the scalability of our approach, we compare the optimized topologies across the five nodal CP resolutions in \Cref{fig top mesh 3D}. We observe that the designed topologies are consistent as the number of nodal CPs is increased from 500K to 8M: in each example, every primary load path appears at all resolutions and the differences between them are mostly limited to the thickness of a few members and the shape of the fillets that round the member junctions. The differences in the final volume fraction are too small to be seen in \Cref{fig top mesh 3D}, except that the L beam with 8M nodal CPs has slightly thicker members near the loading point than the other resolutions. We obtain this consistency without any explicit filtering since the only regularization is the resolution of the PGCAN feature grid, which is the same in all five cases. As discussed in \Cref{sec res 2D mesh}, refining the mesh affects only the evaluation of the loss and not the parameterization of the design.

%========================================
%=======================================

%========================================
\begin{figure*}[!b]
    \centering
    \begin{subfigure}[t]{\textwidth}
        \centering
        \includegraphics{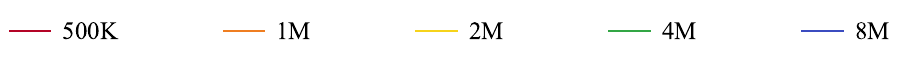}
    \end{subfigure}
    \begin{subfigure}[t]{0.32\textwidth}
        \includegraphics[width=\linewidth]{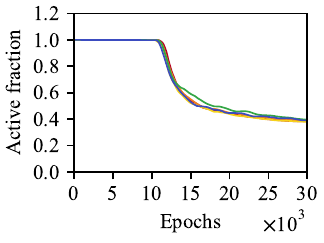}
        \vspace{-14em}
        \captionsetup{justification=raggedright, singlelinecheck=false, skip=-5.3pt, position=top,labelfont=subtxt, textfont=subtxt}
        \caption[]{}
        \label{fig hist mesh L 3D 1}
    \end{subfigure}
    \begin{subfigure}[t]{0.32\textwidth}
        \includegraphics[width=\linewidth]{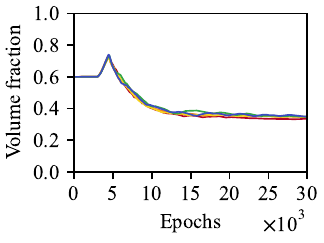}
        \vspace{-14em}
        \captionsetup{justification=raggedright, singlelinecheck=false, skip=-2.9pt, position=top,labelfont=subtxt, textfont=subtxt}
        \caption[]{}
        \label{fig hist mesh L 3D 2}
    \end{subfigure}
    \begin{subfigure}[t]{0.32\textwidth}
        \includegraphics[width=\linewidth]{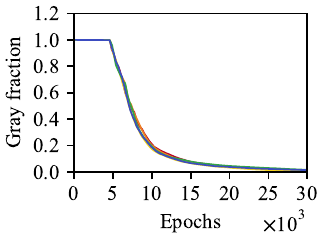}
        \vspace{-14em}
        \captionsetup{justification=raggedright, singlelinecheck=false, skip=-2.9pt, position=top,labelfont=subtxt, textfont=subtxt}
        \caption[]{}
        \label{fig hist mesh L 3D 3}
    \end{subfigure}
    \begin{subfigure}[t]{0.32\textwidth}
        \includegraphics[width=\linewidth]{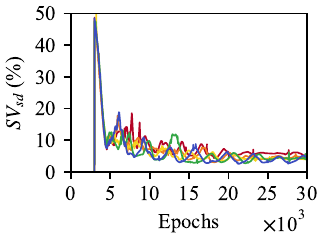}
        \vspace{-14em}
        \captionsetup{justification=raggedright, singlelinecheck=false, skip=-2.9pt, position=top,labelfont=subtxt, textfont=subtxt}
        \caption[]{}
        \label{fig hist mesh L 3D 4}
    \end{subfigure}
    \begin{subfigure}[t]{0.32\textwidth}
        \includegraphics[width=\linewidth]{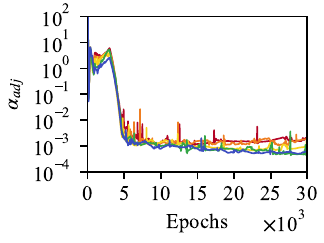}
        \vspace{-14em}
        \captionsetup{justification=raggedright, singlelinecheck=false, skip=-2.9pt, position=top,labelfont=subtxt, textfont=subtxt}
        \caption[]{}
        \label{fig hist mesh L 3D 5}
    \end{subfigure}
    \begin{subfigure}[t]{0.32\textwidth}
        \includegraphics[width=\linewidth]{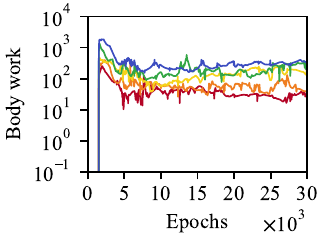}
        \vspace{-14em}
        \captionsetup{justification=raggedright, singlelinecheck=false, skip=-2.9pt, position=top,labelfont=subtxt, textfont=subtxt}
        \caption[]{}
        \label{fig hist mesh L 3D 6}
    \end{subfigure}
    \caption{\textbf{Sensitivity of evolution histories with respect to number of nodal CPs using the 3D L beam benchmark:}
    Evolution histories are compared for the runs with 500K, 1M, 2M, 4M, and 8M nodal CPs. The panels report \textbf{(a)} active element fraction, \textbf{(b)} volume fraction, \textbf{(c)} gray area fraction, \textbf{(d)} stress violation over the realized solid material, \textbf{(e)} adjoint scaling factor, and \textbf{(f)} body work contribution in the adjoint problem.}
    \label{fig hist mesh L 3D}
\end{figure*}
%=========================================================================

The training histories of the L beam in \Cref{fig hist mesh L 3D} corroborate the above arguments on mesh independence. The active element fraction, the gray fraction, and the stress violation ${SV}_{sd}$ (\CrefSub{fig hist mesh L 3D}{fig hist mesh L 3D 1}, \CrefSub{fig hist mesh L 3D}{fig hist mesh L 3D 3}, and \CrefSub{fig hist mesh L 3D}{fig hist mesh L 3D 4}, respectively) collapse onto a single curve at all five resolutions with only minor transient variations. The volume fraction in \CrefSub{fig hist mesh L 3D}{fig hist mesh L 3D 2} follows the same trajectory for the first four resolutions while the 8M case settles at a slightly higher level, consistent with the final values in \Cref{tab 3D obj}.

All histories follow the characteristic 2D trajectory with an initial warm-up plateau, an overshoot of the volume fraction when the stress penalty activates, and a steady convergence under the continuation of $p$ and $\beta$. The late drift of ${SV}_{sd}$ observed on the coarse meshes of the 2D crack example in \CrefSub{fig hist mesh crack 2d}{fig hist mesh crack 2d 4} does not occur at any 3D resolution for the L beam benchmark, indicating that the coarsest 3D mesh with 500K CPs already provides sufficient accuracy for evaluating the loss. Overall, these quantities that describe the design and its constraint satisfaction are insensitive to the mesh.

The number of CPs does leave a trace in the adjoint quantities. The adjoint scaling factor $\alpha_{adj}$ in \CrefSub{fig hist mesh L 3D}{fig hist mesh L 3D 5} and the body work of the adjoint field in \CrefSub{fig hist mesh L 3D}{fig hist mesh L 3D 6} are ordered by resolution. $\alpha_{adj}$ decreases from about $2{\times}10^{-3}$ at 500K to below $10^{-3}$ at 8M, while the body work increases by about an order of magnitude over the same range. We argue that this ordering is a consequence of the normalization in \Cref{eq alpha adj}: $\alpha_{adj}$ is the peak entry of the adjoint force divided by $F_{goal} = N_c$ and therefore decreases as the mesh is refined, whereas the scaled force $\Fb^{(a)}/\alpha_{adj}$ has the prescribed maximum entry $N_c$, so a finer mesh applies a larger adjoint load over more nodes and the adjoint field responds with a larger body work. Neither trend affects the design sensitivity, which is invariant to $\alpha_{adj}$ by construction, as can be verified from \Cref{eq P_rho2}. This invariance is what allows the same hyperparameters to be used at all five resolutions despite the shift in the adjoint load.

The history of $\alpha_{adj}$ is nonetheless informative. It decreases by only a factor of three to four while the number of CPs grows sixteenfold. The peak adjoint force itself therefore grows with background mesh refinement. This is expected as a finer mesh samples the stress concentration at the re-entrant corner more closely and the largest violations become more severe while the active region shrinks. This observation supports the explanation proposed above for the slightly heavier 8M designs. A more accurately resolved peak stress is a stricter constraint and the optimizer satisfies it by placing more material along the load path. The same effect occurs at the surfaces and junctions of the members, where the stress is highest and a denser set of centroid CPs samples it closer to the boundary. The members of the 8M design are therefore slightly thicker throughout rather than only near the corner. This difference originates in the evaluation of the stress constraint and not in the parameterization of the design, which is the same at all five resolutions.

\subsubsection{Stress Constraint Verification}\label{res sec constr 3d}

%========================================
\begin{figure*}[!t]
    \centering
    \captionsetup[subfigure]{
        justification=raggedright,
        singlelinecheck=false,
        skip=2pt,
        position=top,
        labelfont=subtxt,
        textfont=subtxt
    }

    \begin{subfigure}[t]{0.49\textwidth}
        \caption{L beam, 500K}
        \label{fig stress vol 3d 1}
        \includegraphics[width=\linewidth]{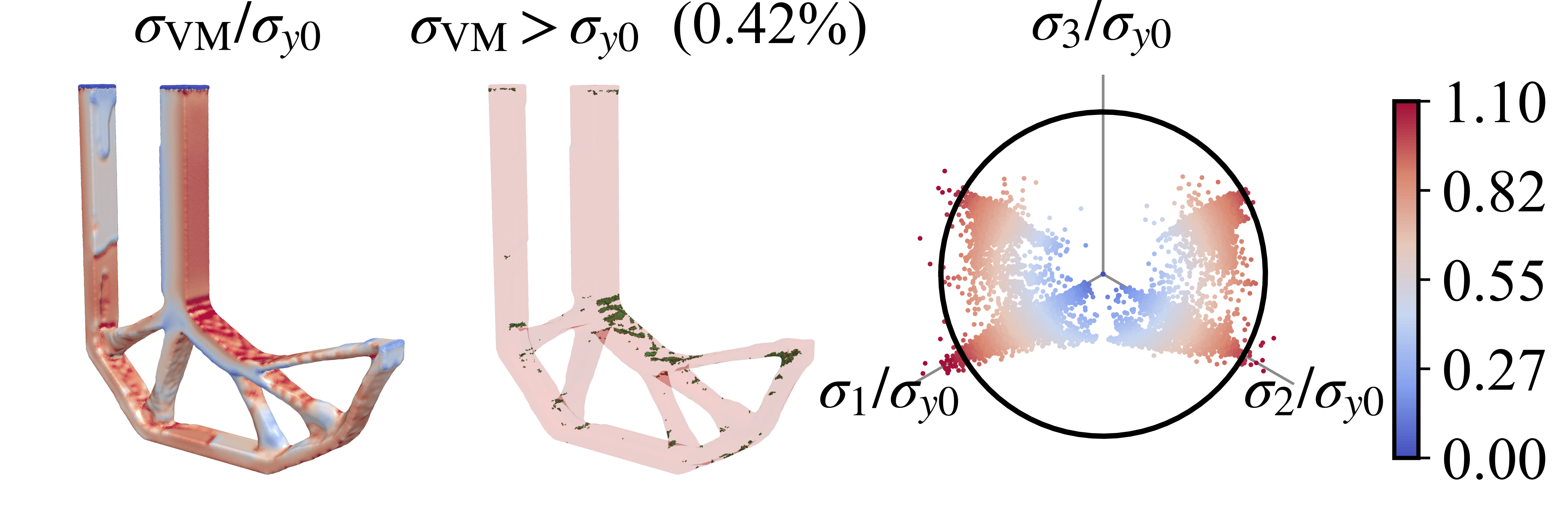}
    \end{subfigure}
    \hfill
    \begin{subfigure}[t]{0.49\textwidth}
        \caption{Portal frame, 500K}
        \label{fig stress vol 3d 2}
        \includegraphics[width=\linewidth]{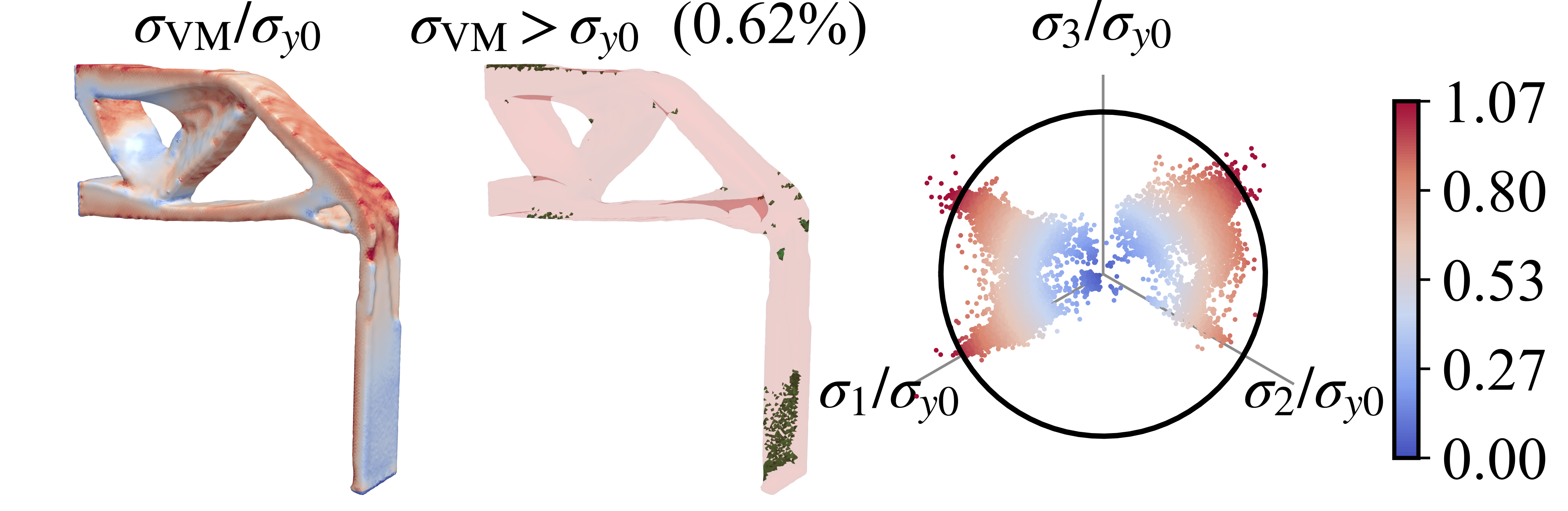}
    \end{subfigure}

    \par\vspace{-0.5em}

    \begin{subfigure}[t]{0.49\textwidth}
        \caption{L beam, 1M}
        \label{fig stress vol 3d 3}
        \includegraphics[width=\linewidth]{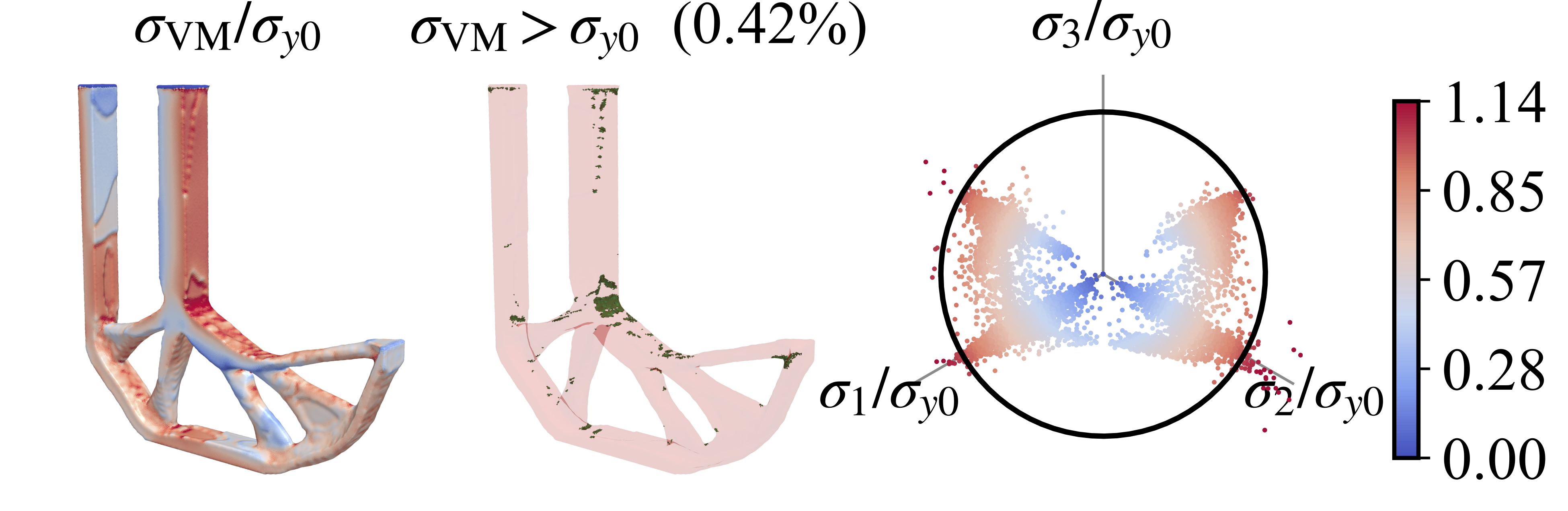}
    \end{subfigure}
    \hfill
    \begin{subfigure}[t]{0.49\textwidth}
        \caption{Portal frame, 1M}
        \label{fig stress vol 3d 4}
        \includegraphics[width=\linewidth]{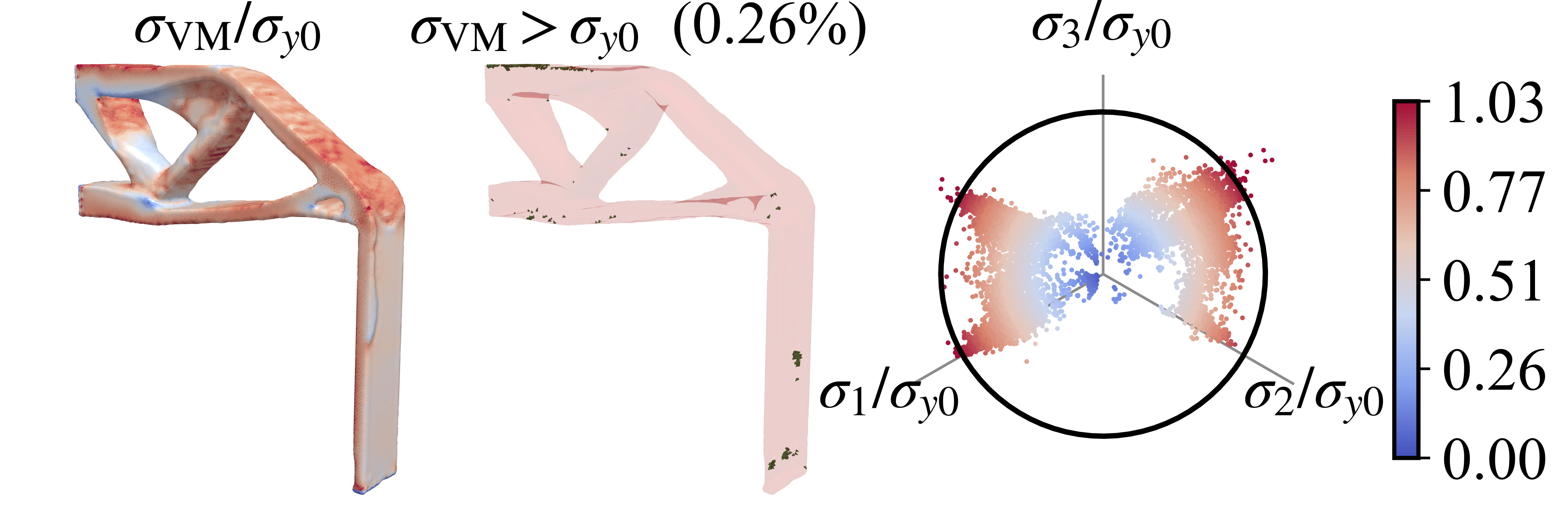}
    \end{subfigure}

    \par\vspace{-0.5em}

    \begin{subfigure}[t]{0.49\textwidth}
        \caption{L beam, 2M}
        \label{fig stress vol 3d 5}
        \includegraphics[width=\linewidth]{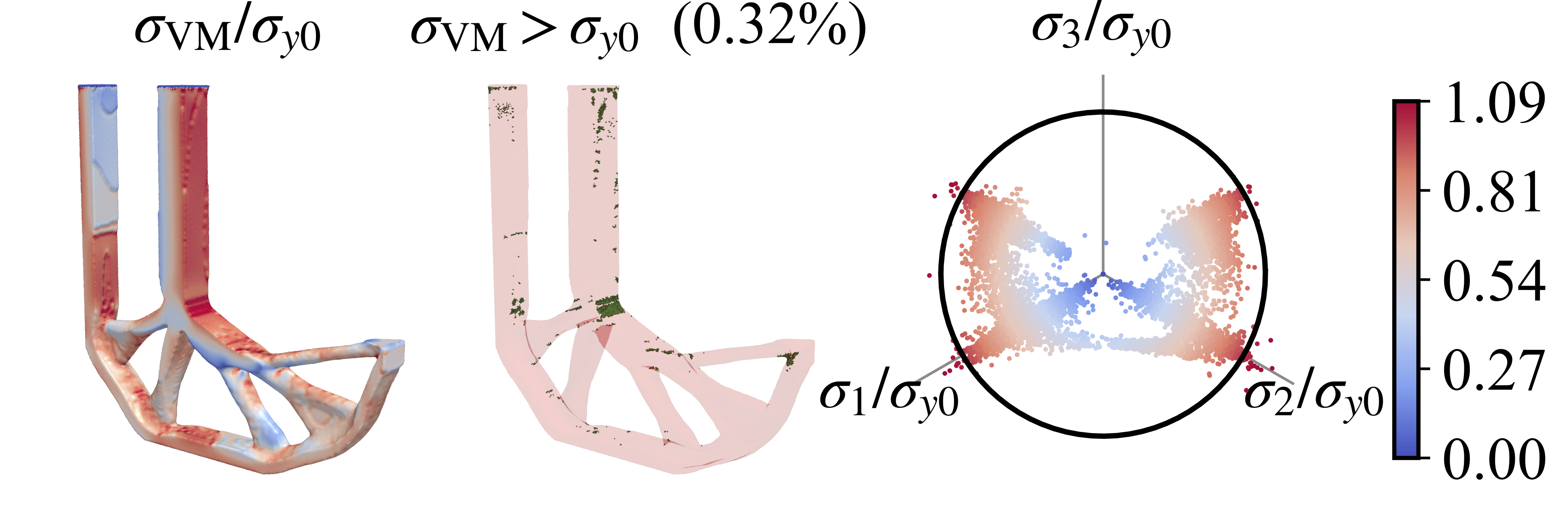}
    \end{subfigure}
    \hfill
    \begin{subfigure}[t]{0.49\textwidth}
        \caption{Portal frame, 2M}
        \label{fig stress vol 3d 6}
        \includegraphics[width=\linewidth]{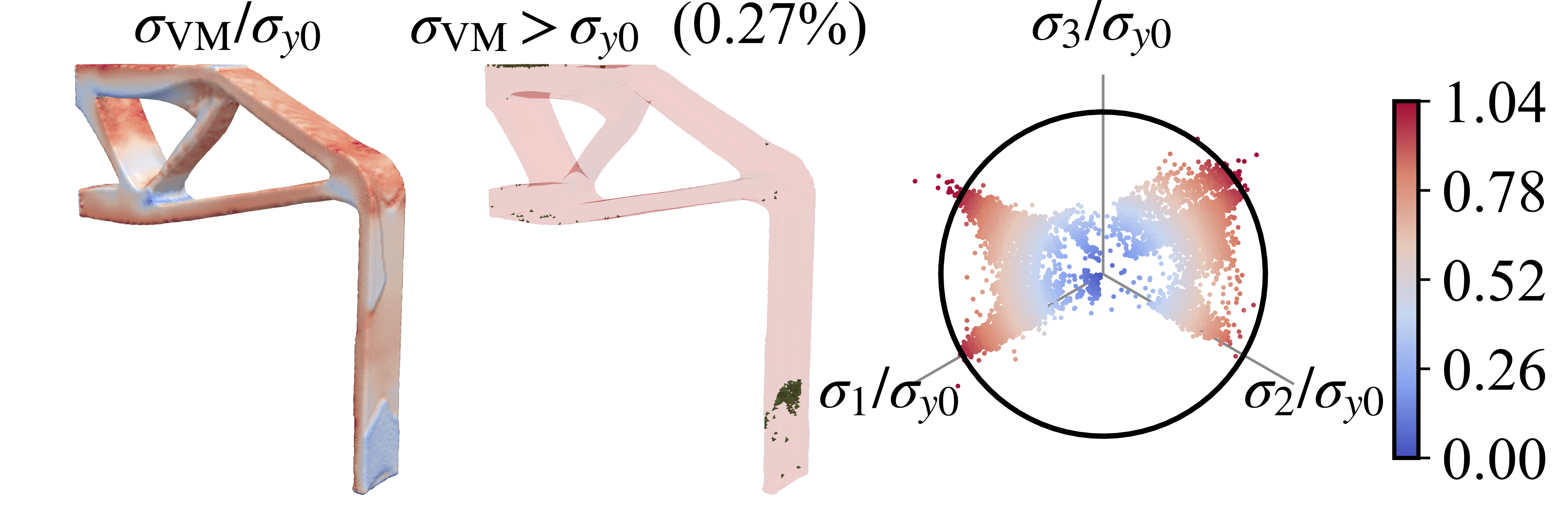}
    \end{subfigure}

    \par\vspace{-0.5em}

    \begin{subfigure}[t]{0.49\textwidth}
        \caption{L beam, 4M}
        \label{fig stress vol 3d 7}
        \includegraphics[width=\linewidth]{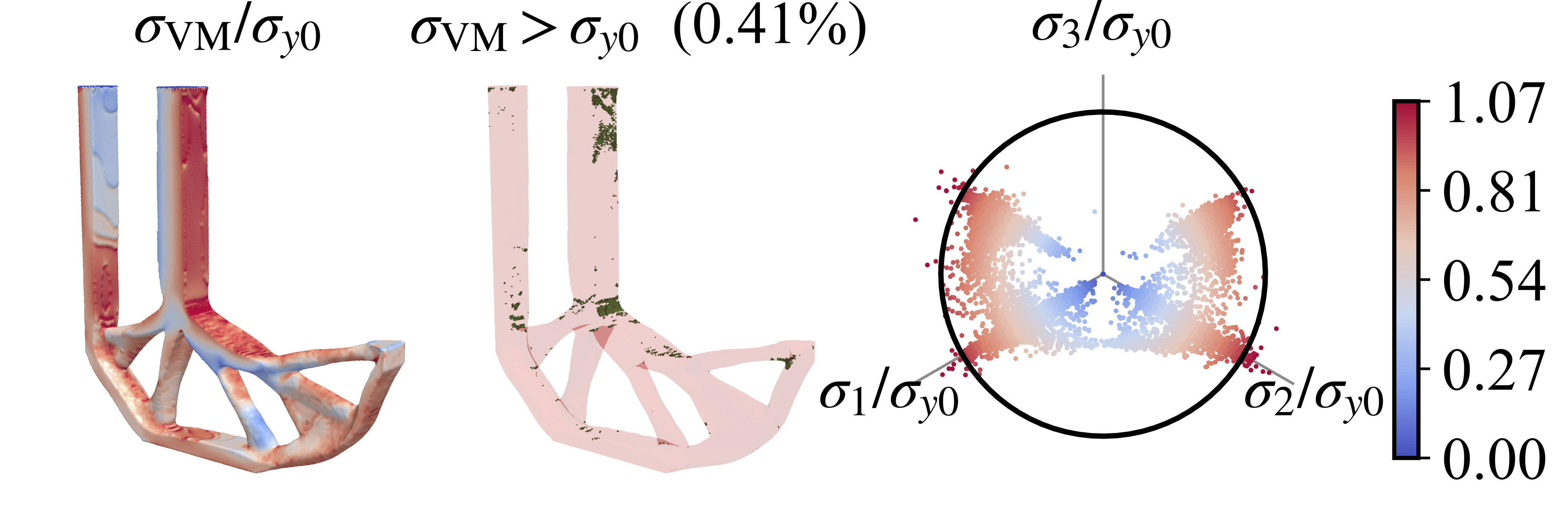}
    \end{subfigure}
    \hfill
    \begin{subfigure}[t]{0.49\textwidth}
        \caption{Portal frame, 4M}
        \label{fig stress vol 3d 8}
        \includegraphics[width=\linewidth]{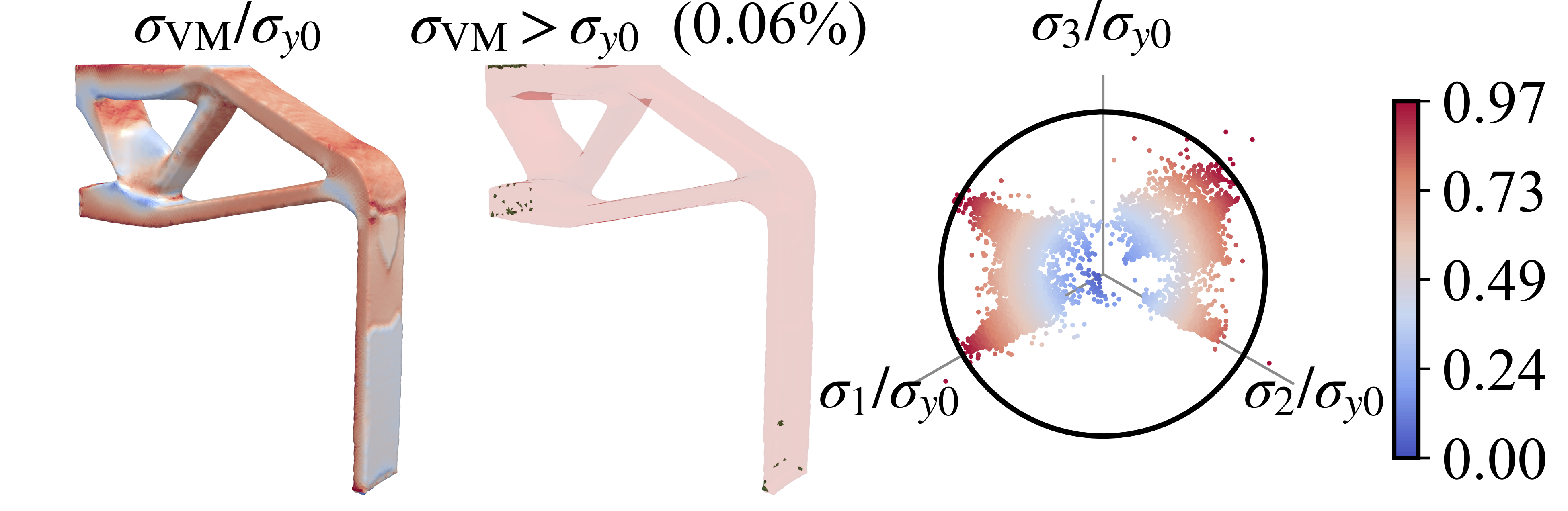}
    \end{subfigure}

    \par\vspace{-0.5em}

    \begin{subfigure}[t]{0.49\textwidth}
        \caption{L beam, 8M}
        \label{fig stress vol 3d 9}
        \includegraphics[width=\linewidth]{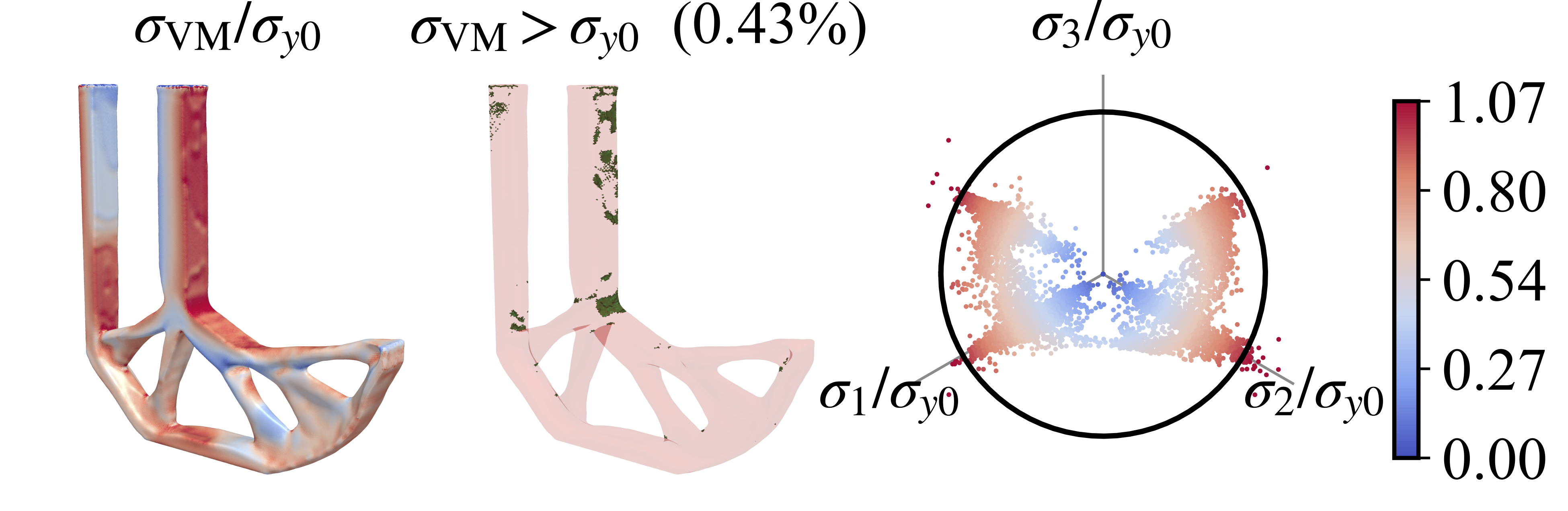}
    \end{subfigure}
    \hfill
    \begin{subfigure}[t]{0.49\textwidth}
        \caption{Portal frame, 8M}
        \label{fig stress vol 3d 10}
        \includegraphics[width=\linewidth]{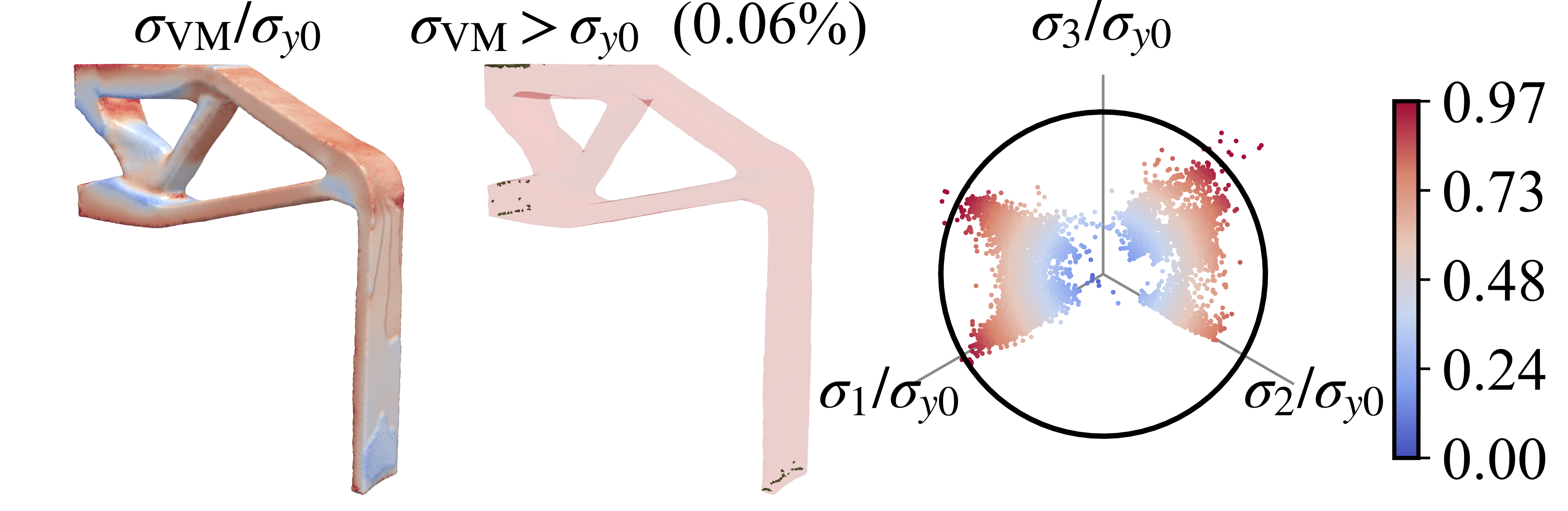}
    \end{subfigure}

    \caption{\textbf{Stress-constraint verification for the optimized 3D designs.}
    Normalized von Mises stress distributions are shown for the L beam in the left column and the portal frame in the right column using 500K, 1M, 2M, 4M, and 8M nodal CPs. Each panel presents the normalized von Mises stress $\sigma_{\mathrm{VM}}/\sigma_{\mathrm{y0}}$ over the optimized geometry, a transparent rendering in which regions with $\sigma_{\mathrm{VM}}>\sigma_{\mathrm{y0}}$ are highlighted in dark green, and the corresponding normalized principal-stress states $(\sigma_1,\sigma_2,\sigma_3)/\sigma_{\mathrm{y0}}$. The black curve denotes the von Mises limit, and the scatter colors indicate the local stress ratio.}
    \label{fig stress vol 3d}
\end{figure*}
%========================================

To further evaluate the enforcement of the stress constraint, we remesh the 3D designs obtained at all five nodal CP resolutions with linear tetrahedral elements and solve the linear elastic problem in FEniCS under the boundary conditions of \Cref{fig exam3D}. The von Mises yield surface in 3D is a cylinder about the hydrostatic axis in principal-stress space. We therefore project the normalized element stress states $(\sigma_1,\sigma_2,\sigma_3)/\sigma_{\mathrm{y0}}$ along that axis onto the deviatoric plane where the yield surface reduces to a circle and points outside it violate the constraint as shown in \Cref{fig stress vol 3d}. A small number of elements exhibit localized stress peaks, e.g., near the clamped boundary, that are far larger than the stresses in the rest of the design and would otherwise dominate the color scale. For visualization only, we therefore cap the color scale at the $99.9$th percentile of $\sigma_{\mathrm{VM}}/\sigma_{\mathrm{y0}}$ over all elements.

In the L beam, shown in the left column of \Cref{fig stress vol 3d}, the highest stresses occur along the rounded corner. A few violating elements also appear along the vertical flanges and at the web joints. The violating elements make up approximately $0.4\%$ of the remeshed designs with no systematic trend as the resolution increases. The $99.9$th percentile of $\sigma_{\mathrm{VM}}/\sigma_{\mathrm{y0}}$ lies between $1.07$ and $1.14$, and most points outside the circle lie along the $\sigma_1$ and $\sigma_2$ axes, which correspond to nearly uniaxial tension or compression as expected for flanges in bending. The violating fraction of the L beam is therefore largely insensitive to the nodal CP resolution, even though the volume fraction of the 8M design is slightly larger.

The right column of \Cref{fig stress vol 3d} shows the results for the portal frame. At 500K, $0.62\%$ of the elements exceed the stress limit. These elements lie along the top and bottom of the horizontal frame and at the supported end of the leg. Increasing the nodal CP resolution reduces this fraction to about $0.26\%$ at 1M and 2M and further to $0.06\%$ at 4M and 8M. The $99.9$th percentile of the stress ratio also drops from $1.07$ at 500K to $0.97$ at 8M. The few violations that remain at 8M are confined to the loading points and the support boundary. A finer resolution captures the stress concentrations at these locations more accurately. The optimizer responds by thickening the members across the whole design, which raises the volume fraction from $0.228$ at 2M to $0.258$ at 8M (\Cref{tab 3D obj}). The FE analysis confirms that the thicker design carries lower stresses and removes most of the violations. 
For this example the training estimate is ${SV}_{sd} \approx 6\%$ (\CrefSub{fig hist mesh portal 3D}{fig hist mesh portal 3D 4}), which is two orders of magnitude larger than the FE value of $0.06\%$ at 8M. The difference comes from the definition of the two quantities. ${SV}_{sd}$ counts violations against the tightened training stress limit $\alpha_s\sigma_{\mathrm{y0}} = 0.9\,\sigma_{\mathrm{y0}}$ (\Cref{sec metd train}), whereas the FE check uses the yield strength $\sigma_{\mathrm{y0}}$.

In both examples the stress violations occur in the load-carrying members or near the loading points and the support boundary. The $99.9$th percentile of the stress ratio is $1.07$--$1.14$ for the L beam and $0.97$--$1.07$ for the portal frame. The corresponding values for the 2D L beam and portal frame are $1.00$ and $0.96$. Scaling up to 3D and to millions of local constraints therefore does not degrade constraint satisfaction. 
Nonetheless, some elements exceed the nominal limit $\sigma_{\mathrm{y0}}$ in every case even though training enforces the tightened limit $\alpha_s \sigma_{\mathrm{y0}}$ with $\alpha_s = 0.9$ (\Cref{sec metd train}). The constraint can be imposed more strictly by reducing $\alpha_s$ further. This only redefines the stress limit and requires no other change to the framework but it leads to a heavier design.

\subsubsection{Computational Cost and Memory}\label{sec res cost}

As discussed in \Cref{sec metd gpu}, we employ two levels of partitioning to manage the computational resources required for large-scale 3D problems. The design domain is first divided evenly across the available GPUs. Each subdomain is then sub-sliced into smaller regions that are processed sequentially on its GPU. We conduct two scaling studies to assess the effect of each level. The first examines the memory and time trade-off introduced by sub-slicing on a fixed number of GPUs. The second evaluates the parallel speedup from additional GPUs while the number of sub-slices per GPU is reduced in proportion to keep the total sub-slice size approximately constant. For each configuration we record the training time over $30{,}000$ epochs, the peak GPU memory usage, and the optimized topology.

%========================================
\begin{figure*}[!t]
    \centering
    \begin{subfigure}[t]{0.49\textwidth}
        \includegraphics[width=\linewidth]{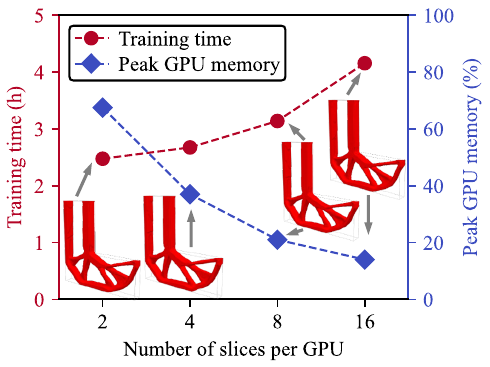}
        \vspace{-19em}
        \captionsetup{justification=raggedright, singlelinecheck=false, skip=-5.5pt, position=top,labelfont=subtxt, textfont=subtxt}
        \caption[]{}
        \label{fig cost 3d 1}
    \end{subfigure}
    \hfill
    \begin{subfigure}[t]{0.49\textwidth}
        \includegraphics[width=\linewidth]{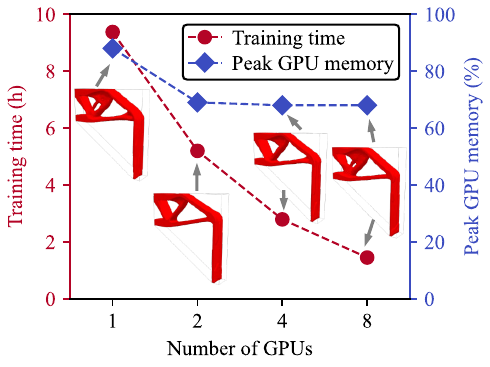}
        \vspace{-19em}
        \captionsetup{justification=raggedright, singlelinecheck=false, skip=-5.5pt, position=top,labelfont=subtxt, textfont=subtxt}
        \caption[]{}
        \label{fig cost 3d 2}
    \end{subfigure}
    \caption{\textbf{Computational cost and memory scaling of the 3D examples:}
    \textbf{(a)} Sub-slicing sweep on the L beam with 1M nodal CPs using two GPUs. The training time (left axis, dark red circles) and the peak GPU memory usage (right axis, dark blue diamonds) are plotted against the number of sub-slices per GPU (2, 4, 8, and 16). Finer sub-slicing lowers the peak memory at the cost of a longer training time. \textbf{(b)} Multi-GPU sweep on the portal frame with 2M nodal CPs. The same two quantities are plotted against the number of GPUs (1, 2, 4, and 8), with the number of sub-slices per GPU reduced accordingly (8, 4, 2, and 1) so that the total number of sub-slices and the peak memory usage stay roughly constant. The training time decreases nearly linearly with the number of GPUs. The optimized topologies are shown next to the corresponding data points.}
    \label{fig cost 3d}
\end{figure*}
%=======================================
In \CrefSub{fig cost 3d}{fig cost 3d 1}, the 1M L beam is trained on two GPUs with $2$ to $16$ sub-slices per GPU. Increasing the number of sub-slices by a factor of eight reduces the peak memory usage from $68\%$ to $14\%$ of the device capacity. The reduction is sublinear because the memory footprint includes both the autograd graph for the current sub-slice and a static baseline of approximately $6\%$ for the network replicas, optimizer states, and cached subdomain data. The training time increases from $2.5$ to $4.2$ hours since each additional sub-slice adds a forward and backward pass and lowers GPU utilization. All configurations produce nearly identical topologies, confirming that gradient accumulation preserves the optimization result.

In \CrefSub{fig cost 3d}{fig cost 3d 2}, the 2M portal frame is trained on $1$, $2$, $4$, and $8$ GPUs with $8$, $4$, $2$, and $1$ sub-slices per GPU, respectively. This configuration keeps the number of nodal CPs per sub-slice fixed. The peak memory usage therefore remains approximately $68\%$ for two or more GPUs. The single-GPU case requires more memory because it stores the CP data for the entire domain. 
Increasing the number of GPUs from one to eight reduces the training time from $9.4$ to $1.5$ hours, which is a speedup of $6.3\times$. Relative to the ideal speedup of $8\times$ for eight GPUs, this corresponds to a parallel efficiency of $79\%$. The deviation from ideal scaling is primarily due to the all-reduce operations used to aggregate the parameter gradients after each epoch. The optimized topologies remain consistent across all four configurations with small variations.

These results demonstrate that sub-slicing reduces peak memory usage, while multi-GPU parallelism reduces training time without significantly affecting the optimized topology. As the peak memory is governed primarily by the size of the resident sub-slice, the configurations in \Cref{tab 3D obj} are selected to contain approximately 200K--250K nodal CPs per sub-slice for 3D examples. This strategy enables the same formulation to scale from the single-GPU 2D examples to 3D problems with up to eight million nodal CPs as well as local stress constraints distributed across four GPUs.
    % \clearpage
\section{Conclusion}\label{sec con}
In this work, we developed SNTO: a simultaneous optimization framework for stress-constrained TO involving a large number of local constraints. In SNTO, the displacement, adjoint, and density fields are represented by three independent custom NNs, each taking the spatial coordinates throughout the design domain as input. All three networks employ the PGCAN architecture to be able to accurately capture stress concentrations and fine-scale design features. Distance-based analytical functions are used to impose the BCs and design constraints exactly, eliminating the need for boundary penalty loss terms. The networks are trained simultaneously by minimizing a unified composite loss that combines the volume fraction objective, a quadratic aggregation of the local stress constraints, and the potential energy functionals associated with the state and adjoint fields. To improve the stability of this simultaneous optimization process, we leverage curriculum training consisting of a staged warm-up of the equilibrium conditions, continuation of the SIMP penalization exponent and projection sharpness parameter, and a gradual increase in the constraint weight. In addition, normalizing the adjoint load stabilizes the magnitude of the adjoint field and further improves the enforcement of adjoint equilibrium.

To enable large-scale applications, SNTO exploits the additive structure of the loss gradient with respect to any partition of the CPs. The design domain is first uniformly partitioned into subdomains distributed across multiple GPUs. When GPU memory is limited, the CPs within each subdomain can be further divided into smaller sub-slices that are processed sequentially. As the total loss is a sum of contributions from all CPs, the gradients with respect to the network parameters can be accumulated over the sequential sub-slices on each GPU and then summed across GPUs before each parameter update. This strategy improves the computational efficiency with multiple GPUs and substantially reduces peak memory requirements. Moreover, continuity of all three fields across subdomains and sub-slices is guaranteed by the global neural representations, without requiring interface conditions or additional interface residual loss terms.

We evaluate SNTO on a series of 2D and 3D stress-constrained TO benchmarks, and compare the 2D results with those obtained from PolyStress. The 2D problems are solved using up to one million nodal CPs, whereas the 3D problems involve up to eight million nodal CPs. The results demonstrate that SNTO: (1) simultaneously maintains the displacement and adjoint networks in equilibrium with the evolving design throughout training; (2) produces topologies with sharper boundaries and smaller gray area fractions than PolyStress; (3) reaches lower volume fractions than PolyStress in all four examples, especially for the hook where SNTO removes the unstressed material around the fixed hole that PolyStress retains; and (4) satisfies the local stress constraints to the same degree as PolyStress, with only isolated elements exceeding the stress limit.
The scalability and robustness of the framework are further evaluated by refining the mesh. The results show that (1) the resulting topologies are largely insensitive to the background mesh resolution even without density filtering, except for a slightly heavier design in the 3D cases with 8M nodal CPs; (2) the training time decreases roughly in inverse proportion to the number of GPUs; and (3) subdividing each subdomain into smaller sub-slices further reduces the peak GPU memory usage at the cost of a longer training time. We believe our findings establish a viable pathway for applying ML-based methods to large-scale TO problems with many local constraints.

The current framework assumes linear elasticity but could be extended to nonlinear material behavior including hyperelasticity and elastoplasticity. Another promising direction is extension to transient problems where stress constraints must be satisfied over time and the adjoint equations must be solved backward in time. A continuous space–time representation of the state and adjoint fields may address these requirements without storing the complete state history, as typically required by nested approaches. In addition, adaptively sampling more points in regions of high stress could reduce the remaining constraint violations near geometric singularities. These extensions are fully compatible with the additive structure of the proposed framework.

\ifanon\else
\section*{CRediT authorship contribution statement} \label{sec credit}
\textbf{Xiangyu Sun}: conceptualization, methodology, formal analysis, software implementation, data curation, figure preparation, and writing (original draft, review and editing); 
\textbf{Shirin Hosseinmardi}: conceptualization, methodology, and writing (original draft, review and editing); 
\textbf{Ramin Bostanabad}: supervision, funding acquisition, conceptualization and writing (review and editing).

% \section*{Declaration of competing interest} \label{sec compet}
% The authors have no competing interests to declare that could have appeared to influence the work reported in this paper.

\section*{Acknowledgments} \label{sec ack}
We appreciate the support from the Office of Naval Research (award number N000142312485) and National Science Foundation (award number 2238038).

\fi

\section*{Declarations} \label{sec dec}

\noindent \textbf{Conflict of interest} The authors declare that they have no conflicts of interest.

\ifanon
\noindent \textbf{Replication of results} The codes of the model will be made publicly available in an online repository upon publication.
\else
\noindent \textbf{Replication of results} The codes of the model are accessible via \href{https://github.com/Bostanabad-Research-Group/Scalable-Stress-TO}{GitHub} upon publication. 
\fi

    % \clearpage
\section*{Appendix}
\appendix
\section{Sensitivity of the training histories to the number of nodal CPs}\label{app hist mesh}
\begin{figure*}[!b]
    \centering
    \begin{subfigure}[t]{\textwidth}
        \centering
        \includegraphics{figures/legend_2D.pdf}
    \end{subfigure}
    \begin{subfigure}[t]{0.32\textwidth}
        \includegraphics[width=\linewidth]{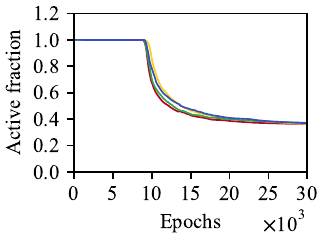}
        \vspace{-14em}
        \captionsetup{justification=raggedright, singlelinecheck=false, skip=-5.3pt, position=top,labelfont=subtxt, textfont=subtxt}
        \caption[]{}
        \label{fig hist mesh L 1}
    \end{subfigure}
    \begin{subfigure}[t]{0.32\textwidth}
        \includegraphics[width=\linewidth]{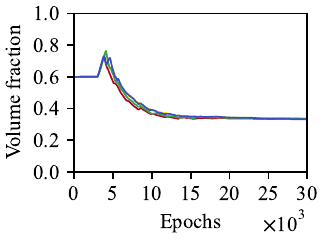}
        \vspace{-14em}
        \captionsetup{justification=raggedright, singlelinecheck=false, skip=-2.9pt, position=top,labelfont=subtxt, textfont=subtxt}
        \caption[]{}
        \label{fig hist mesh L 2}
    \end{subfigure}
    \begin{subfigure}[t]{0.32\textwidth}
        \includegraphics[width=\linewidth]{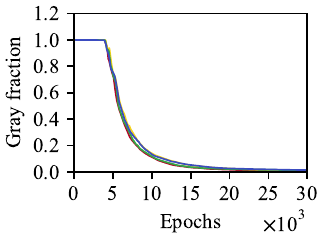}
        \vspace{-14em}
        \captionsetup{justification=raggedright, singlelinecheck=false, skip=-2.9pt, position=top,labelfont=subtxt, textfont=subtxt}
        \caption[]{}
        \label{fig hist mesh L 3}
    \end{subfigure}
    \begin{subfigure}[t]{0.32\textwidth}
        \includegraphics[width=\linewidth]{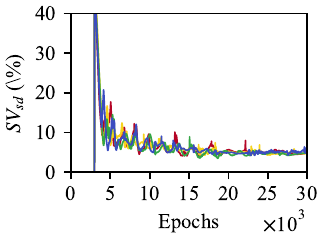}
        \vspace{-14em}
        \captionsetup{justification=raggedright, singlelinecheck=false, skip=-2.9pt, position=top,labelfont=subtxt, textfont=subtxt}
        \caption[]{}
        \label{fig hist mesh L 4}
    \end{subfigure}
    \begin{subfigure}[t]{0.32\textwidth}
        \includegraphics[width=\linewidth]{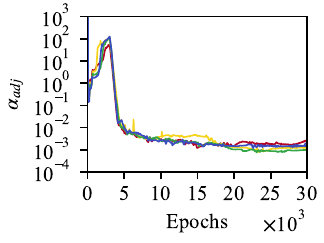}
        \vspace{-14em}
        \captionsetup{justification=raggedright, singlelinecheck=false, skip=-2.9pt, position=top,labelfont=subtxt, textfont=subtxt}
        \caption[]{}
        \label{fig hist mesh L 5}
    \end{subfigure}
    \begin{subfigure}[t]{0.32\textwidth}
        \includegraphics[width=\linewidth]{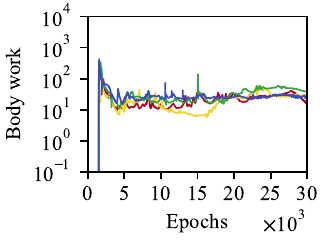}
        \vspace{-14em}
        \captionsetup{justification=raggedright, singlelinecheck=false, skip=-2.9pt, position=top,labelfont=subtxt, textfont=subtxt}
        \caption[]{}
        \label{fig hist mesh L 6}
    \end{subfigure}
    \caption{\textbf{Sensitivity of evolution histories with respect to number of nodal CPs using the L beam benchmark:}
    Evolution histories are compared for the runs with 125K, 250K, 500K, and 1M nodal CPs. The panels report \textbf{(a)} active element fraction, \textbf{(b)} volume fraction, \textbf{(c)} gray area fraction, \textbf{(d)} stress violation over the realized solid material, \textbf{(e)} adjoint scaling factor, and \textbf{(f)} body work contribution in the adjoint problem.}
    \label{fig hist mesh L}
\end{figure*}

\begin{figure*}[!t]
    \centering
    \begin{subfigure}[t]{\textwidth}
        \centering
        \includegraphics{figures/legend_2D.pdf}
    \end{subfigure}
    \begin{subfigure}[t]{0.32\textwidth}
        \includegraphics[width=\linewidth]{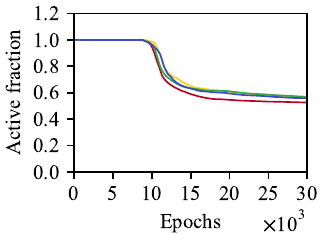}
        \vspace{-14em}
        \captionsetup{justification=raggedright, singlelinecheck=false, skip=-5.3pt, position=top,labelfont=subtxt, textfont=subtxt}
        \caption[]{}
        \label{fig hist mesh hook 1}
    \end{subfigure}
    \begin{subfigure}[t]{0.32\textwidth}
        \includegraphics[width=\linewidth]{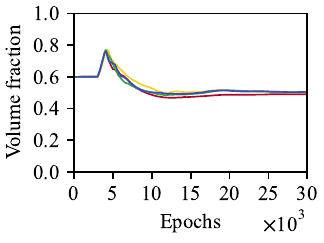}
        \vspace{-14em}
        \captionsetup{justification=raggedright, singlelinecheck=false, skip=-2.9pt, position=top,labelfont=subtxt, textfont=subtxt}
        \caption[]{}
        \label{fig hist mesh hook 2}
    \end{subfigure}
    \begin{subfigure}[t]{0.32\textwidth}
        \includegraphics[width=\linewidth]{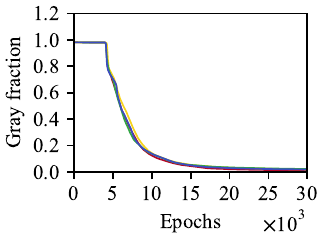}
        \vspace{-14em}
        \captionsetup{justification=raggedright, singlelinecheck=false, skip=-2.9pt, position=top,labelfont=subtxt, textfont=subtxt}
        \caption[]{}
        \label{fig hist mesh hook 3}
    \end{subfigure}
    \begin{subfigure}[t]{0.32\textwidth}
        \includegraphics[width=\linewidth]{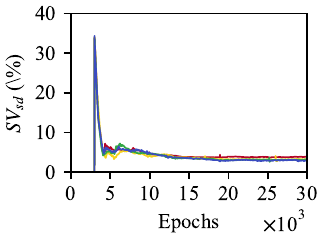}
        \vspace{-14em}
        \captionsetup{justification=raggedright, singlelinecheck=false, skip=-2.9pt, position=top,labelfont=subtxt, textfont=subtxt}
        \caption[]{}
        \label{fig hist mesh hook 4}
    \end{subfigure}
    \begin{subfigure}[t]{0.32\textwidth}
        \includegraphics[width=\linewidth]{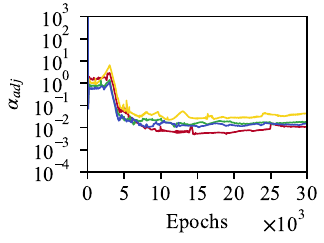}
        \vspace{-14em}
        \captionsetup{justification=raggedright, singlelinecheck=false, skip=-2.9pt, position=top,labelfont=subtxt, textfont=subtxt}
        \caption[]{}
        \label{fig hist mesh hook 5}
    \end{subfigure}
    \begin{subfigure}[t]{0.32\textwidth}
        \includegraphics[width=\linewidth]{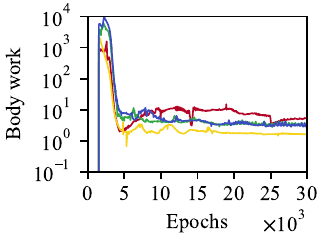}
        \vspace{-14em}
        \captionsetup{justification=raggedright, singlelinecheck=false, skip=-2.9pt, position=top,labelfont=subtxt, textfont=subtxt}
        \caption[]{}
        \label{fig hist mesh hook 6}
    \end{subfigure}
    \caption{\textbf{Sensitivity of evolution histories with respect to number of nodal CPs using the hook benchmark:}
    Evolution histories are compared for the runs with 125K, 250K, 500K, and 1M nodal CPs. The panels report \textbf{(a)} active element fraction, \textbf{(b)} volume fraction, \textbf{(c)} gray area fraction, \textbf{(d)} stress violation over the realized solid material, \textbf{(e)} adjoint scaling factor, and \textbf{(f)} body work contribution in the adjoint problem.}
    \label{fig hist mesh hook}
\end{figure*}

\begin{figure*}[!t]
    \centering
    \begin{subfigure}[t]{\textwidth}
        \centering
        \includegraphics{figures/legend_2D.pdf}
    \end{subfigure}
    \begin{subfigure}[t]{0.32\textwidth}
        \includegraphics[width=\linewidth]{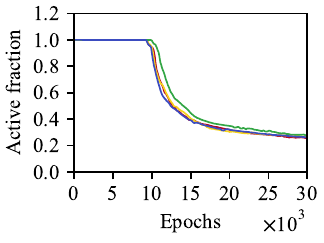}
        \vspace{-14em}
        \captionsetup{justification=raggedright, singlelinecheck=false, skip=-5.3pt, position=top,labelfont=subtxt, textfont=subtxt}
        \caption[]{}
        \label{fig hist mesh portal 1}
    \end{subfigure}
    \begin{subfigure}[t]{0.32\textwidth}
        \includegraphics[width=\linewidth]{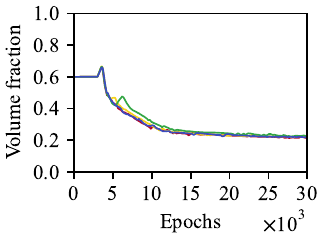}
        \vspace{-14em}
        \captionsetup{justification=raggedright, singlelinecheck=false, skip=-2.9pt, position=top,labelfont=subtxt, textfont=subtxt}
        \caption[]{}
        \label{fig hist mesh portal 2}
    \end{subfigure}
    \begin{subfigure}[t]{0.32\textwidth}
        \includegraphics[width=\linewidth]{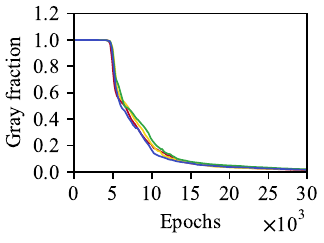}
        \vspace{-14em}
        \captionsetup{justification=raggedright, singlelinecheck=false, skip=-2.9pt, position=top,labelfont=subtxt, textfont=subtxt}
        \caption[]{}
        \label{fig hist mesh portal 3}
    \end{subfigure}
    \begin{subfigure}[t]{0.32\textwidth}
        \includegraphics[width=\linewidth]{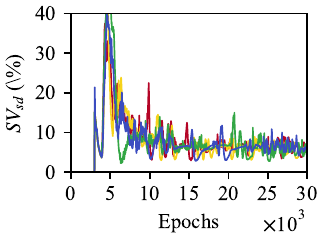}
        \vspace{-14em}
        \captionsetup{justification=raggedright, singlelinecheck=false, skip=-2.9pt, position=top,labelfont=subtxt, textfont=subtxt}
        \caption[]{}
        \label{fig hist mesh portal 4}
    \end{subfigure}
    \begin{subfigure}[t]{0.32\textwidth}
        \includegraphics[width=\linewidth]{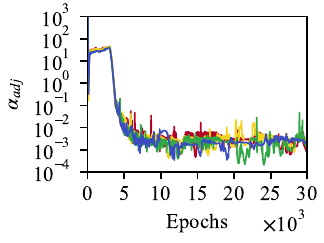}
        \vspace{-14em}
        \captionsetup{justification=raggedright, singlelinecheck=false, skip=-2.9pt, position=top,labelfont=subtxt, textfont=subtxt}
        \caption[]{}
        \label{fig hist mesh portal 5}
    \end{subfigure}
    \begin{subfigure}[t]{0.32\textwidth}
        \includegraphics[width=\linewidth]{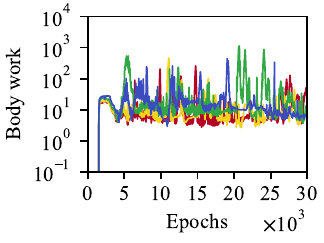}
        \vspace{-14em}
        \captionsetup{justification=raggedright, singlelinecheck=false, skip=-2.9pt, position=top,labelfont=subtxt, textfont=subtxt}
        \caption[]{}
        \label{fig hist mesh portal 6}
    \end{subfigure}
    \caption{\textbf{Sensitivity of evolution histories with respect to number of nodal CPs using the portal frame benchmark:}
    Evolution histories are compared for the runs with 125K, 250K, 500K, and 1M nodal CPs. The panels report \textbf{(a)} active element fraction, \textbf{(b)} volume fraction, \textbf{(c)} gray area fraction, \textbf{(d)} stress violation over the realized solid material, \textbf{(e)} adjoint scaling factor, and \textbf{(f)} body work contribution in the adjoint problem.}
    \label{fig hist mesh portal}
\end{figure*}

\Cref{fig hist mesh L,fig hist mesh hook,fig hist mesh portal} extend the mesh-resolution study of the crack example in \Cref{sec res 2D mesh} by presenting the training histories for the L beam, hook, and portal frame benchmarks with 125K, 250K, 500K, and 1M nodal CPs. Each figure reports the same six quantities as \Cref{fig hist mesh crack 2d}: the active element fraction, volume fraction, gray area fraction, solid stress violation $SV_{sd}$, adjoint scaling factor $\alpha_{adj}$, and body work of the adjoint field. All three benchmarks reproduce the behavior observed in the crack example. The active element, volume, and gray area fractions closely follow common trajectories across the four resolutions. Consequently, the warm-up period, peak volume fraction, material removal, and design sharpening occur at nearly the same stages of training, regardless of the number of nodal CPs. The stress violation also converges to similar levels across the resolutions, approaching $5\%$ for the L beam and portal frame examples, and $4\%$ for the hook example. After the initial design transient, $\alpha_{adj}$ and the adjoint body work remain within narrow bands around common levels, although the coarser cases exhibit somewhat larger late-stage fluctuations. The agreement of the complete training histories, rather than only the final designs, across an eightfold increase in the number of nodal CPs demonstrates that the training dynamics of SNTO are governed primarily by the network parameterization and training schedule rather than by the discretization. This behavior allows the same training schedule to be used for all 2D examples.

\begin{figure*}[!h]
    \centering
    \begin{subfigure}[t]{\textwidth}
        \centering
        \includegraphics{figures/legend_3D.pdf}
    \end{subfigure}
    \begin{subfigure}[t]{0.32\textwidth}
        \includegraphics[width=\linewidth]{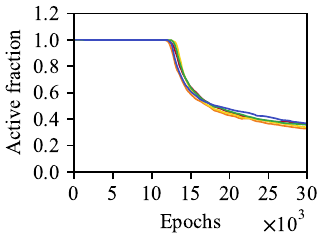}
        \vspace{-14em}
        \captionsetup{justification=raggedright, singlelinecheck=false, skip=-5.3pt, position=top,labelfont=subtxt, textfont=subtxt}
        \caption[]{}
        \label{fig hist mesh portal 3D 1}
    \end{subfigure}
    \begin{subfigure}[t]{0.32\textwidth}
        \includegraphics[width=\linewidth]{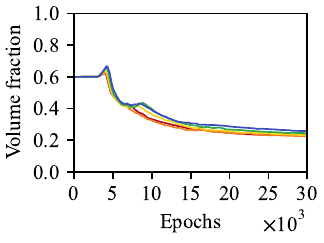}
        \vspace{-14em}
        \captionsetup{justification=raggedright, singlelinecheck=false, skip=-2.9pt, position=top,labelfont=subtxt, textfont=subtxt}
        \caption[]{}
        \label{fig hist mesh portal 3D 2}
    \end{subfigure}
    \begin{subfigure}[t]{0.32\textwidth}
        \includegraphics[width=\linewidth]{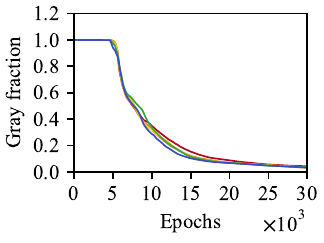}
        \vspace{-14em}
        \captionsetup{justification=raggedright, singlelinecheck=false, skip=-2.9pt, position=top,labelfont=subtxt, textfont=subtxt}
        \caption[]{}
        \label{fig hist mesh portal 3D 3}
    \end{subfigure}
    \begin{subfigure}[t]{0.32\textwidth}
        \includegraphics[width=\linewidth]{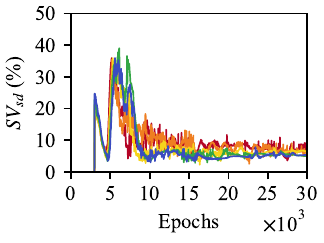}
        \vspace{-14em}
        \captionsetup{justification=raggedright, singlelinecheck=false, skip=-2.9pt, position=top,labelfont=subtxt, textfont=subtxt}
        \caption[]{}
        \label{fig hist mesh portal 3D 4}
    \end{subfigure}
    \begin{subfigure}[t]{0.32\textwidth}
        \includegraphics[width=\linewidth]{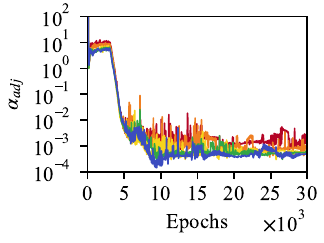}
        \vspace{-14em}
        \captionsetup{justification=raggedright, singlelinecheck=false, skip=-2.9pt, position=top,labelfont=subtxt, textfont=subtxt}
        \caption[]{}
        \label{fig hist mesh portal 3D 5}
    \end{subfigure}
    \begin{subfigure}[t]{0.32\textwidth}
        \includegraphics[width=\linewidth]{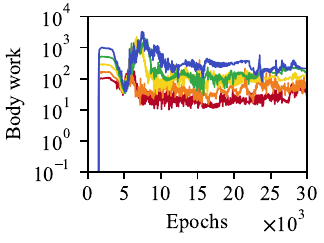}
        \vspace{-14em}
        \captionsetup{justification=raggedright, singlelinecheck=false, skip=-2.9pt, position=top,labelfont=subtxt, textfont=subtxt}
        \caption[]{}
        \label{fig hist mesh portal 3D 6}
    \end{subfigure}
    \caption{\textbf{Sensitivity of evolution histories with respect to number of nodal CPs using the 3D portal frame benchmark:}
    Evolution histories are compared for the runs with 500K, 1M, 2M, 4M, and 8M nodal CPs. The panels report \textbf{(a)} active element fraction, \textbf{(b)} volume fraction, \textbf{(c)} gray area fraction, \textbf{(d)} stress violation over the realized solid material, \textbf{(e)} adjoint scaling factor, and \textbf{(f)} body work contribution in the adjoint problem.}
    \label{fig hist mesh portal 3D}
\end{figure*}

\Cref{fig hist mesh portal 3D} presents the corresponding histories for the 3D portal frame example with 500K, 1M, 2M, 4M, and 8M nodal CPs. The results largely reproduce the behavior observed for the 3D L beam example in \Cref{fig hist mesh L 3D}. Across all five resolutions, the active element, volume, and gray area fractions follow similar trajectories, with consistent timing of the warm-up period, peak volume fraction, material removal, and design sharpening. After the initial transient, the stress violation and adjoint scaling factor also remain within narrow bands around common levels. The adjoint body work, however, depends on the resolution. The 8M case maintains consistently higher values throughout training as shown in \CrefSub{fig hist mesh portal 3D}{fig hist mesh portal 3D 6}. This is the same ordering observed for the L beam and it follows from the normalization of the adjoint force in \Cref{eq alpha adj} as discussed in \Cref{sec res scal}.

%========================================
%=======================================

%========================================
%=======================================

%========================================
%=======================================

%========================================
%=======================================
\clearpage
    % \pagebreak
    \printbibliography
\end{document}